\documentclass{article}

     \PassOptionsToPackage{numbers, compress}{natbib}
\usepackage[preprint]{neurips_2024}

\usepackage[utf8]{inputenc} % allow utf-8 input
\usepackage[T1]{fontenc}    % use 8-bit T1 fonts
\usepackage{hyperref}       % hyperlinks
\usepackage{url}            % simple URL typesetting
\usepackage{booktabs}       % professional-quality tables
\usepackage{amsfonts}       % blackboard math symbols
\usepackage{nicefrac}       % compact symbols for 1/2, etc.
\usepackage{microtype}      % microtypography
\usepackage{xcolor}         % colors
\usepackage{amsmath}
\usepackage{graphicx}
\usepackage{tikz}
\usepackage{wrapfig}
\usepackage{subcaption}

\usetikzlibrary{arrows.meta, positioning}
\usepackage{amsmath,amsfonts,bm}

\def\eqref#1{equation~\ref{#1}}
\def\1{\bm{1}}

\def\mA{{\bm{A}}}

\def\mE{{\bm{E}}}

\def\mI{{\bm{I}}}

\def\mM{{\bm{M}}}

\def\mW{{\bm{W}}}
\def\mX{{\bm{X}}}
\def\mY{{\bm{Y}}}

\DeclareMathAlphabet{\mathsfit}{\encodingdefault}{\sfdefault}{m}{sl}
\SetMathAlphabet{\mathsfit}{bold}{\encodingdefault}{\sfdefault}{bx}{n}

\def\sR{{\mathbb{R}}}

\newcommand{\R}{\mathbb{R}}

\usepackage{amssymb}
\usepackage{mathtools}
\usepackage{amsthm}
\theoremstyle{plain}
\newtheorem{theorem}{Theorem}[section]

\theoremstyle{definition}

\theoremstyle{remark}

\title{Identifying Sub-networks in Neural Networks via Functionally Similar Representations}

\author{%
  Tian Gao, Amit Dhurandhar, Karthikeyan Natesan Ramamurthy, Dennis Wei\\
  IBM Research\\
  \texttt{ \{tgao, adhuran, knatesa, dwei\}@us.ibm.com} \\
}

\begin{document}

\maketitle

\begin{abstract}
Providing human-understandable insights into the inner workings of neural networks is an important step toward achieving more explainable and trustworthy AI.
Existing approaches to such mechanistic interpretability typically require substantial prior knowledge and manual effort, with strategies tailored to specific tasks. In this work, we take a step toward automating the understanding of the network by investigating the existence of distinct  sub-networks. Specifically, we explore a novel automated and task-agnostic approach based on the notion of functionally similar representations within neural networks to identify similar and dissimilar layers, revealing potential sub-networks. We achieve this by proposing, for the first time to our knowledge, the use of Gromov-Wasserstein distance, which overcomes challenges posed by varying distributions and dimensionalities across intermediate representations—issues that complicate direct layer-to-layer comparisons.
On algebraic, language, and vision tasks, we observe the emergence of sub-groups within neural network layers corresponding to functional abstractions. 
Through downstream applications of model compression and fine-tuning, we show the proposed approach offers meaningful insights into the behavior of neural networks with minimal human and computational cost.  %\footnote{Code will be made publicly available.}
\end{abstract}

\section{Introduction}
Rapid progress in transformer language models has directed attention towards understanding the underlying causes of new capabilities. Like many other neural methods, large language models (LLMs) are mostly black-box models and explainable artificial intelligence aims to offer insights and improve human understanding of these LLMs. Recently, mechanistic interpretability %(MINT)
research has gained popularity,  focusing on reverse-engineering
models into human-understandable algorithms, using methods such as computational graphs and circuits \citep{conmy2024towards}.

Current approaches primarily rely on extensive manual inspection and trial-and-error to reverse engineer  networks by discovering a sequence of learned functions that produce a desired output. This process requires significant human prior knowledge with strategies tailored to specific tasks such as algebraic problems~\citep{charton2023learning}. Automatic discovery of these functions is a difficult task due to the large search space involved. We investigate a simpler task that can be considered as an initial step towards automating the discovery of these functions: \textit{is it possible  to detect how many distinct (complex) functions exist in a learned network, and which layers correspond to each such function?} Understanding neural networks through the identification of subnetworks is essential due to the complexity and opacity of modern deep learning models. Neural networks, especially those with many layers and parameters, often exhibit behaviors that are difficult to interpret holistically. Identifying subnetworks allows us to break down the model into smaller, more interpretable units, providing insights into how individual components contribute to the model’s overall performance.

To address the above question, we propose an automatic and task-agnostic approach to identify 
sub-components in neural networks that are functionally different from each other. Specifically, we treat the intermediate layers of neural networks as computing different \emph{functions} of the input and base our approach on  comparing \emph{functional similarity}. These distinctive subnetworks would represent one function in a sequence of different ones, and knowing  such a partition could facilitate further decoding these functions leading a better understanding of the network. 
\begin{figure}
    \centering
    \includegraphics[width=0.8\textwidth]{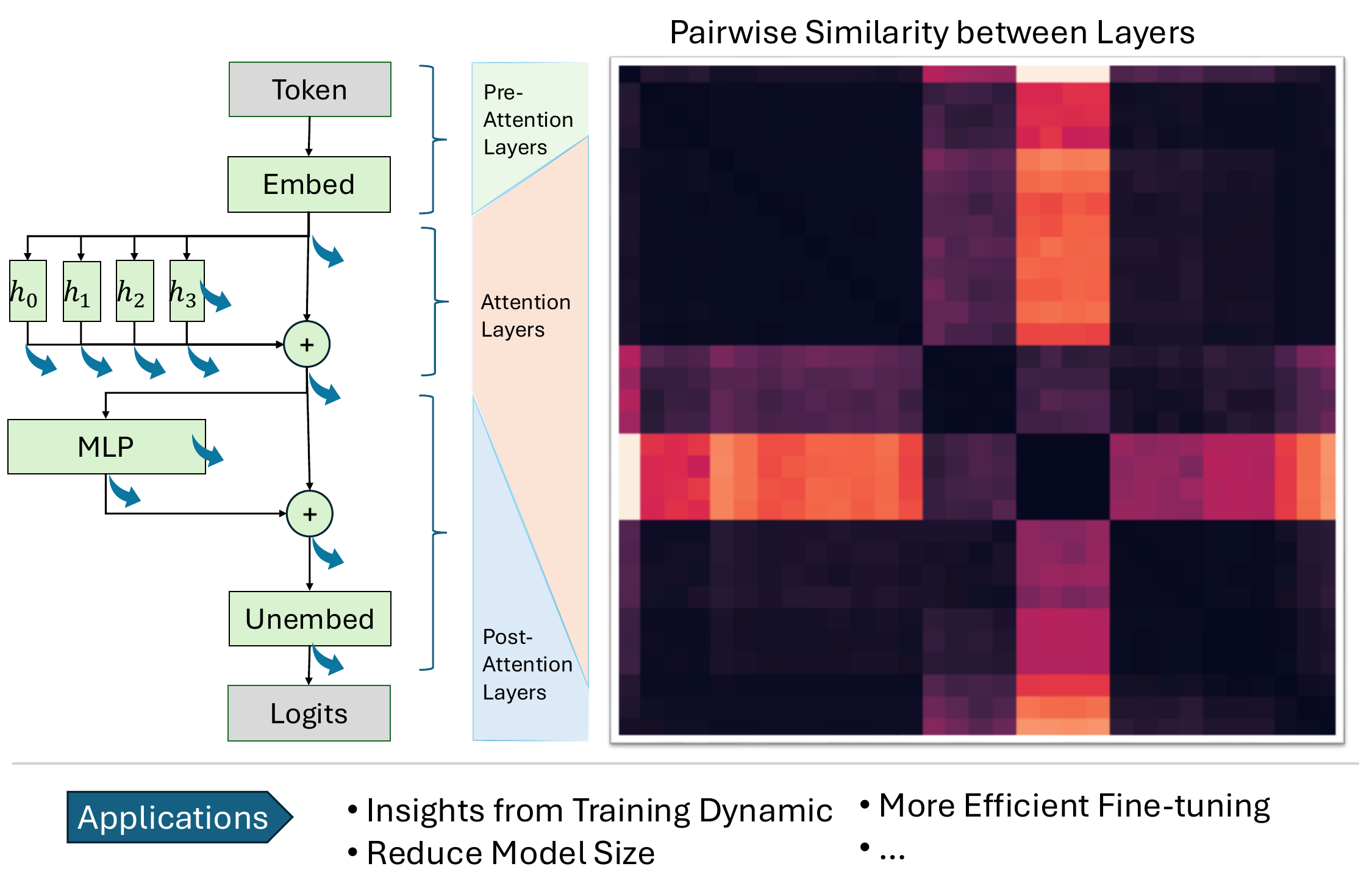}
    \caption{\small{Overview of our approach, where we use  representations from different neural network layers to identify functionally distinct sub-networks (darker blocks) leveraging GW distance. The figure is an illustration based on a single  transformer block, and the proposed technique can be applied to other types of network structures.}}
    \label{fig:overview}
\end{figure}

The nature of functional similarities differ depending on which of the two cases is true: either we have a hypothesized target function that part of the network may (approximately) compute, or we lack such a hypothesis. In the case of a known target function,  functional similarity relates to the output of a target function, and the problem is one of search to find a network layer that best approximates it.   The use of probes attached to representations is a popular method for detecting such similarities. 

The more realistic case with neural networks is the absence of a target function, which is the primary focus of this paper. In this case, we propose to measure the functional similarity \emph{between} layers of the network, which is akin to comparing two functions based on the values they take for the same input samples. The problem then becomes to identify distinctive sub-networks by finding which layers are functionally less similar to previous layers as shown in Figure \ref{fig:overview}. The idea is that less similar layers (brighter colors) may indicate boundaries of a sub-component, while more similar layers (darker colors) likely belong to the same sub-component. While it may be tempting to also use probes here for layer comparisons, the lack of targets makes it unclear how the representations in these layers should be transformed by the probes for comparison. We need a more direct way to compare these representations.

We propose to measure function similarity using Gromov-Wasserstein (GW) distance \citep{gwsurvey} between representations from different layers of the network. 
As elaborated further in Section~\ref{sec:gw_property}, GW allows distance computation between distributions supported on two different metric spaces with different supports and potentially different dimensions, which is common across different layers in neural networks. GW is also invariant to permutation of the representation within a layer, a crucial property since neural networks are known to have permutation symmetries \citep{goodfellow2016deep}. %When the function target is not known, The GW distance can be used to find different sub-components within a network and show differences in representations by models trained with different strategies.
As such, GW can effectively identify genuinely distinct behaviors across (groups of) layers.

We validate our approach on algebraic, language, and vision tasks, showing that GW distance provides a systematic way to analyze and identify subnetworks.  Additionally, our findings provides a holistic view on differences in representations of models trained with different strategies. 
We observe clear patterns in the form of block structures among different layers, suggesting there exist sub-networks that have different functions, particularly at the transition layers where major functional changes may occur. Moreover, our approach is applicable to various downstream tasks, such as tracking the emergence of subnetworks during training process (\S~\ref{sec:emergence_training}) and identifying potentially redundant layers in model compression and fine-tuning (\S~\ref{exp:yelp}) Overall,  our method can improve the efficiency of mechanistic interpretation by finding subnetworks in larger models, reducing the need for extensive human effort and contributing to a further understanding of neural network behaviors.

\section{Background and Related Work}
With its popularity, mechanistic interpretability has become a disparate area, with many different applications in vision~\citep{palit2023towards}, and language~\citep{ortu2024competition,hernandez2023linearity,yu2024mechanisms}, and we survey a few closely related directions within it.  %\red{update related works}

\noindent\textbf{Algorithm discovery in algebraic problems} Modular sum algorithms \citep{nanda2023progress,zhong2024clock} and related math problems \citep{he2024learning,charton2023learning} are often studied in mechanistic interpretability due to their precise function definitions. These works have shown that there may exist many different sub-functions in the computation. However, most of these works require extensive human effort to reverse engineering these algorithms. We investigate whether we can automate the discovery process by studying a simpler problem of sub-function detection.

\noindent\textbf{Subnetwork and Neuron Semantics} Automated circuit discovery \citep{conmy2024towards,shi2024hypothesis} aims to find a computational graph that is much more sparse without sacrificing performance. Heuristic core \citep{bhaskar2024heuristic} hypothesizes that there exists a set of attention heads that encompasses all subnetworks, learning shallow and non-generalizable features. 
Moreover, Various work have explored neuron semantics and possible disentanglement of semantics ~\citep{bricken2023monosemanticity,dreyer2024pure,huang2024ravel} and concepts~\citep{park2024geometry}. Instead of studying circuits and neurons, we investigate differences among neural network layers as a whole, based on an existing line of work \citep{nanda2023progress}. Related, block structures within neural network layers have been observed in previous studies \citep{nguyen2022origins}.

\noindent\textbf{Methods of Studying Mechanisms} Weight inspection and manipulation \citep{ortu2024competition,meng2022locating} are  commonly employed techniques to gain insights into the inner work of networks, including studying periodicity \citep{nanda2023progress} and  weight gradients \citep{zhong2024clock}. Modifying or ablating the activation of a specific model components \citep{huang2024inversionview,kramar2024atp}, including attention knockout \citep{wang2022interpretability}, and even direct modification of attention matrix~\citep{ortu2024competition,geva2023dissecting} are prevalent. Another popular approach involves representation and output inspection~\citep{meng2022locating}, including logit patterns~\citep{zhong2024clock}, residual stream \citep{ortu2024competition}, and periodicity \citep{nanda2023progress}. Causal mediation analysis is used to compute the indirect treatment effect \citep{meng2022locating,yu2024mechanisms}  with perturbed embedding. %, often with Gaussian noises. 
%\citep{nanda2023progress} studies the similarity between output and a known function (trigonometric identity).
Rather than just inspection of outputs, many works have proposed to map the output to some target and is a popular technique for analyzing how neural activations correlate with high-level concepts \citep{huang2024ravel,hou2023towards}. Linear probes are generally used. \citep{hou2023towards} uses a nonparametric probe ($k-$nearest-neighbor) to classify outputs for reasoning tasks. 
Here we instead focus on studying similarity measure between layers.

\noindent\textbf{Similarity Measure between Neural Network Layers}   There are studies that quantify the similarity between different groups of neurons \citep{klabunde2023similarity}, typically layers \citep{ding2021grounding,yu2024mechanisms}, to compare different neural networks.
Generally a normalized representation with desired properties is used to compare different transformer blocks, such as invariance to invertible linear transformation in canonical correlation analysis \citep{morcos2018insights}, orthogonal transformation, isotropic scaling, and different initializations in centered kernel alignment  \citep{kornblith2019similarity}. Other measures include representational similarity analysis \citep{mehrer2020individual}, which studies all pairwise distances across different inputs. Wasserstein distance has been explored in measuring similarities in the context of neural networks~\citep{dwivedi2019representation,cao2022towards,lohit2022model}, but they assume that different layer representations belong to the same metric space, which is very unlikely even if they have the same dimensionality as the semantics captured by each layer are likely to differ significantly. Several similarity measures \citep{tsitsulin2019shape,demetci2023revisiting} are related to, but not exactly equivalent to, GW distance. While GW distance has been used for model merging as a regularization~\citep{singh2020model,stoica2023zipit}, it has not been fully explored in the area of mechanistic interpretability, particularly  for the subnetworks identification.

\noindent\textbf{Similarity Measure between Neural Network Layers}  Similarity measure between layers have been studied. Cross product between each layer output and final output \citep{yu2024mechanisms} is used to approximate each layer's contribution to the final prediction. There are studies  that quantify the similarity between different groups of neurons \citep{klabunde2023similarity}, typically layers \citep{ding2021grounding}, to understand and compare different neural networks. Generally a normalized representation is used to compare different transformer blocks, with different desired invariance properties, such as invariance to invertible linear transformation in canonical correlation analysis \citep{morcos2018insights}, orthogonal transformation, isotropic scaling, and different initializations in centered kernel alignment  \citep{kornblith2019similarity}. Other measures include representational similarity
analysis \citep{mehrer2020individual}, which studies all pairwise distances across  different inputs.  Wasserstein distance has been explored in measuring similarities in the context of neural networks~\citep{dwivedi2019representation,cao2022towards,lohit2022model}, but they assume that different layer representations belong to the same metric space, which is very unlikely even if they have the same dimensionality as the semantics captured by each layer are likely to differ significantly.

\section{Functional Similarity Within Neural Networks }
We aim to identify  sub-components of a neural network based on their mechanistic functions. When we have a hypothesized target function that the sub-component may compute, this can be formulated as a similarity search problem. %In this context, we search for candidate representations within neural networks that represents changes in function computation. 
The search problem consists of three key elements: the search space, the search target, and the similarity measures used to evaluate how closely the candidates in the search space match the target.   We first discuss  the notation of function similarity as the measure for searching and the search space for candidates. We then discuss specific choices on the similarity measure, given whether the search target is known or not.

\subsection{Similarity Measures}
Let $f: X \rightarrow Y$ be a function that map $x$ in a set of input $X = \{x_i: x_i \in \sR^{d_x } \}_{i=1}^{n} $ to $y$ in a set of output $Y = \{y_i: y_i \in \sR^{ d_y}\}_{i=1}^{n}$.  Each element in $Y$ and $X$ are assumed to be a vector with dimensions $d_y$ and $d_x$, respectively, with $n$ being the set size, without loss of generality. Let $f_0: X \rightarrow Y^0$ be another function that produces $Y^0 = \{y^0_i: y^0_i \in \sR^{ d_y}\}_{i=1}^{n}$ given $X$. Note sets can be concatenated into matrix forms as $Y^0, Y \in \sR^{n\times d_y}$ and $X \in \sR^{n \times d_x}$.

% \blue{
% \begin{definition} \label{ref:funcsim}
%  \textbf{Function similarity}.  We say two functions, $f_0$ and $f$, with a set of input $X$, are \textit{similar} if and only if $Y_0 = f_0(X)$ and $Y = f(X)$ are similar, with respect to $X$.  $Y_0$ and $Y$ are similar if $d(Y_0, Y) \leq \epsilon$.  In addition, $f$ is more similar to $f_0$ than function $\tilde{f} $ with $\tilde{Y} = \tilde{f}(X)$ ,with respect to $X$, if $d(Y_0, Y) < d(Y_0, \tilde{Y})$.
% \end{definition}
% }
% \blue{
% \begin{definition} \label{ref:funcsim}
%  \textbf{Function similarity}. Similarity between two functions, $f_0$ and $f$, over a  set  of input $X$,  can be measured by the similarity between their outputs $Y_0 = f_0(X)$ and $Y = f(X)$.
% \end{definition}
% }

\textbf{Functional Similarity.} Similarity between two functions, $f_0$ and $f$, over a  set  of input $X$,  can be measured by the similarity between their output sets $Y_0 = f_0(X)$ and $Y = f(X)$.

% \red{Choose one of above: formal statement or just bold?}

We can use a scoring or distance function $D(Y^0, Y)$ as a measure between output similarity and hence the functional similarity between $f_0$ and $f$, where we regard $Y^0$ as the function/search target. If they are close according to $D$, then the function values should be similar to each other locally at a set of points $X$. Otherwise, these functions should be different at $X$. Popular measures such as Euclidean distance have been used for this purpose \citep{klabunde2023similarity}. Each intermediate representations of a neural network can be naturally treated as function outputs, given inputs $X$.

\textbf{The Need for Complex Functional Similarity Measure.} Since we cannot exactly control the behavior of a trained neural network, the layer-wise functions $f$ that it learns can be complex and thus the learned representation $Y$ from each layer may be a complex function of the target $Y^0$ rather than a simpler transformation. For example, let $Y^0 = \sin(X)$ and a candidate $Y = \sin^2(X) = (Y^0)^2$. They share strong similarity, but a linear transformation will not be able to capture their functional similarity. If we want to truly understand where function $f_0$ might be approximately computed, we should consider functions of target $Y^0$, but naively listing out all possibilities can be prohibitive. As a consequence, one may need to use more complex measures to deal with such a space. %, for which we will elaborate on in Section~\ref{sec:searchtargets} together with search targets. % In appendix~\ref{sec:appendix_knowntarget}, we show  use non-linear probes of limited capacity and show a study on the $f_{\text{mod3}}$ dataset with Models 1 and 2 that motivates the need for non-linear probes. 

\subsection{Search Space}
 We consider multiple candidate $Y$'s to form the search space for target $Y^0$.  In the context of MLP neural networks for example, where $\sigma(.)$ denotes the non-linearity and $W$s are the parameter matrices, we have $Y^* = W_n (\sigma(W_{n-1} \dots \sigma(W_1 X)))$ for the whole network. We can extract many $Y$'s from intermediate functions of the model, for instance $Y_1 = W_1X$, $Y_2 = \sigma(W_1 X)$, and so on. These $Y$'s are often called representations, activations, or sometimes even ``outputs'' from each layer. We use these terms interchangeably here. For attention modules in transformer neural networks \citep{vaswani2017attention}, we can similarly extract $Y$'s from attention key, query, and value functions as well as MLP functions. 
We list the exact  equations and locations of representations considered in the transformer models in Table~\ref{table:location} in Appendix~\ref{apendix:rep}, which serves as the focus of this paper. We consider attention-based models first, and later we also consider convolutional neural networks with residual layers, with candidate representations listed in Table~\ref{table:location_resnet} in in Appendix~\ref{apendix:rep}.

\subsection{Search Targets}
\label{sec:searchtargets}
 \paragraph{Known Targets}
When  the  search target, denoted as $T$, is a  value from a known function,  we can directly compare outputs between representations from each  layer and  known function output $T$.  Representations from each layer can be directly compared with the target via a probe. Popular linear probes can be used to assess the similarity between a target and any layer's representation.  For instance, linear regression can be used to model each target $T$ from each representation candidate $Y$, and the residual error is used as the search criteria  between $Y$ and $T$. As discussed previously, to deal with the potentially large search space of functions of the target, a more powerful probe %(such as a nonlinear MLP function)
may have to be used so that it can detect more complex similarities to $T$. 

% \paragraph{Probes} 
% Popular linear probes can be used to assess the similarity between a target and any layer's representation.  We perform linear regression of each target $T$ on each representation candidate $Y$, and report the residual error as the search criteria  between $Y$ and $T$. As discussed previously, to deal with the potentially large search space of functions of the target, a more powerful probe (such as a nonlinear MLP function) may have to be used so that it can detect more complex similarities to $T$. %Therefore, we use a two-layer MLP\footnote{We use the neural network classifier from the scikit-learn package, with default parameters, to constraint the power of neural models, to prevent it reproduce the original network.} to predict $T$'s. Our experiment show indicates that non-linear probes can be used to find subgroups of layers while linear probes fail due to limited power (refer to \ref{sec:appendix_knowntarget} for the experimental details). Unlike existing work that primarily focuses on linear probes, we show that non-linear probes, still with limited capacity, are useful. \red{TODO: remove mentioning of probes entirely}

% \blue{include raw values in appendix? maybe in a figure}
%usefulness on scale and complexity: multiple-levels, explanation in term of audience understanding and complexity, pattern emerging at ensemble/high level,

One challenge  with using predictive probes to compute the distance measure $D$ is that the target function must be known. In  practice, however, we often lack knowledge of specific targets. While it’s possible to experiment with various target functions with power probes,
the vast  number of potential targets makes this approach inefficient. This calls for an alternative strategy to distinguish sub-components in a network through representation similarity.

\section{ Gromov-Wasserstein Distance as a Similarity Measure}
\paragraph{Unknown Targets } When the search target is unknown,  %functionally similar parts cannot be identified by comparison to a predefined set of target functions. Instead, 
we propose to identify the similarities and subnetworks among the representations at each intermediate layer. Each layer, however, posits a representation that potentially has a different distribution, not to mention even different dimensionality depending on the architectures and layers one considers (viz. mlp  and attention layers in transformer blocks). Consequently, representations across layers may be incomparable using standard distance metrics, such as the $\ell_p$ norm amongst others.

To address  these challenges, we propose computing distances between representations at the same layers for  different inputs, and match the vertices of a weighted graph -- where each dimension of the representation  are vertices and the distances indicate weights on the edges -- with the vertices of a similarly constructed weighted graph  from another layer. Essentially,  we assume the representations in  a layer are  samples of the underlying distribution, and we want the best permutation of representation dimensions in one layer that aligns with vertices in another layer, thereby deriving the inter-layer distance. If this inter-layer distance is low, then we consider  the two layers  functionally similar for the given input data. 

Formally, without loss of generality, let $Y_1 = \{y_{1i}: y_{1i} \in \sR^{ d_1}\}_{i=1}^{n}$  and $Y_2 = \{y_{2i}: y_{2i} \in \sR^{ d_2}\}_{i=1}^{n}$ be representations of $n$ examples from two different layers, where the discrete distributions over the representations are $\mu_1$ and $\mu_2$ respectively, with dimension $d_1$ possibly being different from $d_2$. Direct distance computation between them is not reasonable. Instead, we seek  to compute a coupling or matching $\pi\in \Pi(\mu_1, \mu_2)$ between the $n$ examples in each set such that given the pairwise distances $D_1, D_2\in \sR^{n\times n}$ within representations $\mY_1$ and $\mY_2$ respectively, the sum of differences between the distances of the matched examples is minimized. Loosely speaking, we aim to find a matching that preserves the pairwise distance as much as possible. In particular, we want to minimize the following:
\begin{align} \label{eq:gw_def}
\rho&(\mY_1,\mY_2,\mu_1,\mu_2,D_1,D_2)  \triangleq \nonumber\\ \min\limits_{\pi\in \Pi(\mu_1, \mu_2)} & \sum_{i,j,k,l}(D_1(i,k)-D_2(j,l))^2\pi_{i,j}\pi_{k,l} \nonumber\\
    &\text{s. t.} \quad \pi \mI = \mu_1; \pi^T\mI = \mu_2; \pi \geq 0.
\end{align}
%where, $w_1$ and $w_2$ represent the weights of each sample in the source and target distribution (positive and sum to $1$). 

It turns out that $\rho$  corresponds to  the Gromov-Wasserstein (GW) distance \citep{gw}, used to map two sets of points in optimal transport. We thus utilize  this distance as a measure of inter-layer functional similarity in the setting where the target is unknown.
%have the same dimensionality and distribution, which is a strong assumption that we do not make.

\subsection{Justification for GW Distance as a  Functional Similarity Measure}
\label{sec:gw_property}

%\red{  more formal statement on justification with  GW over other metrics } much broader assumptions on changes across layers, encompasses other distance functions, ground distance = euclidean (other distance possible)

Let $(\mY_1, D_1, \mu_1)$ and $(\mY_2, D_2, \mu_2)$ be two given metric measure space (mm-space), where $(\mY, D)$ is a compact metric space and $\mu$ is a Borel probability measure with full support: $\text{supp} (\mu) = \mY$. An isomorphism between $\mY_1, \mY_2$ is any isometry  $\Psi: \mY_1 \rightarrow \mY_2$, i.e., a distance-preserving transformation between metric spaces, such that $\Psi_{\#\mu_{1}} = \mu(\Psi^{-1} ) = \mu_{2}$.

\begin{theorem}
   \citep{memoli2011gromov}.  The Gromov-Wasserstein distance in \eqref{eq:gw_def} defines a proper distance on the collection of isomorphism classes of the mm-spaces.  
\end{theorem}

\textit{Remark.} The Gromov-Wasserstein distance itself is defined on isomorphism-classes of metric measure spaces, which means that any distance preserving (isometric) transformation of a space should preserve GW distance between the points in that space and any other space~\citep{memoli2011gromov}. These isometric transformations include rigid motions (translations and rotations) and reflections or compositions of them. Additionally, permutations of points in a space also preserve GW distances, as the points are unlabeled. Hence, GW distance captures much richer transformations across layers.

The computed GW distance represents the minimal distance over all possible transportation plans between two sets of points from different spaces. % In our context, we can also view
Since GW can be viewed as a measure that quantifies the distance between  distance-based (i.g., Euclidean-distance)  graphs, with a set of points as its nodes, GW distance would be low if the graph undergoes (nearly) isomorphic transformations between layers. Conversely, a high GW distance indicates a non-distance preserving transformation across layers,  potentially reflecting a highly non-linear operation. While  GW distance does not reveal the exact function operation, it highlights specific layers for further investigations.

\begin{wrapfigure}{r}{0.3\textwidth}
    %  \begin{subfigure}{.5\linewidth}
    %     \includegraphics[width=\linewidth]{figs/YELP_bert-base-uncased_target_unknown_spectral_GW_pre_scores_histo.png}
    %     \caption{Pretrained}
    %     \label{fig:yelp0_hist}
    % \end{subfigure}\hfill
    % \begin{subfigure}{.5\linewidth}
        \includegraphics[width=\linewidth]{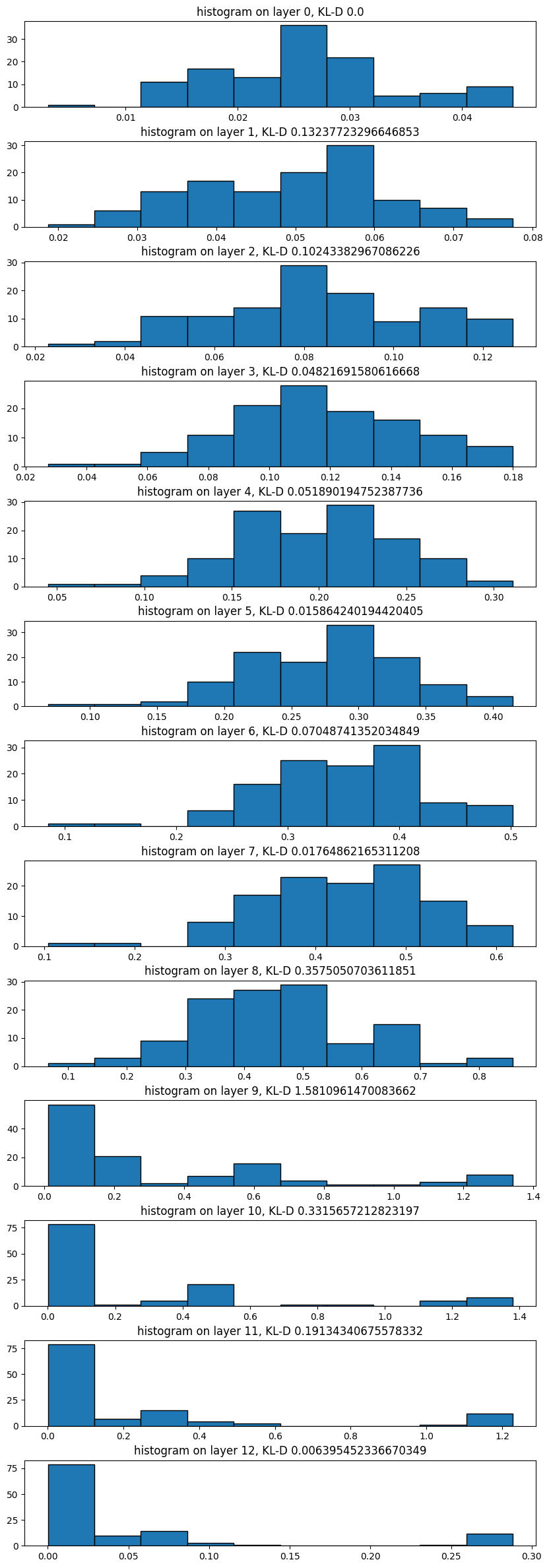}
        % \caption{Fine-Tuned}
        \label{fig:yelp1_hist}
    % \end{subfigure}\hfill
    % \begin{subfigure}{.33\linewidth}
    %     \includegraphics[width=\linewidth]{figs/YELP_bert-base-uncased_target_unknown_spectral_GW_scores_tsne.png}
    %     \caption{tSNE projection}
    %     \label{fig:yelp2_hist}
    % \end{subfigure}\hfill
    % \begin{subfigure}{.20\linewidth}
    %     \includegraphics[width=\linewidth]{figs/YELP_bert-base-uncased-sparse70_target_unknown_spectral_GW_scores_histo.png}
    %     \caption{}
    %     \label{fig:yelp3_hist}
    % \end{subfigure}\hfill
    % \begin{subfigure}{.20\linewidth}
    %     \includegraphics[width=\linewidth]{figs/YELP_bert-base-uncased-sparse95_target_unknown_spectral_GW_scores_histo.png}
    %     \caption{}
    %     \label{fig:yelp4_hist}
    % \end{subfigure}
    \caption{\small{Histogram on pairwise distances for outputs from all transformer blocks in a fine-tuned BERT model trained on YELP dataset.}} % $c)$ shows the neighborhood of tSNE projection for a batch of data across layers.}}
    \label{fig:yelp_hist}
\end{wrapfigure}

\noindent\textbf{Favorable Properties of GW.} Besides the above noteworthy property of GW, it also has other favorable properties \citep{gwsurvey,gw}: 
%\begin{enumerate}
    \noindent i) It is symmetric and satisfies the triangle inequality. ii) It is invariant under any isometric transformation of the input, which is advantageous because we do not want rotations and reflections to affect our similarity search. This invariance  also includes permutation invariance, which is a beneficial  property since we want the distance between layer representations to remain unaffected by  permutations within the representations in each layer.
    %and not necessarily the exact mapping between the representations across layers.
    % \item It is permutation invariant in the sense that it mainly matches pairwise distances that is edges and hence vertices could be permuted in some situations leading to the same GW distance. This is acceptable to us since we mainly care about the distance between layer representations and not necessarily the exact mapping.
    iii) GW is scalable since it does not require estimating high-dimensional distributions,  which is often the case with intermediate representations in large models; instead, it only compare them to obtain a distance measure.
    %as the mapping occurs by comparing $1$-dimensional real values (i.e. pairwise distances).  
    iv) GW is monotonic in (positive) scaling of pairwise distances, and hence the same layers should appear to be closer than others even with scaling.

\noindent\textbf{Distance Distributions.} As an illustrative example, we plot the histogram on pairwise distances for a batch of samples across all transformer blocks in BERT models from the YELP review dataset in Figure~\ref{fig:yelp_hist}. For more details on YELP, we  provide a comprehensive discussion of experiments  \S~\ref{exp:yelp}.    The results in Figure~\ref{fig:yelp_hist} show the distributions on pairwise distances begin to differ from block 9, consistent with GW distance observed in Figure~\ref{fig:yelp_w},  suggesting that  significant transformations occur and can be effectively captured by GW. We include the full results and discussion in Appendix~\ref{appendix:alignment}.
     
%         \vskip -2mm
% \begin{figure}
%     \centering
%     \includegraphics[width=0.4\linewidth]{figs/YELP_bert-base-uncased_target_unknown_spectral_GW_scores_histo.png}
%       \caption{\small{Histogram on pairwise distances for outputs from all transformer blocks in a fine-tuned BERT model trained on YELP dataset.}} % $c)$ shows the neighborhood of tSNE projection for a batch of data across layers.}}
%     \label{fig:yelp_hist}
%     \vskip -10mm
% \end{figure}

% \vspace{-.1cm}
%     \centering
%     \includegraphics[width=0.35\textwidth]{figs/gw_model0.png}
%     \caption{\small{Model 0 Pairwise GW distance.}}
%     \label{fig:sidecap}
%     \vspace{-.5cm}
% \end{wrapfigure}

\noindent\textbf{Neighborhood Change.} Complementary to the distribution of  pairwise distances, the changing representations of samples could also alter their relative neighborhoods across transformer blocks. We plot a tSNE projection \citep{van2008visualizing} of representations from  a batch of samples on YELP, and visualize it in Figure~\ref{fig:yelp_hist}b and Figure~\ref{fig:yelp4_tsne} of Appendix~\ref{sec:appendix_align}. The Jaccard similarity, measuring the overlap between top-5-neighbors of 3 selected samples across different transformer blocks, ranges from 0.0 to 0.43, with average values of $\{0.27, 0.26, 0.26\}$. The full details are shown in Table~\ref{table:jaccard_tSNE}  of Appendix~\ref{sec:appendix_align}. Hence, the sample neighborhood changes across blocks, which can be indicative of functional changes that are not captured by comparing distributions alone. However, GW can account for such changes as well. 

\noindent\textbf{Computation Details.} We use an existing optimal transport toolbox, \url{pythonot} \citep{flamary2021pot}, for computing GW distance. Specifically, we use an approximate conditional gradient algorithm proposed in \citep{titouan2019optimal}, which has a complexity of 
$O(mn^2 + m^2n)$, where $m$  and $n$ are the dimensions of two spaces (here the number of data samples from two layers being compared). In comparison, the Wasserstein distance \cite{lohit2022model} may require $O(n^3 log(n))$ for exact computation. %, and the Sinkhorn solver \citep{cuturi2013sinkhorn} operates at nearly  $O(n^2)$. 
When the dataset is large, we can also sub-sample the dataset to improve the computational efficiency. % (for example, we choose 1000 examples randomly for YELP datasets). 

\section{Empirical Study and Findings}

We compare the proposed similarity  measure for sub-network identification against a set of  baselines across multiple datasets, including those from algebraic operation, NLP, and computer vision tasks. For a list of baselines with their implementation details, please see Appendix~\ref{appendix:baselines}. 

\subsection{Synthetic Modular Sum Tasks}
\begin{figure*}
   % \caption{\small{Pairwise GW distance on Yelp, across BERT models. (a) is pre-trained BERT and (b)-(e) are fine tuned BERT models with increasing sparsity (dense, $25\%$, $70\%$ and $95\%$ sparsity).}}
   % \label{fig:yelp}
% \end{figure}
% \begin{figure}[h]
 \vskip -3mm
  \begin{subfigure}{.30\linewidth}
        \centering
        \includegraphics[width=\textwidth]{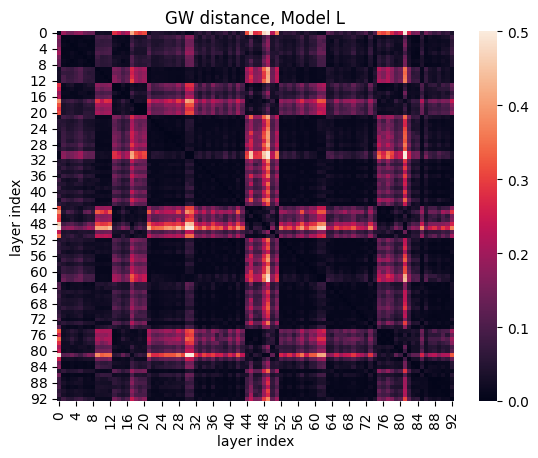} % First figure
        \caption{\small{Model L (layer-wise training) pairwise GW distance, on $f_{\text{mod3}}$ dataset.}}
        \label{fig:1}
    \end{subfigure}\hfill
      \begin{subfigure}{.30\linewidth}
        \centering
        \centering
        \includegraphics[width=\textwidth]{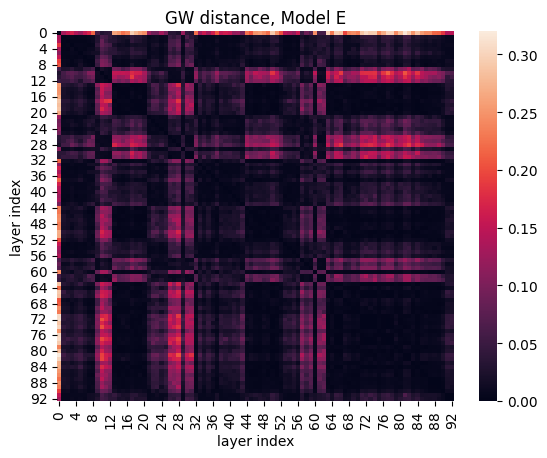} % Second figure
        \caption{\small{Model E (end-to-end training) pairwise GW distance, on $f_{\text{mod3}}$ dataset.}}
          \label{fig:2}
    \end{subfigure}\hfill
     \begin{subfigure}{.30\linewidth}
        \centering
              \includegraphics[width=\textwidth]{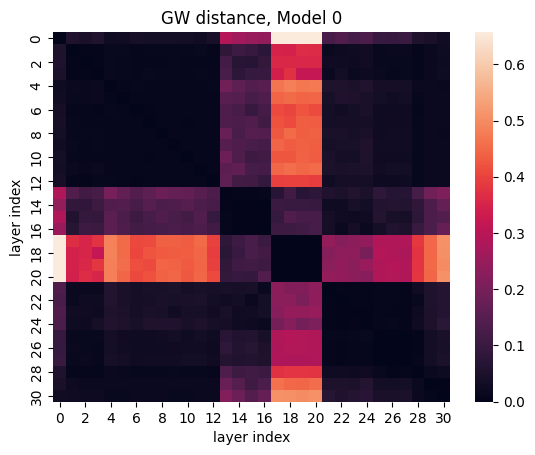}
    \caption{\small{Model 0 Pairwise GW distance, on $f_{\text{mod}}$ dataset.}}
    \label{fig:sidecap}
    \end{subfigure}\hfill
    \caption{Pairwise GW distance on the synthetic Modular Sum dataset.}
     \vskip -6mm
\end{figure*}
We begin by validating the Gromov-Wasserstein distance by comparing it against known partitions of the networks to determine whether it can successfully identify sub-networks.  We first introduce the setup for the experiment, including data generation and models to be investigated.

\noindent\textbf{Setup} As a test case, we focus on a modular sum problem, following existing works \citep{nanda2023progress}. We consider two datasets: the first generated by a single modular sum function with  $c =f_{\text{mod}}(a+b) =  (a + b) \text{mod p}$, where $a, b, c = 0, 1, \dots, p-1$, with $p=59$.  The second dataset is more complex, with $c = f_{\text{mod3}} (a, b)$ of three levels of modular sums, namely: $ c_1 = (a + b) \text{mod } p_1, c_2 = (c_1 + b) \text{mod } p_2, c = (c_2 + b) \text{mod } p_3$, where $p = [59, 31, 17]$. 

\noindent\textbf{Training procedure} We train 3 different neural networks with transformer blocks to predict $c$ given $(a,b)$. These networks contain input embeddings for $a$ and $b$, each of size $d$, i.e., $[\mE_a, \mE_b] \in \sR^{2d}$, and predict a categorical output $c$ via an unembedding/decoding layer. All parameters in the network are learned. For the first simpler $f_{\text{mod}}$ dataset, we train a neural network consisting of a one-block ReLU transformer \citep{vaswani2017attention}, following the same protocol and hyperparameter choices as previous works \citep{nanda2023progress,zhong2024clock}. We call this \textbf{Model 0}. For the more complex $f_{\text{mod3}}$ dataset, we train two neural networks consisting of three-block ReLU transformers, with 3 transformer blocks corresponding to the three levels of modular sum functions, and 4 attention heads within each block. The first network, which we call \textbf{Model E}, employs an end-to-end training procedure to directly learn output $c$ given input $(a, b)$.  For the second network, which we call \textbf{Model L}, we use the same architecture as Model E but with a layer-wise training approach instead of end-to-end training. Specifically, we use the following 3-step procedure: 

%\begin{enumerate}
%\item (optional) In order to speed up training, we take the trained embedding layer from the first problem $c=f_{\text{mod}}(a,b)$ as the embedding layer of Model 2, and freeze it during subsequent training. \cyan{The optional word is confusing here, just remove?}
\noindent    1) We train the first transformer block  of Model L to predict $(c_1, b)$ using an additional  linear layer on top, given inputs $(a, b)$. 
\noindent    2) Once block 1 is fully trained, we discard the linear layer, freeze everything before the linear layer, and use its representations of $(c_1, b)$  to train the second block to predict $(c_2,b)$, again incorporating an additional layer on top. % 
\noindent 3) Finally, we repeat the above step by freezing the first and second block and training the last block to predict $c$, using representations of $(c_2,b)$.  
%\end{enumerate}

In all these models, we are able to achieve $100\%$ prediction accuracy on a separate validation dataset. %Note that extra linear layers can also be considered as probes but used in training. 
More details can be found in appendix~\ref{sec:appendix_alg}. We use the GW measure on \textbf{Model L} with layer-wise training to verify if there is consistency between GW distances and known output $c$'s. 
%In the appendix~\ref{sec:appendix_knowntarget}, we discuss the exact choices of probes and validate them that they can find the correct targets. 
To evaluate the capability of handling different dimensions, we directly measure GW distance between the 93 intermediate representation $Y$ (see appendix~\ref{sec:appendix_alg} for search space details) and $c$'s.  To speed up computation of GW distance, we randomly sub-sample $1000$ data from a total of $3600$ samples, reducing time from $2$ min to $5$ seconds for each computation.
\begin{table}[h]
\small
 \vskip -3mm
  \caption{Gromov-Wasserstein Distance Results for Various Targets in Model L, for $f_{\text{mod3}}$ dataset.}
  \centering
 \begin{tabular}{l|c|l|cl}
    \toprule
 {GW-D for} & Top Similar Layers & $D_{\text{min}}=$\\
   $c_1$   & Resid-Post$^1$ &$0.02$\\
   $c_2$ & Resid-Post$^2$ & $0.03$  \\
     $c$  & Resid-Post$^2$, Resid-Post$^3$, and 7 others & $0.04$  \\
% \midrule
% Model 1&{GW-D for}  & Top Similar Layers& $D_{\text{min}}=$ \\
%  &     $c_1$  &  $z_1^2$ & $0.02$ \\
%  &   $c_2$ &  $z_1^2$ & $0.02$\\
% &     $c$ & $z_1^2$, 6 others, and Resid-Post$^3$  & $0.02$ \\
    \bottomrule
  \end{tabular}
  \label{table:GW_known}
   \vskip -3mm
\end{table}
\noindent\textbf{Results} The results are shown in the Table~\ref{table:GW_known}. We see that in the \textbf{Model L}, the GW distance correctly identifies the most similar layers in accordance with different intermediate $c$'s. The final target $c$ contains 9 similar layers all with distance around $0.04$. In Appendix~\ref{sec:appendix_knowntarget}, we also test probes since the targets are known. Results shows GW distance can be a reliable alternative to the probes to find subnetworks. % even when the target is unknown. 
%We also test it on \textbf{Model 1} where we do not know the exact location of targets $c$'s. GW distance finds the first attention  head from block 2 to be the most similar to all $c$'s. In addition, the correct Resid-Post$^3$ is shown to be in the top 7 most similar layers out of 93 to the final output $c$, 
%We also tested GW distance between $Y$'s and the embedded value of targets, as Emb($c$) with Emb as the embedding layer of the networks. The top similar layers found are the same. 
% \paragraph{Modular Sum: When Target is Unknown } 
% \begin{figure}
% \vspace{-.1cm}
%     \centering
%     \small
%     \begin{minipage}{0.45\textwidth}
%         \centering
%         \includegraphics[width=\textwidth]{figs/gw_model2.png} % First figure
%         \caption{\small{Model L (layer-wise training) pairwise GW distance, on $f_{\text{mod3}}$ dataset.}}
%         \label{fig:1}
%     \end{minipage}\hfill
%     \begin{minipage}{0.45\textwidth}
%         \centering
%         \includegraphics[width=\textwidth]{figs/gw_model1.png} % Second figure
%         \caption{\small{Model E (end-to-end training) pairwise GW distance, on $f_{\text{mod3}}$ dataset.}}
%           \label{fig:2}
%     \end{minipage}
%     \vspace{-.4cm}
% \end{figure}

Moreover, as previously  mentioned GW distance can naturally compare representations across and within transformer blocks with different dimensions. In Fig~\ref{fig:1} and Fig~\ref{fig:2}, we visualize the pairwise GW distance between layer representations without a target for \textbf{Model L} and \textbf{Model E}. Looking at \textbf{Model L} we see predominantly 3 groupings of layers: i) layers roughly from $20$ to $44$ are similar to each other and to layers $52$ to $72$, ii) layers roughly $12$ to $19$ are similar to each other and layers $45$ to $51$ and iii) the initial and last few layers are mainly similar to themselves. Interestingly, the number of groupings corresponds to the 3 functions trained layer-wise in \textbf{Model L}. %Interestingly, there exist a lot of block structures along the diagonal of the GW matrices, more than the expected 3 blocks. This indicates functionally similar layers within these block structures may share similar distance, but then undergo significant chagnes forming a large distance division. 
We also observe differences in patterns across \textbf{Model L} and \textbf{Model E}, suggesting layer-wise and end-to-end training return different networks. Compared to the fixed layer-wise training, end-to-end training in \textbf{Model E} may learn faster in the earlier layers  and may not have much to learn in later layers, as the function may not be particularly  challenging for it. This could explain why, starting from layer 64, all layers in \textbf{Model E} exhibit similar representations.
%This is possibly the reason that from layer $64$ all the layers are similar for \textbf{Model E}. 
Moreover,  magnitudes of the distances are also different, with \textbf{Model L} showing larger distances, indicating that learning the targets $c_1, c_2$ result in more functional differences. One possible  explanation could be that \textbf{Model E} directly operates in the trigonometry space \citep{nanda2023progress}, without having to predict the exact integer values until later, thereby suppressing the distances. We include results from baseline methods capable of handling different dimensions between subspaces in Appendix~\ref{appendix:sum}.
%  in task model L between layer 2044 and layer 52 72: The similarity across transformer blocks arises from the residual connections between the input and outputs from each block. The GW measure is invariant to translation, and hence such linear operation result in a low GW distance, which is expected. 
% \begin{figure}
% \vspace{-.5cm}
%     \centering
%       \includegraphics[width=0.35\textwidth]{figs/gw_model0.png}
%     \caption{\small{Model 0 Pairwise GW distance, on $f_{\text{mod}}$ dataset.}}
%     \label{fig:sidecap}
%     \vspace{-.25cm}
% \end{figure}
% \begin{wrapfigure}{r}{0.35\textwidth}
% \vspace{-.5cm}
%     \centering
%     \includegraphics[width=0.35\textwidth]{figs/gw_model0.png}
%     \caption{\small{Model 0 Pairwise GW distance, on $f_{\text{mod}}$ dataset.}}
%     \label{fig:sidecap}
%     \vspace{-.25cm}
% \end{wrapfigure}
To gain a deeper understanding of the operations within each transformer block, we  visualize pairwise GW distances among layers for \textbf{Model 0} for dataset $f_{\text{mod}}$ in Figure \ref{fig:sidecap}. In this case, we have a total of $31$ representations since only one transformer block is used. We notice the first major difference occurs between layers $13$ and $16$, which are $4$ Attn-Pre (computing key and value product). The second difference occurs between layers $17$ and $20$, which are the first $3$ Attn (computing $\mA(\mX)$). This suggests  that major computation seems to be done by the attention mechanism. Note that distances are not monotonically increasing across layers, which is expected as the representation spaces can change significantly given the heterogeneity of the operations such as those performed by residual connections and attention within a transformer block.   

\subsection{NLP Tasks} 
\label{exp:yelp}
\noindent\textbf{Setup} We now apply GW distance to real natural language processing tasks. We experiment on benchmark sentiment analysis datasets, Yelp reviews %\footnote{ \scriptsize{\url{https://www.kaggle.com/code/suzanaiacob/sentiment-analysis-of-the-yelp-reviews-data}}} 
and Stanford Sentiment Treebank-v2 (SST2) from the GLUE NLP benchmark \citep{glue}, with the goal to predict of the text has positive or negative sentiment, and analyze how different layers from fine-tuning BERT(-base) \citep{bert} models perform on these datasets. We use the pretrained BERT to generate 4 fine-tuned models, corresponding to a dense model and 3 sparse models with sparsity levels of $25\%$, $70\%$ and $95\%$ using a state-of-the-art structured pruning algorithm \citep{np}. Sparsity are used to force models to condense  information into the limited remaining weights, enabling us to examine potential links between this constraint and their structural similarity. Training details are in Appendix~\ref{sec:appendix_nlp}. Due to the size of BERT models, we limit our analysis to comparing the final representations from each of the 12 transformer blocks,  rather than examining all intermediate representations. % within the blocks. % This study highlights how layers differ across different blocks. 

\begin{figure*}[h]
     \begin{subfigure}{.20\linewidth}
        \includegraphics[width=\linewidth]{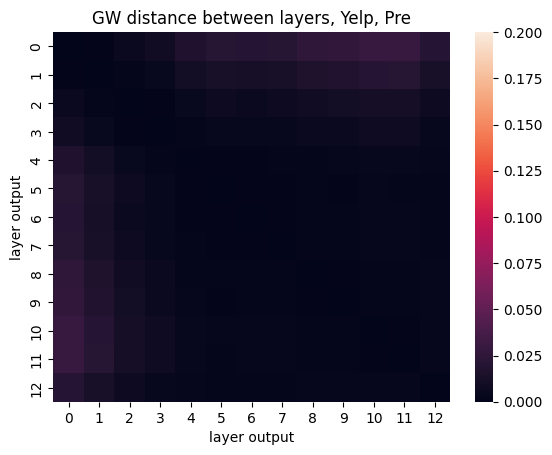}
        \caption{Pre-trained}
        \label{fig:yelp0}
    \end{subfigure}\hfill
    \begin{subfigure}{.20\linewidth}
        \includegraphics[width=\linewidth]{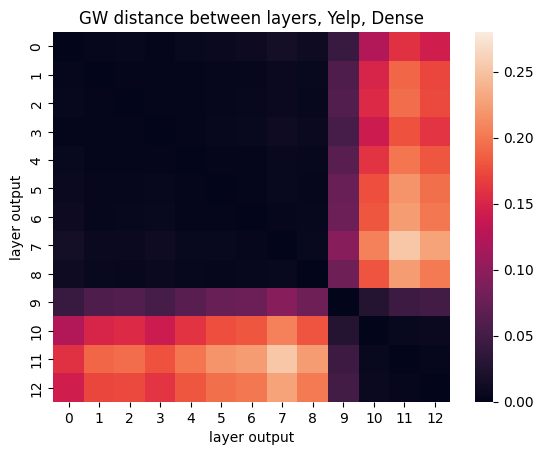}
        \caption{Dense}
        \label{fig:yelp1}
    \end{subfigure}\hfill
    \begin{subfigure}{.20\linewidth}
        \includegraphics[width=\linewidth]{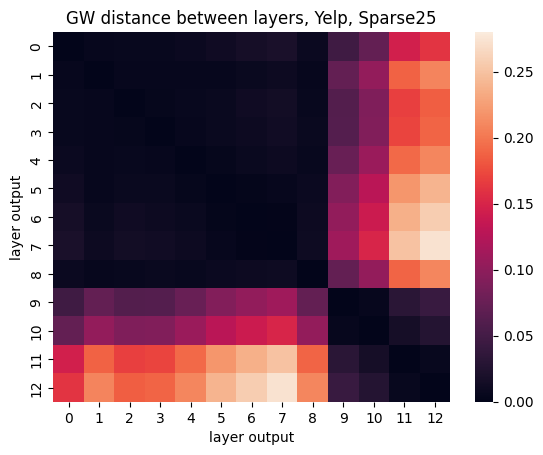}
        \caption{25\% Sparse}
        \label{fig:yelp2}
    \end{subfigure}\hfill
    \begin{subfigure}{.20\linewidth}
        \includegraphics[width=\linewidth]{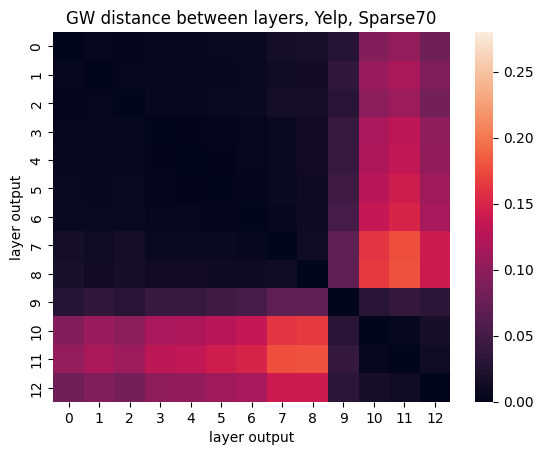}
        \caption{70\% Sparse}
        \label{fig:yelp3}
    \end{subfigure}\hfill
    \begin{subfigure}{.20\linewidth}
        \includegraphics[width=\linewidth]{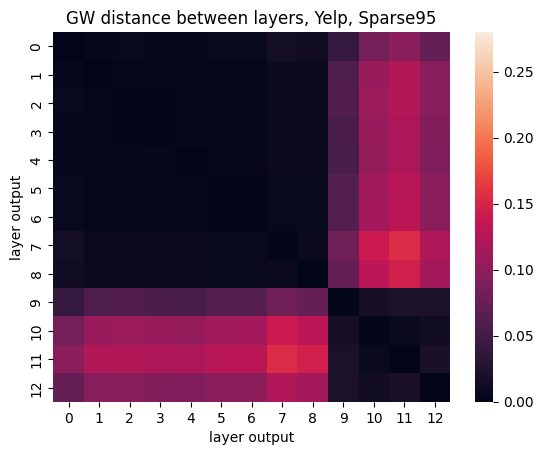}
        \caption{95\% Sparse}
        \label{fig:yelp4}
    \end{subfigure}
    \caption{\small{Pairwise (layer) distances on Yelp, across different BERT models, using the proposed GW distance}, from top to bottom. Different columns: first column is the pre-trained BERT and the rest are fine tuned BERT models with increasing sparsity (dense, $25\%$, $70\%$ and $95\%$ sparsity). As can be seen GW clearly demarcates the (functional) sub-network blocks. Due to page limit, we show baseline results in Appendix~\ref{appendix:yelp_morebaselines}.}
    \label{fig:yelp_w}
\end{figure*}

\noindent\textbf{Results}  In the last row of Figure~\ref{fig:yelp0}, we see that the pre-trained BERT does not have major differences among blocks, which is not surprising given its accuracy on YELP is only 49.3\% (roughly equivalent to random guessing). In Figures~\ref{fig:yelp1} to \ref{fig:yelp4},  we see an interesting pattern emerge,  revealing  two-to-three major block structures in the (sparse) fine-tuned BERT models identified by our approach. The first major differences occur at block $9$ and then the last three blocks ($10$, $11$, $12$) seem to form a distinct block. This seems to indicate that most of the function/task fitting occurs at these later blocks. % This may be explained by the fact that gradient values are potentially larger at these later layers so they change faster from the pretrained models.

\noindent\textbf{Model Compression and Fine-Tuning} The presence of block structures in the GW-distance matrices indicates  major functional changes may concentrate at these transition blocks. This finding may suggest that for other downstream tasks, we may consider freezing the model up to Block 8 and only fine-tuning the blocks after that. %using  the representation from layer $8$ and only fine-tuning the later layers.
We validate this observation in appendix~\ref{sec:appendix_finetune}. We also consider a model compression application where only 4 transformer blocks can achieve similarly good performance, as discussed in Appendix~\ref{appendix:pruning}.

When sparsifying these models, we observe the more sparse models have lesser differences among the blocks (with 95\% sparse model in \ref{fig:yelp4} having the least differences). This is expected as fewer parameters contribute to the final function output besides others being suppressed. Nonetheless, a similar pattern persists, indicating that later blocks differ significantly from earlier ones. This observation is consistent with fine-tuning and sparsification literature \citep{Li2021_DSP,np}, where it has been observed that later blocks typically undergo substantial changes during fine-tuning as they focus on task-specific solutions, while the earlier middle blocks remain stable as they capture syntactic and semantic patterns of the language necessary  for various tasks. In appendix~\ref{appendix:cross}, we further investigate the GW distance between blocks from different models, providing insights into how representations vary across architectures. In Appendix~\ref{appendix:yelp_morebaselines}, we  include  results from baseline similarity measures. Overall, CKA produces also similar block structures to the proposed GW distance, though with greater variability within block structures. In contrast, other baselines fail to reveal such clear block structures. 

%We also compare the proposed GW distance with Euclidean, Cosine, and Wasserstein distance as baselines in Figure~\ref{fig:yelp_w_2old}, on the same YELP dataset and with the same settings.  Euclidean distance between two layers' outputs, shown in the first row of Figure~\ref{fig:yelp_w_2old}, can be seen as the GW distance with a fixed identity-mapping transportation plan for each sample. This validates the low-valued diagonal elements. Off-diagonal elements show greater variation, and it is less obvious there are two distinct sub-groups within layers. The similar pattern is also observed with Cosine and Wasserstein distances, with similar strong diagonal pattern but more pronounced  block structures than Euclidean distance.

%On SST2 dataset, we also observe very similar patterns with the GW distance and 3 baselines, for which we refer the readers to appendix~\ref{appendix:sst} for detailed results.

%However, the exact sub-components are not as clear as with GW distance. 
On SST2 dataset, we also observe very similar patterns with the GW distance and 3 baselines, for which we refer the readers to appendix~\ref{appendix:sst} for detailed results. In both datasets, low distance measure are consistently  observed  in the diagonal elements, but the overall block structures are not as obvious in the baselines as they are with GW distance, highlighting the effectiveness of the GW distance. 

\noindent\textbf{Clustering} Besides visualization above, one can also utilize clustering methods to automatically identify  the subnetworks from the GW distance. We tested spectral clustering \citep{von2007tutorial} on a similarity matrix computed as the reverse pairwise GW distance matrix.  This method successfully identified 2 groups with block $1\sim8$ and block $9\sim12$.

\subsection{Emergence of Subnetworks During Training}
\label{sec:emergence_training}
\begin{figure}[h]
\vskip -3mm
\begin{subfigure}{.25\linewidth}
        \includegraphics[width=\linewidth]{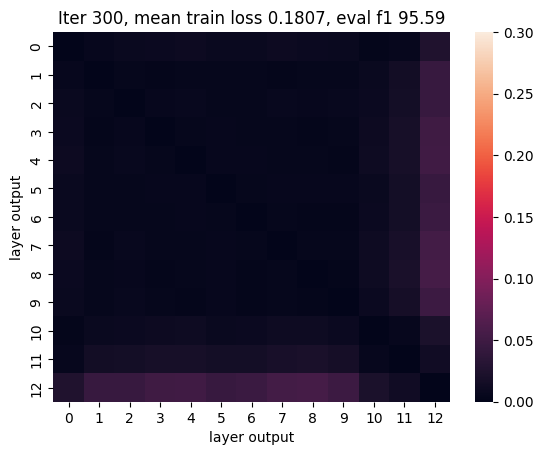}
        \caption{Iteration = $300$}
    \end{subfigure}\hfill
    \begin{subfigure}{.25\linewidth}
        \includegraphics[width=\linewidth]{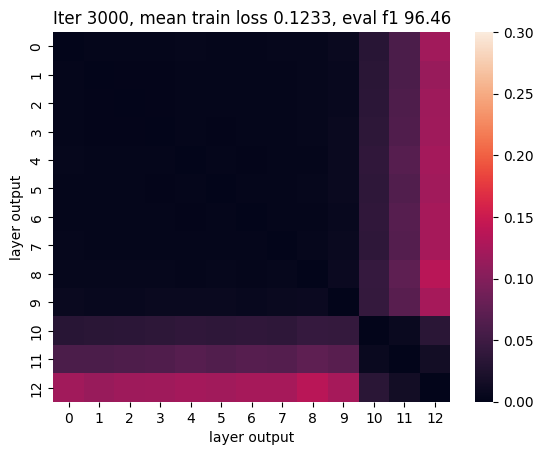}
        \caption{Iteration  = $3$k}
    \end{subfigure}\hfill
    \begin{subfigure}{.25\linewidth}
        \includegraphics[width=\linewidth]{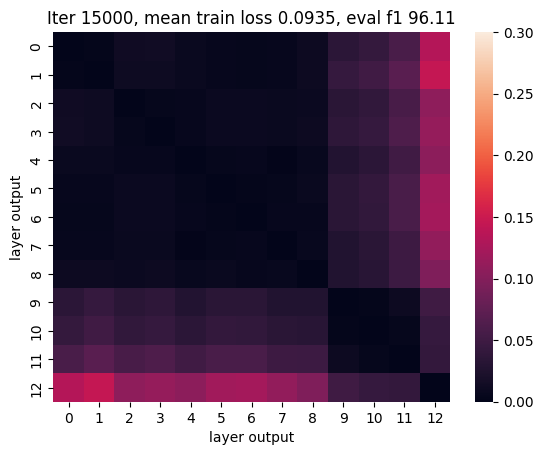}
        \caption{Iteration  = $15$k}
    \end{subfigure}\hfill
        \begin{subfigure}{.25\linewidth}
        \includegraphics[width=\linewidth]{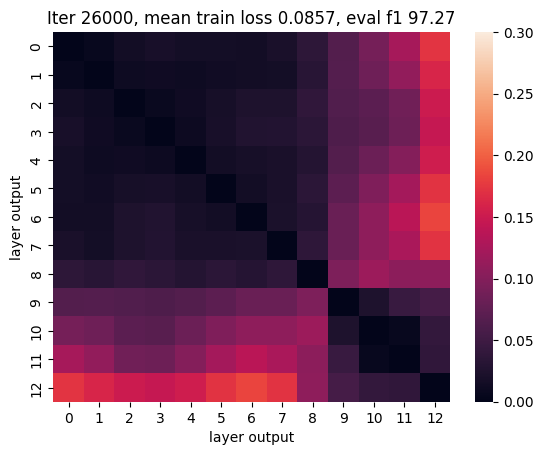}
        \caption{Iteration  = $26$k}
    \end{subfigure}\hfill
    \caption{\small{Pairwise GW distance in YELP datasets, over training iterations.}}
    \label{fig:yelp_train_prog}
    \vskip -3mm
\end{figure}

% In Appendix~\ref{appendix:emergence_training}, we 
We also visualize the GW distance between blocks while fine-tuning the pretrained BERT  model on YELP datasets in the entire training process, in order to observe when these subnetwork structures begin to emerge. Figure~\ref{fig:yelp_train_prog} show a few visualization on GW distances at selected training iterations. Block distances are low in the beginning (observed in Figure~\ref{fig:yelp0}), but by  iteration 300  the last block begin to differ  from other blocks. As training progresses, block 9, 10, and 11 begin to show at iteration 3k and 15k. These growing differences in GW distance  reflect the model’s increasing F1 score on the test data. Overall it show the gradual specialization of blocks into distinct sub-networks, with each sub-network potentially focusing on different aspects of the task.

\begin{figure}[h]
\vskip -3mm
\begin{subfigure}{.33\linewidth}
        \includegraphics[width=\linewidth]{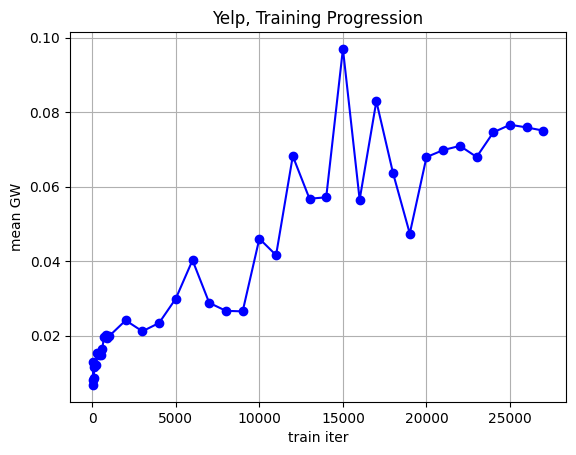}
        %\caption{}
    \end{subfigure}\hfill
    \begin{subfigure}{.33\linewidth}
        \includegraphics[width=\linewidth]{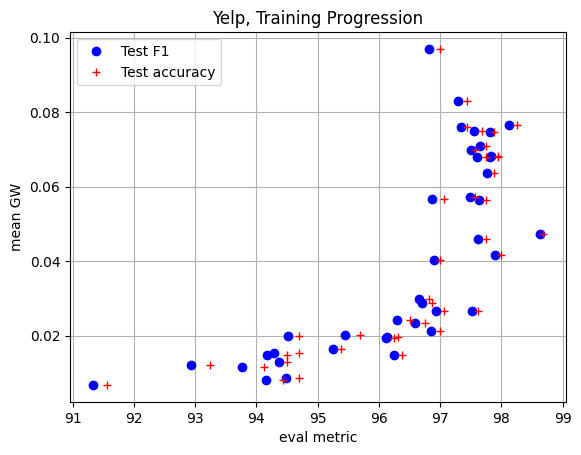}
        %\caption{}
    \end{subfigure}\hfill
\begin{subfigure}{.33\linewidth}
        \includegraphics[width=\linewidth]{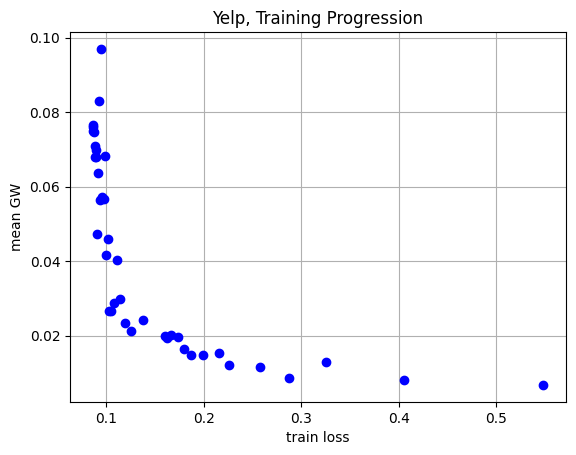}
     %   \caption{}
    \end{subfigure}\hfill
    \caption{\small{Pairwise GW distance in YELP datasets, over training iterations.}}
    \label{fig:yelp_plot_prog}
    \vskip -5mm
\end{figure}

We also plot the mean GW distance of all block pairs  in Figure~\ref{fig:yelp_plot_prog}. Figure~\ref{fig:yelp_plot_prog}a show the mean GW distance over training iterations, and show it grows over time. Figure~\ref{fig:yelp_plot_prog}b shows that mean GW distance versus two different accuracy metric on the test dataset. GW distance grow slowly at first, followed by a rapid increase as the model achieves better accuracy and F1 scores. Such observation is consistent with existing ``grokking" behavior, where validation accuracy can suddenly increases well after achieving near perfect training accuracy \citep{nanda2023progress}. Similarly,  Figure~\ref{fig:yelp_plot_prog}c shows a rapid increase in mean GW distance in order to achieve a lower training loss.
%mean GW distance increases rapidly in order to achieve lower training loss. 

\subsection{ResNet and Computer Vision Dataset}
\begin{figure*}
\vskip -3mm
\centering
\includegraphics[width=0.95\linewidth]{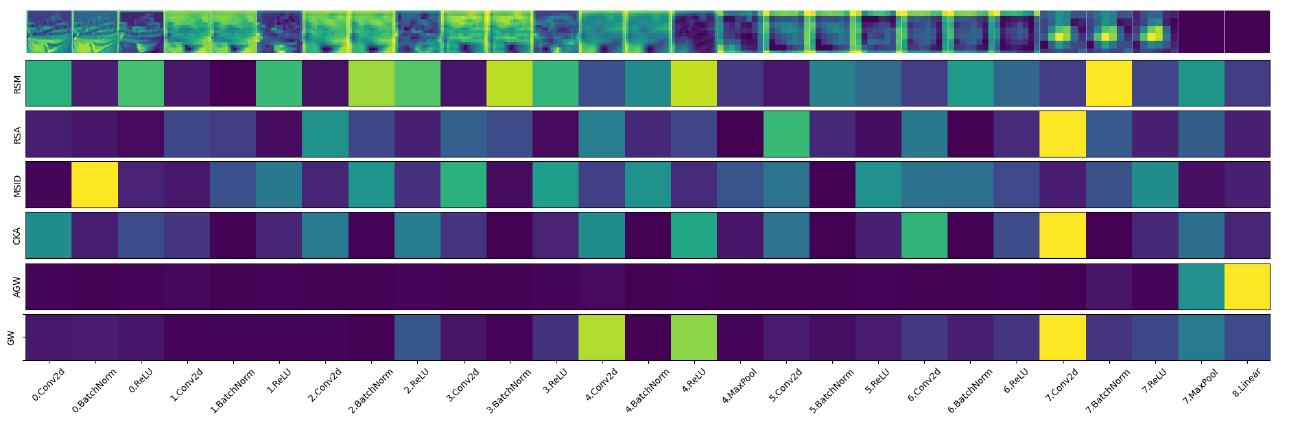}
    \caption{\small{Pairwise layer distance between  every layer and its previous layer on CIFAR-10, for various baselines. Each Row Represents one method.}}
    \label{fig:cifar_visual2}
\end{figure*}
In addition to the attention-based architectures,  we  also test our approach on  ResNet9, 
 a popular convolutional neural network architecture~\citep{he2016deep,park2023trak}. We compare a randomly initialized ResNet9 and a trained one on CIFAR 10 image dataset  CIFAR-10 \citep{krizhevsky2009learning}, achieving 91.63\% accuracy on the test data. For more details on the setup, we refer the readers to Appendix~\ref{appendix:vision}. In Figure~\ref{fig:cifar_w_new}, we show the pairwise distance among layers using baselines that handle different input dimensions. Overall, GW distance show the most clear divisions of subnetworks.

To further examine  how the sub-network structures align with learned representation, we visualize  the computed distances alongside the learned representations of a "ship" image across all layers in Figure~\ref{fig:cifar_visual2}. The top row shows the representations of a ship at each layer.  To see the gradual changes over layers, we visualize the distance between every layer and its previous layer, using various methods capable of handling different dimensions between compared spaces. Overall, RSM, RSA, MSID, and CKA show  indicate significant changes across many layers, without clear evidence of sub-network structures. AGW highlights the changes in the final few layers only. In comparison, GW distance demonstrates the most consistency with the image representations visually. Specifically, the 3rd convolution layer (Layer ID 2.ReLU) introduces the first notable differences, where the ship's shape becomes less distinct, signaling the learning of mid-level features. The shapes become increasingly blurred in the 5th convolution layers (Layer ID 4.Conv2d) and by Layer 4.ReLU  the ship's shape is nearly absent. The final convolutional layer (Layer ID 7.Conv2d) shows significant changes from its preceding layer (Layer ID 6.ReLU), marking the point where class-specific information is consolidated. These results suggest that GW distance aligns most effectively with the learned  representations, providing strong evidence that it  reveals meaningful subnetwork structures.

\vskip -4mm
\section{Discussion}
\vskip -2mm
We proposed a novel approach to model interpretation based on functional similarity within intermediate layers of neural networks, using GW distance to compute such similarities. To the best of our knowledge, our application of GW distance in this context is novel. On algebraic, real NLP, and vision tasks, we identified  the existence of major sub-components amongst layers, corresponding to functionally meaningful abstractions. 
Overall, our method provides an automatic low-cost approach to find sub-components within neural networks, facilitating human understanding. Future work could investigate larger models to observe general trends and applications of functional similarity. Theoretical study of special properties of GW distances within the context of neural network interpretability is also an interesting future direction. 

\paragraph{Limitations and Broader Impact} Our approach first assumes we have access to the intermediate layer representations, which may not be available for some black-box models. Our approach is general, but assumes the proposed distances correctly represent the functional similarities. Our findings are also limited to the datasets and models studied and are not guaranteed to be observed in other scenarios. In terms of broader impact, our approach could be applied widely given its simplicity for identifying functional units in neural networks. However, more investigations on inner mechanisms will have to be done, perhaps building on our approach, in order to fully understand the behavior of neural models.

\begin{ack}
We thank   colleagues and anonymous reviewers for their helpful feedback in many versions of this paper.  
% Use unnumbered first level headings for the acknowledgments. All acknowledgments
% More information about this disclosure can be found at: \url{https://neurips.cc/Conferences/2024/PaperInformation/FundingDisclosure}.
% Do {\bf not} include this section in the anonymized submission, only in the final paper. You can use the \texttt{ack} environment provided in the style file to automatically hide this section in the anonymized submission.
\end{ack}

\bibliographystyle{abbrv}
\bibliography{ref}

%%%%%%%%%%%%%%%%%%%%%%%%%%%%%%%%%%%%%%%%%%%%%%%%%%%%%%%%%%%%
\newpage
\appendix
\section{Representations in Transformer-based and Convolution Neural Network}
\label{apendix:rep}
 We consider multiple candidate $Y$'s to form the search space for target $Y^0$.  In the context of MLP neural networks for example, where $\sigma(.)$ denotes the non-linearity and $W$s are the parameter matrices, we have $Y^* = W_n (\sigma(W_{n-1} \dots \sigma(W_1 X)))$ for the whole network. We can extract many $Y$'s from intermediate functions of the model, for instance $Y_1 = W_1X$, $Y_2 = \sigma(W_1 X)$, and so on. These $Y$'s are often called representations, activations, or sometimes even ``outputs'' from each layer.
 
For attention modules in transformer neural networks \citep{vaswani2017attention}, we can similarly extract $Y$'s from attention key, query, and value functions as well as MLP functions. More specifically, a deep transformer architecture of depth $l$ is formed by sequentially stacking $l$ transformer blocks. Each transformer block takes the representations of a sequence $\mX_{\text{in}} \in \R^{T \times d}$, where $\mX_{\text{in}}$ = Emb($\mX$) with embedding layer Emb and input $\mX$, $T$ is the number of tokens and $d$ is the embedding dimension, and outputs $\mX_{\text{out}}$, where:
% \begin{wrapfigure}{r}{.7\textwidth}
\begin{equation}
\begin{split}
   \mX_{\text{out}} &= \alpha_{\text{FF}} \hat{\mX} + \beta_{\text{FF}} \text{MLP}(\text{Norm}(\hat{\mX})) \\
\text{where,} \quad  \text{MLP}(\mX_m)&= \sigma(\mX_m \mW^1)\mW^2\\
\hat{\mX} &= \alpha_{\text{SA}} \mX_{\text{in}} + \beta_{\text{SA}} \text{MHA}(\text{Norm}(\mX_{\text{in}})),\\
\text{MHA}(\mX) &= [\text{Attn}_1(\mX), \dots, \text{Attn}_H(\mX))]\mW^P, \\
    \text{Attn}(\mX) &= \mA(\mX)\mX \mW^V, \\ 
    \mA(\mX) &= \text{softmax}\left(\frac{1}{\sqrt{d_k}} \mX \mW^Q {\mW^K}^\top \mX^\top + \mM\right),
    \end{split}
\label{eq:transformers}
\end{equation}
%\end{wrapfigure}
with scalar weights $ \alpha_{\text{FF}}$, $\beta_{\text{FF}} $ , $\alpha_{\text{SA}}$, and $\beta_{\text{SA}}$ usually set to $1$ by default. Here $\text{FF}$ stands for feedforward network,  $\text{SA}$ stands for self-attention,  $\text{MHA}$ is Multi-Head Attention, and \text{Norm} is a normalization layer. MLP usually has a single hidden layer with dimension $d$ and ReLU activation. The MHA sub-block shares information among tokens by using self-attention with $\mW^Q,~\mW^K$ and $\mW^V$ indicating query, key and value matrices. We list the exact locations of representations considered in the transformer models in Table~\ref{table:location}.

\begin{table}[h]
\small
  \caption{Representations $Y$ in the attention-based  model considered in experiments as per \eqref{eq:transformers}. Omitting $Y$ in most names for readability.}
%  \label{table:math}
  \centering
  \begin{tabular}{l|llllllllll}
    \toprule
%  \multicolumn{2}{c}{Across Block}   & \multicolumn{2}{c}{Within Block}   \\
   % \cmidrule(r){2-3}
 \multicolumn{5}{l}{(Across Blocks)}    &      & \\

    Name  & Resid-Pre$^l$ &    $Y^l$, at each block \\
    Value    & $=\mX_{\text{in}}^l$ & $=\mX_{\text{out}}^l$   \\
\midrule
 \multicolumn{5}{l}{(Within Each Block $l$)}    &      & \\

    Name   & Attn-Out$^l$ & Resid-Mid$^l$ & Pre & Post & MLP-out$^l$ &  Resid-Post$^l$   \\
    Value  & $=$MHA($\mX$)$^l$ & $=\hat{\mX}$& $=\hat{\mX}\mW^1$  & $=$\text{MLP}($\hat{\mX}$) & $=\text{MLP}(\hat{\mX})$  & $=\mX_{\text{out}}$ \\
    
\midrule
\multicolumn{5}{l}{(Within Each Attention Head $h$)}   &      & \\
  Name & $k_h$ &  $q_h$ & Attn-Pre$_h$ & Attn$_h$ &  v$_h$  & z$_h$   \\
    Value    & $=\mX \mW^K$ & $=\mX \mW^Q$ &$= q_h k_h^T $ & $=\mA(\mX)$ & $=\mX \mW^V$ & $=$Attn($\mX$)    \\
    \bottomrule
  \end{tabular}
\label{table:location}
\end{table}

We also consider convolution neural networks for computer vision datasets. Specifically, we use a relatively lightweight  ResNet9 \citep{he2016deep,park2023trak}. The exact locations of the candidate representations considered are listed in Table~\ref{table:location_resnet}.

 \begin{table}[h]
\vskip -4mm
\small
  \caption{All representations $Y$ considered in ResNet 9 in experiments.}
%  \label{table:math}
  \centering
  \begin{tabular}{l|llllllllll}
    \toprule
%  \multicolumn{2}{c}{Across Block}   & \multicolumn{2}{c}{Within Block}   \\
   % \cmidrule(r){2-3}
 \multicolumn{5}{l}{(Module 0)}    &      & \\

    Name  & 0.Conv2d  & 0.BatchNorm & 0.ReLU &   \\
    Details & in-channel = 3, out =64, kernel size = (3,3)  &  Batch Normalization &  activation\\
\midrule
  \multicolumn{5}{l}{(Module 1)}    &      & \\

    Name  & 1.Conv2d  & 1.BatchNorm & 1.ReLU  &  \\
    Details & in-channel = 64, out =128, kernel size = (5,5)  &  Batch Normalization &  activation\\
\midrule
  \multicolumn{5}{l}{(Module 2 \& 3: Residual Block )}    &      & \\

    Name  & 2.Conv2d  & 2.BatchNorm & 2.ReLU  &  \\
      Name  & 3.Conv2d  & 3.BatchNorm & 3.ReLU  &  \\
    Details & in-channel = 128, out =128, kernel size = (3,3)  &  Batch Normalization &  activation\\
\midrule
 \multicolumn{5}{l}{(Module 4)}    &      & \\
    Name  & 4.Conv2d  & 4.BatchNorm & 4.ReLU  & 4. MaxPool  \\
    Details & in-channel = 128, out =256, kernel size = (3,3)  &  Batch Normalization &  activation & Kernel (2,2) \\
\midrule
  \multicolumn{5}{l}{(Module 5 \& 6: Residual Block )}    &      & \\
      Name  & 5.Conv2d  & 5.BatchNorm & 5.ReLU  &  \\
          Name  & 6.Conv2d  & 6.BatchNorm & 6.ReLU  &  \\
    Details & in-channel = 256, out =256, kernel size = (3,3)  &  Batch Normalization & \\
    \midrule
 \multicolumn{5}{l}{(Module 7)}    &      & \\
    Name  & 7.Conv2d  & 7.BatchNorm & 7.ReLU  &  7. MaxPool &  \\
    Details & in-channel = 256, out =128, kernel size = (3,3)  &  Batch Normalization &  activation & Adaptive \\
\midrule
 \multicolumn{5}{l}{(Module 8)}    &      & \\
    Name  & 8.Linear (classification) &  \\
    Details & in-feature = 128, out =10 &  \\
    \bottomrule
  \end{tabular}
\label{table:location_resnet}
\vskip -3mm
\end{table}

\section{Modular Sum Experiment Details}
\label{sec:appendix_alg}
We use the same architecture and protocols in training, as previous modular papers \citep{nanda2023progress,zhong2024clock}, based on their available Github repos. Specifically, we use transformer width $d=128$, and each attention head has $32$ dimensions. As a result, MLP has $512$ hidden neurons. ReLU is used as the activation throughout the models, 

\paragraph{Data} Among all  data points  ($59^2 = 3481$ of them), we randomly select 80\% as training
 samples and 20\% as validation samples. 

\paragraph{Hyperparameters}  We used AdamW optimizer 
\citep{loshchilov2017decoupled} with learning rate $\gamma = 0.001$ and weight decay factor
 $\beta =2$. We use the shuffled data as one batch  in every epoch. We train models from scratch and train for 26,000 epoches.

\paragraph{Search Space} For the $f_{\text{mod3}}$ dataset, we consider all layers  in the network, including all representations within  transformer blocks. As shown in Table~\ref{table:location}, each attention head has 6 intermediate layers, for a total of 24. Each block has an additional 7 layers (1 input layer, Resid-Pre, and 6 intermediate layers). Hence,  for three blocks each with four attention heads, we have a total of $93$ representations to evaluate, as each block has $31 = 24 + 7$ representations.

\section{Probes on Modular Sum Dataset: When Target is Known}
\label{sec:appendix_knowntarget}

When  the  target is a  value from a known function,  we can directly compare outputs between representations from each  layer and the known function output.   Representations from each layer can be directly compared with the target via a probe. We  first consider \textbf{Model E} and then \textbf{Model L}.

% \paragraph{Search Space} For the $f_{\text{mod3}}$ dataset, we consider all layers  in the network, including all representations within  transformer blocks. As shown in Table~\ref{table:location}, each attention head has 6 intermediate layers, for a total of 24. Each block has an additional 7 layers (1 input layer, Resid-Pre, and 6 intermediate layers). Hence,  for three blocks each with four attention heads, we have a total of $93$ representations to evaluate, as each block has $31 = 24 + 7$ representations.

% \blue{The above description is confusing}

\paragraph{Linear Probe}   Popular linear probes can be used to assess the similarity between a target and any layer's representation.  We perform linear regression of each target $(c_1, c_2, c)$ on each of the 93 representations $Y$, and report the residual error as the scoring distance function between $Y$ and $c$'s. 

\begin{table}[h]
\small
  \caption{Linear and Nonlinear Probe Results, for $f_{\text{mod3}}$ dataset.}
  %\label{sample-probe-table}
  \centering
 \begin{tabular}{l|c|c|l|cl}
    \toprule
Model L & {Linear Probe for} & Perfect Match? & Top Similar Layers & $D_{\text{min}}=$\\

 &  $c_1$  & $\checkmark$ & Resid-Post$^1$ and 21 others &$0$\\
 &   $c_2$ & $\checkmark$ & Resid-Post$^2$ and 21 others& $0$  \\
 &    $c$ & $\checkmark$ & Resid-Post$^3$ and Post$^2$& $0$  \\
\midrule
Model E &{Linear Probe for} & Perfect Match?  & Top Similar Layers& $D_{\text{min}}=$ \\
 &     $c_1$ & $\times$ &  Post$^2$& $0.522$ \\
 &   $c_2$ & $\times$ & Post$^1$& $0.93$\\
&     $c$ & $\checkmark$ & Resid-Post$^3$ and 5 others& $0$ \\
\midrule
Model E &{Nonlinear Probe for} & Perfect Match?  & Top Similar Layers& $D_{\text{min}}=$ \\
 &     $c_1$ & $\checkmark$ &Resid-Post$^1$ and 15 others & $0$ \\
 &   $c_2$ & $\checkmark$ & Resid-Post$^1$ and 4 others & $0$\\
&     $c$ & $\checkmark$ & Resid-Post$^3$ and 9 others& $0$ \\
    \bottomrule
  \end{tabular}
  \label{table:known}
\end{table}

\paragraph{Results} Since we perform layer-wise training with \textbf{Model L}, we know the true locations of $c_1$ and $c_2$, which sit at $\mX_{\text{out}}^1$ and $\mX_{\text{out}}^2$ with names Resid-Post$^1$ and Resid-Post$^2$, respectively. As shown in the top part of Table~\ref{table:known}, a linear regression probe can predict targets perfectly with these two layers. In fact, there are 21 other layers which also show perfect accuracy. For $c_1$, these consist of Post$^0$ and MLP-out$^0$ from the same  block and some layers from the next block, including linear operations with all $k$'s, $q$'s, $v$'s. The final prediction $c$ can be linearly predicted as expected, due to the model's perfect prediction accuracy.  

Naturally we would like to confirm if the same happens with \textbf{Model E}: if we use the same linear probe, does each block in \textbf{Model E} learn the corresponding $c$ at the output of the transformer block? As shown in the mid part of Table~\ref{table:known}, we are not able to find any layer that produces a representation that is linearly predictive of $c_1$ and $c_2$, with the lowest prediction errors at $52\%$ and $93\%$, respectively. Moreover, the most similar layers to $c_1$ and $c_2$ are in the 2nd block and 1st block respectively, instead of the expected 1st and 2nd blocks. This seems to suggest that  \textbf{Model E} does not actually learn any function of  $c_1$ and $c_2$. 

\paragraph{Non-linear Probe} As discussed previously, to deal with the potentially large search space of functions of the target, a more powerful probe (such as a nonlinear MLP function) may have to be used so that it can detect more complex similarities to $c$. Therefore, we train a two-layer MLP\footnote{We use the neural network classifier from the scikit-learn package, with default parameters.} to predict $c$'s. As shown at the bottom of Table~\ref{table:known}, these two-layer MLPs have more predictive power and can perfectly predict the targets, while still showing differences among various layers indicating that the matched layers do capture the intended target functions while other layers do not. Many layers in the 3rd block, for example, have only $1\%$ accuracy relative to $c_1$. This indicates that non-linear probes can be used to find subgroups of layers  in neural networks. Unlike existing work that primarily focuses on linear probes, we show that non-linear probes, still with limited capacity, are useful. 

% \blue{include raw values in appendix? maybe in a figure}
%usefulness on scale and complexity: multiple-levels, explanation in term of audience understanding and complexity, pattern emerging at ensemble/high level,

One issue with using predictive probes to compute the distance measure $D$ is that the target function has to be known. In  practice, however, we may not know any intermediate targets, as suggested in the end-to-end training of \textbf{Model E}. While we still can try different target functions and use non-linear probes, the infinite number of possible targets makes such an approach inefficient. This calls for a different strategy to differentiate sub-components in a network through representation similarity.

\section{Baseline Comparison Results on Modular Sum}
\label{appendix:sum}
\begin{figure}[h]
   % \caption{\small{Pairwise GW distance on Yelp, across BERT models. (a) is pre-trained BERT and (b)-(e) are fine tuned BERT models with increasing sparsity (dense, $25\%$, $70\%$ and $95\%$ sparsity).}}
   % \label{fig:yelp}
% \end{figure}
% \begin{figure}[h]
 \vskip -3mm
  \begin{subfigure}{.30\linewidth}
        \includegraphics[width=\linewidth]{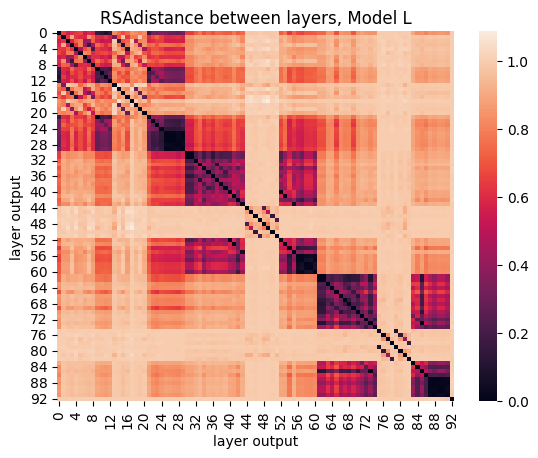}
        %\caption{}
        %\label{fig:yelp0}
    \end{subfigure}\hfill
  \begin{subfigure}{.30\linewidth}
        \includegraphics[width=\linewidth]{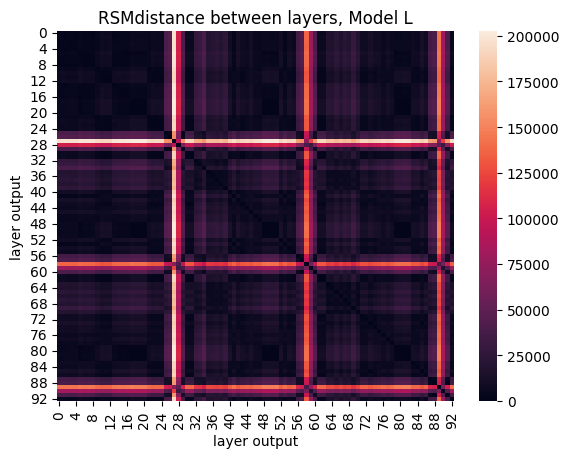}
        %\caption{}
        %\label{fig:yelp0}
    \end{subfigure}\hfill
    \begin{subfigure}{.30\linewidth}
        \includegraphics[width=\linewidth]{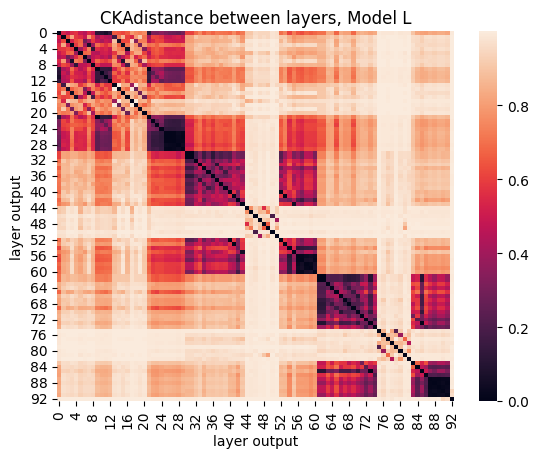}
        %\caption{}
         %\label{fig:yelp1}
    \end{subfigure}\hfill
    \newline
  \begin{subfigure}{.30\linewidth}
        \includegraphics[width=\linewidth]{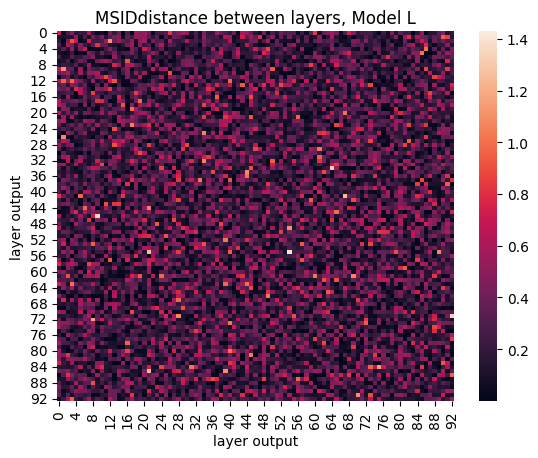}
        %\caption{}
        %\label{fig:yelp0}
    \end{subfigure}\hfill
    \begin{subfigure}{.30\linewidth}
        \includegraphics[width=\linewidth]{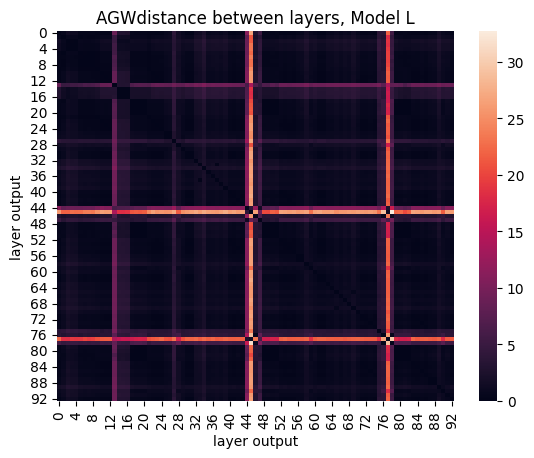}
        %\caption{}
         %\label{fig:yelp1}
    \end{subfigure}\hfill
    \begin{subfigure}{.30\linewidth}
        \includegraphics[width=\linewidth]{figs/gw_model2.png}
        %\caption{}
         %\label{fig:yelp4}
    \end{subfigure}
    \caption{\small{Pairwise (layer) distances on Modular Sum dataset, with layer-wise trained models. Different figures from  left to right, top to bottom:  \textit{RSA, RSM, CKA, MSID, AGW, and the proposed GW distance}.} }
    \label{fig:mod_w_new}
    \vskip -3mm
\end{figure}

We have also tested  a few baselines that can handle different space dimensions, shown in Figure~\ref{fig:mod_w_new}. RSA and CKA reveal different levels of subnetworks within  attention layers and across transformer blocks. AGW demonstrates the highest sensitivity to attention computations, while RSM finds the last few layers within each transformer block. 

\section{Real NLP Experiment Details}
\label{sec:appendix_nlp}

We analyze a BERT-base-uncased \citep{bert} model based on our optimal matching inspired mechanistic interpretability approach. We fine tune it on two well known datasets in NLP; i) Yelp reviews (\url{https://www.kaggle.com/code/suzanaiacob/sentiment-analysis-of-the-yelp-reviews-data}) and ii) Stanford Sentiment Treebank-v2 (SST2), which is part of the GLUE NLP benchmark \citep{glue}. Both of these are sentiment analysis tasks, where the goal is to predict if a piece of text has positive or negative sentiment. The Yelp dataset has hundreds of thousands of reviews, while the SST2 dataset has tens of thousands of sentences. The training details are as follows: i) Hardware: $1$ A100 Nvidia GPU and $1$ intel CPU, ii) Max. Sequence Length : $256$, iii) Epochs: $1$, iv) Batch Size: $16$ and v) Learning Rate: $2e^{-5}$ with no weight decay. The accuracy on Yelp was $97.87\%$, while that on SST2 was $92.4\%$. Without fine tuning the pre-trained BERT models accuracy on Yelp and SST2 was $49.29\%$ and $50.34\%$ respectively indicative of random chance performance.

We also fine tuned a series of sparse models on these datasets. The method we used to sparsify was a state-of-the-art dynamic sparse training approach NeuroPrune \citep{np}, which leads to high performing structured sparse models. Using this approach and the same training settings as above we created BERT models with $25\%$, $70\%$ and $95\%$ sparsity which had accuracies of $96.31\%$, $97.53\%$ and $96.22\%$ respectively for the Yelp dataset and accuracies of $90.25\%$, $88.5\%$ and $84.4\%$ respectively for the SST2 dataset. We then used the resultant models for our analysis.

\section{GW Justification and Alignment }
\label{sec:appendix_align}
% \begin{wrapfigure}{r}{0.2\textwidth}
%     %  \begin{subfigure}{.5\linewidth}
%     %     \includegraphics[width=\linewidth]{figs/YELP_bert-base-uncased_target_unknown_spectral_GW_pre_scores_histo.png}
%     %     \caption{Pretrained}
%     %     \label{fig:yelp0_hist}
%     % \end{subfigure}\hfill
%     % \begin{subfigure}{.5\linewidth}
%         \includegraphics[width=\linewidth]{figs/YELP_bert-base-uncased_target_unknown_spectral_GW_scores_histo.png}
%         % \caption{Fine-Tuned}
%         \label{fig:yelp1_hist}
%     % \end{subfigure}\hfill
%     % \begin{subfigure}{.33\linewidth}
%     %     \includegraphics[width=\linewidth]{figs/YELP_bert-base-uncased_target_unknown_spectral_GW_scores_tsne.png}
%     %     \caption{tSNE projection}
%     %     \label{fig:yelp2_hist}
%     % \end{subfigure}\hfill
%     % \begin{subfigure}{.20\linewidth}
%     %     \includegraphics[width=\linewidth]{figs/YELP_bert-base-uncased-sparse70_target_unknown_spectral_GW_scores_histo.png}
%     %     \caption{}
%     %     \label{fig:yelp3_hist}
%     % \end{subfigure}\hfill
%     % \begin{subfigure}{.20\linewidth}
%     %     \includegraphics[width=\linewidth]{figs/YELP_bert-base-uncased-sparse95_target_unknown_spectral_GW_scores_histo.png}
%     %     \caption{}
%     %     \label{fig:yelp4_hist}
%     % \end{subfigure}
%     \caption{\small{Histogram on pairwise distances for outputs from all transformer blocks in a fine-tuned BERT model trained on YELP dataset.}} % $c)$ shows the neighborhood of tSNE projection for a batch of data across layers.}}
%     \label{fig:yelp_hist}
% \end{wrapfigure}
\noindent\textbf{Distance Distributions.} As an illustrative example, we plot the histogram on pairwise distances for a batch of samples across all transformer blocks in BERT models from the YELP review dataset in Figure~\ref{fig:yelp_hist}.  The results in Figure~\ref{fig:yelp_hist} show the distributions on pairwise distances begin to differ from block 9, consistent with GW distance observed in Figure~\ref{fig:yelp_w},  suggesting that  significant transformations occur and can be effectively captured by GW. 

\begin{figure}[h]
     \begin{subfigure}{.20\linewidth}
        \includegraphics[width=\linewidth]{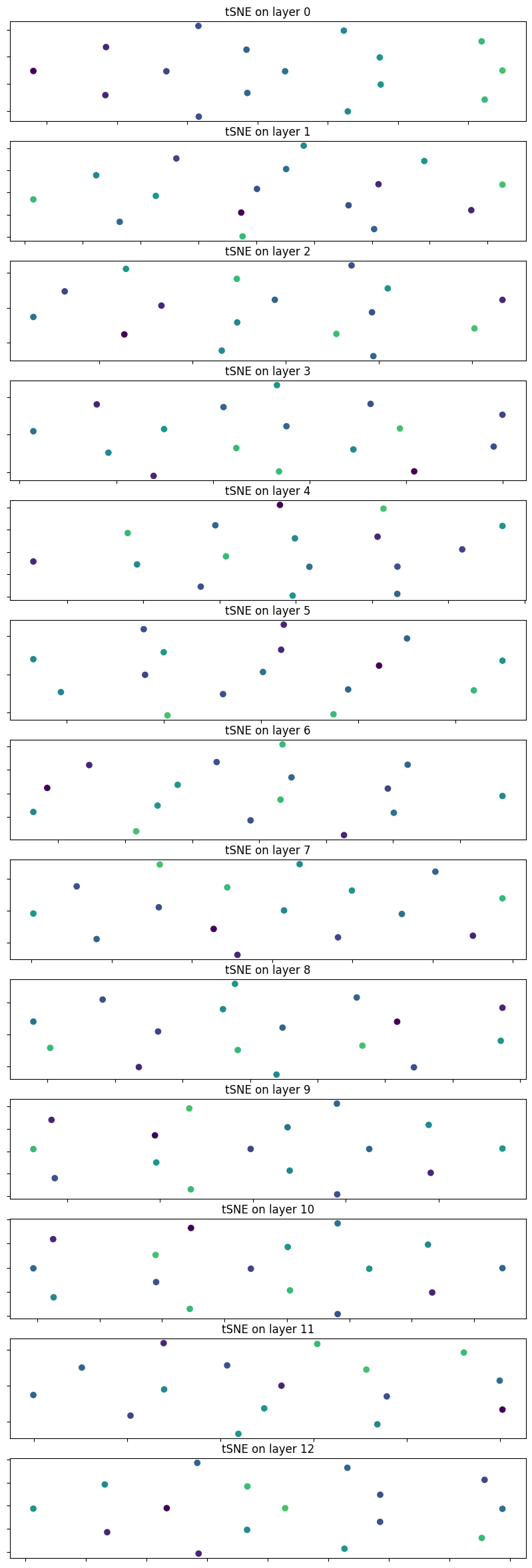}
        \caption{Pretrained}
        \label{fig:yelp0_tsne}
    \end{subfigure}\hfill
    \begin{subfigure}{.20\linewidth}
        \includegraphics[width=\linewidth]{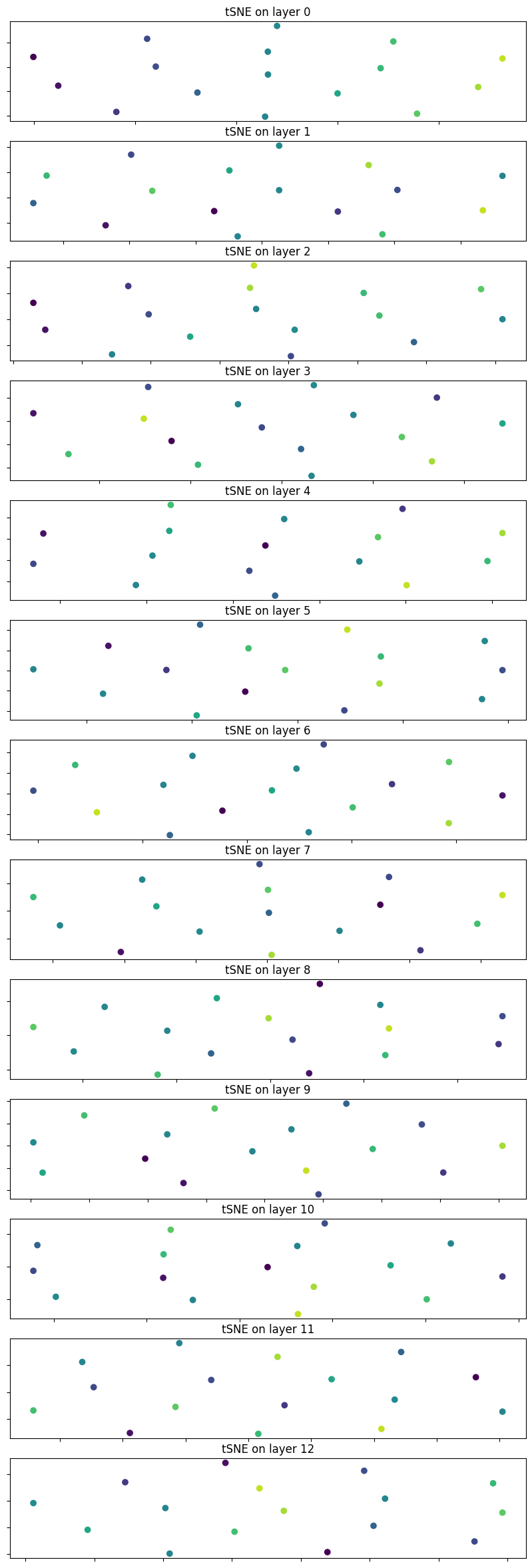}
        \caption{Dense}
        \label{fig:yelp1_tsne}
    \end{subfigure}\hfill
    \begin{subfigure}{.20\linewidth}
        \includegraphics[width=\linewidth]{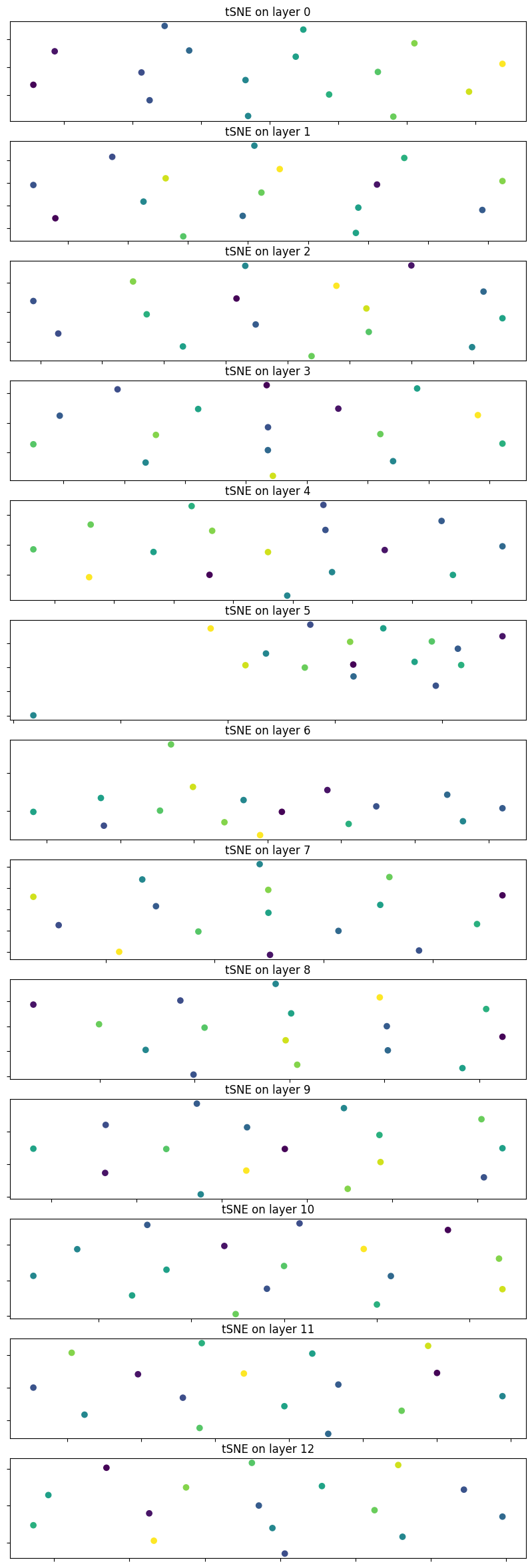}
        \caption{25\% Sparse}
        \label{fig:yelp2_tsne}
    \end{subfigure}\hfill
    \begin{subfigure}{.20\linewidth}
        \includegraphics[width=\linewidth]{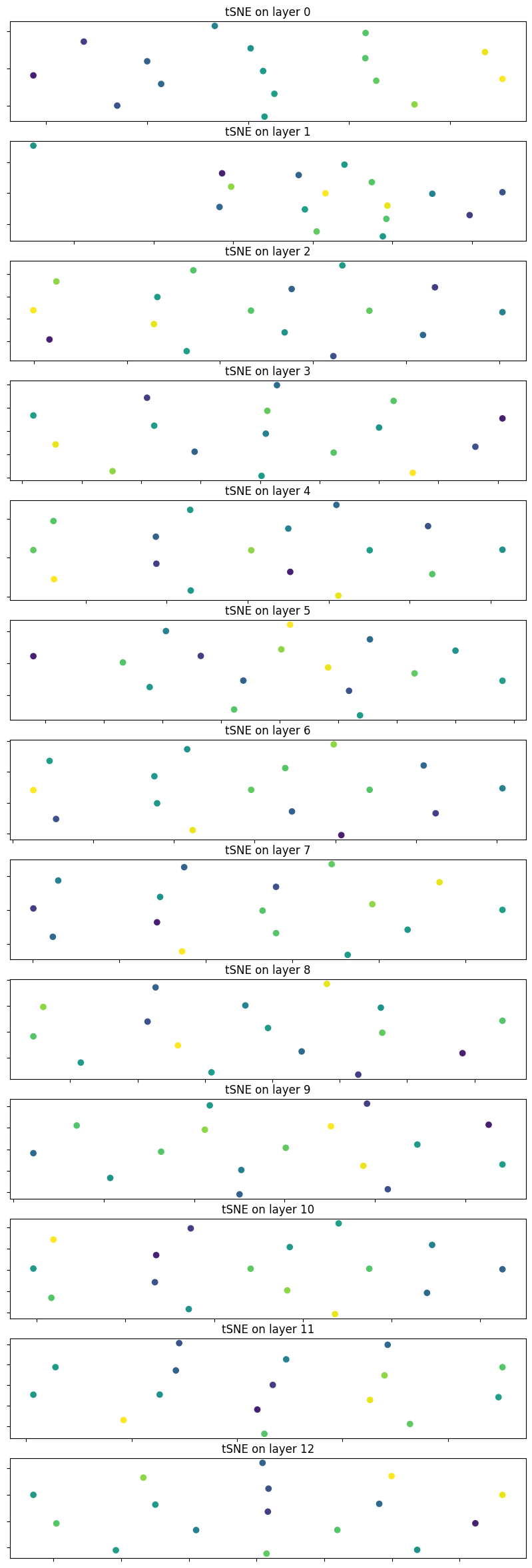}
        \caption{70\% Sparse}
        \label{fig:yelp3_tsne}
    \end{subfigure}\hfill
    \begin{subfigure}{.20\linewidth}
        \includegraphics[width=\linewidth]{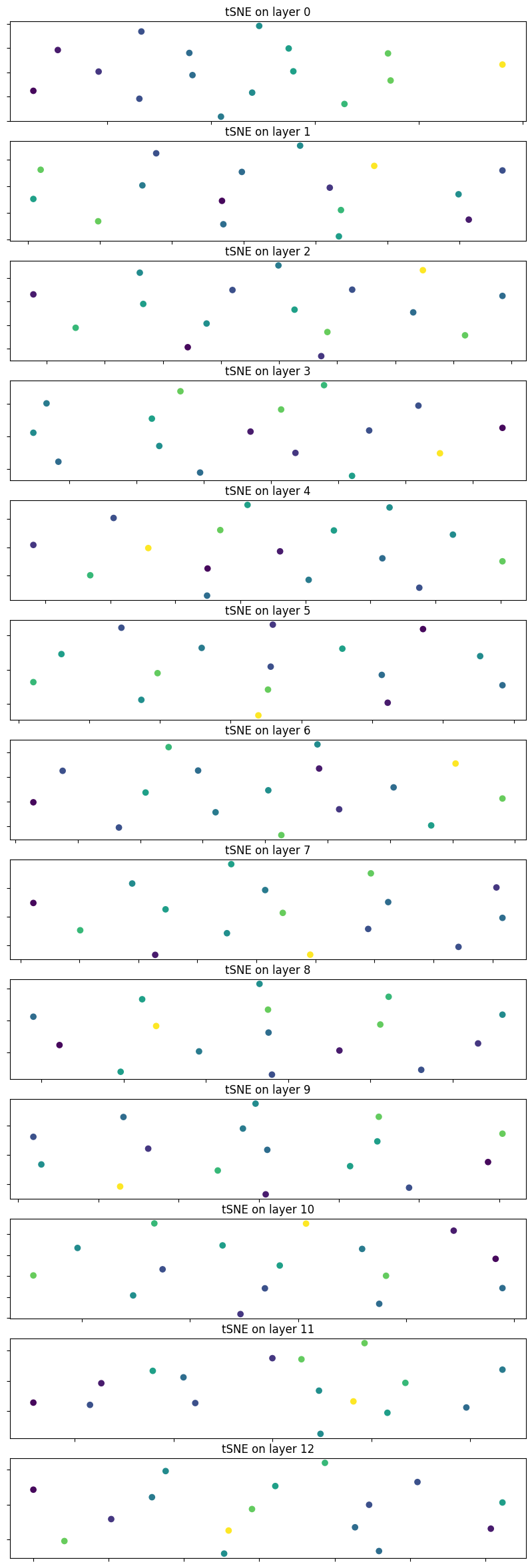}
        \caption{90\% Sparse}
        \label{fig:yelp4_tsne}
    \end{subfigure}
    \caption{\small{tSNE projection on intermediate representations on Yelp, across  BERT models with different sparsity levels. Different Rows: Results from all 12 transformer blocks, from top to bottom. Different columns: first column is the pre-trained BERT and the rest are fine tuned BERT models with increasing sparsity (dense, $25\%$, $70\%$ and $95\%$ sparsity).}}
    \label{fig:yelp_tsne}
\end{figure}

\noindent\textbf{Neighborhood Change.} Complementary to the distribution of  pairwise distances, the changing representations of samples could also alter their relative neighborhoods across transformer blocks. We plot a tSNE projection \citep{van2008visualizing} of representations from  a batch of samples on YELP, and visualize it in Figure~\ref{fig:yelp_hist}b and Figure~\ref{fig:yelp4_tsne}. The Jaccard similarity, measuring the overlap between top-5-neighbors of 3 selected samples across different transformer blocks, ranges from 0.0 to 0.43, with average values of $\{0.27, 0.26, 0.26\}$. The full details are shown in Table~\ref{table:jaccard_tSNE}, as discussed below. Hence, the sample neighborhood changes across blocks, which can be indicative of functional changes that are not captured by comparing distributions alone. However, GW can account for such changes as well. 

We plot a tSNE projection \citep{van2008visualizing} down to 2 dimensions, on a batch of 16 samples (color indicative of sample) on YELP, and visualize it in Figure~\ref{fig:yelp4_tsne}. As one can see, the sample neighborhood changes across layers, which can be indicative of functional changes but something that is not captured by comparing distributions. However, GW can also account for such changes. 

We also show Jaccard similarity measure on top-5-neighbors, per Euclidean distances on tSNE projection, of each of 3 samples across different transformer blocks. Jaccard similarity is a measure of two sets, computed as their intersection divided by their union.  Results are shown in Table~\ref{table:jaccard_tSNE}. This further shows the sample neighborhood changes across layers, and representation similarity measures should account for such changes. 

%\red{check layer vs block usage to avoid confusion}

\begin{table}[t]
\tiny
  \caption{Jaccard Similarity on top-5-neighbors of Selected Samples across all transformer blocks.}
  %\label{sample-probe-table}
  \centering
 \begin{tabular}{l|c|c|c|c|c|c|c|c|c|c|c|c|c|c|c|c|c|c|c}
    \toprule
Sample 1 &  1 &2&3&4&5&6&7&8&9&10&11&12 & Mean\\
Block 0 v.s. &  0.25  & 0.25 & 0.25& 0.11&0.43& 0.11&0.25&0.25&0.11&0.11&0.25&0.25 &\bf{0.27}\\
\midrule
Sample 2 &  1 &2&3&4&5&6&7&8&9&10&11&12& Mean\\
Block 0 v.s. &  0.11  & 043 & 0.11& 0.11&0.11& 0.0&0.11&0.25&0.25&0.43&0.25&0.25 &\bf{0.26}\\
\midrule
Sample 3 &  1 &2&3&4&5&6&7&8&9&10&11&12& Mean\\
Block 0 v.s. &  0.0  & 0.11 & 0.43& 0.25&0.25& 0.25&0.11&0.66&0.11&0.11&0.11&0.0 &\bf{0.26}\\
    \bottomrule
  \end{tabular}
  \label{table:jaccard_tSNE}
\end{table}

\begin{figure}[h]
  \begin{subfigure}{.20\linewidth}
        \includegraphics[width=\linewidth]{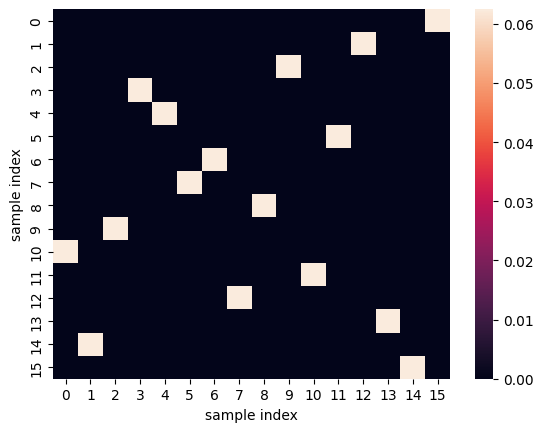}
        \caption{}
        %\label{fig:yelp0}
    \end{subfigure}\hfill
    \begin{subfigure}{.20\linewidth}
        \includegraphics[width=\linewidth]{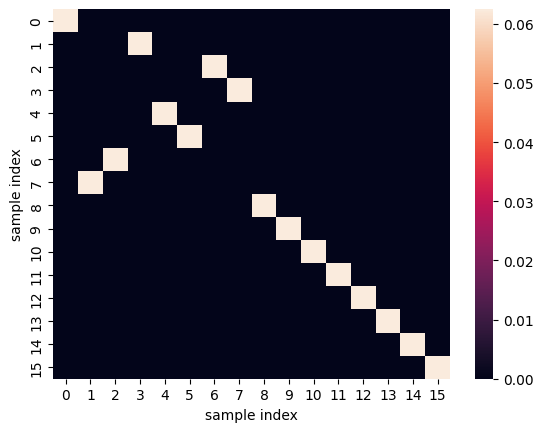}
        \caption{}
         %\label{fig:yelp1}
    \end{subfigure}\hfill
    \begin{subfigure}{.20\linewidth}
        \includegraphics[width=\linewidth]{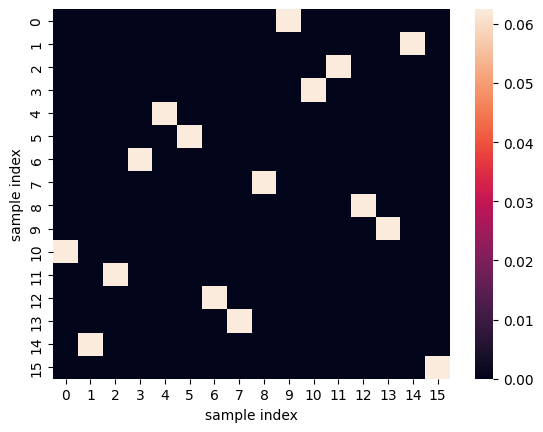}
        \caption{}
         %\label{fig:yelp2}
    \end{subfigure}\hfill
    \begin{subfigure}{.20\linewidth}
        \includegraphics[width=\linewidth]{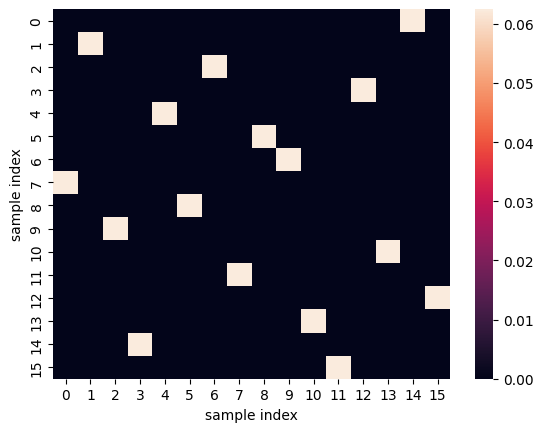}
        \caption{}
         %\label{fig:yelp3}
    \end{subfigure}\hfill
    \begin{subfigure}{.20\linewidth}
        \includegraphics[width=\linewidth]{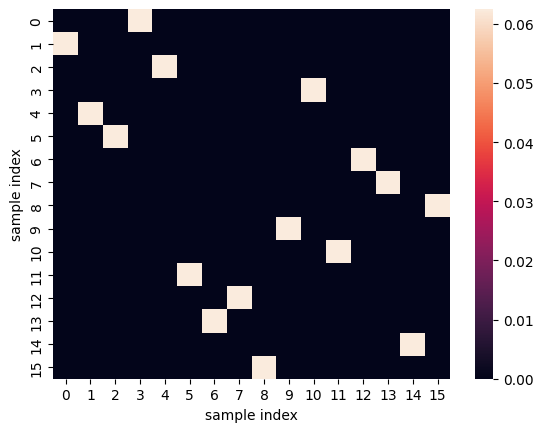}
        \caption{}
         %\label{fig:yelp4}
    \end{subfigure}
    \caption{\small{Pairwise GW transportation plan on Yelp, across BERT models. 5 of randomly chosen layer pairs are shown.}}
    \label{fig:gw_align}
\end{figure}

To show the exact transportation plan from GW distances, we choose plot one batch of data with size 16, and show the transportation plan over 5 random layer pairs in Figure~\ref{fig:gw_align}. As one can see, the transportation plan does not conform to identity-mapping. Both Wasserstein and Euclidean distance will likely have trouble handle in this case. We also note that the transportation plan shown Figure~\ref{fig:gw_align} is a permutation  of the original data, rather than a distributed transportation plan. This behavior is consistent with existing Wasserstein optimal transport plan under certain conditions \citep{peyre2019computational}.

\label{appendix:alignment}

\begin{figure}[h]
     \begin{subfigure}{.20\linewidth}
        \includegraphics[width=\linewidth]{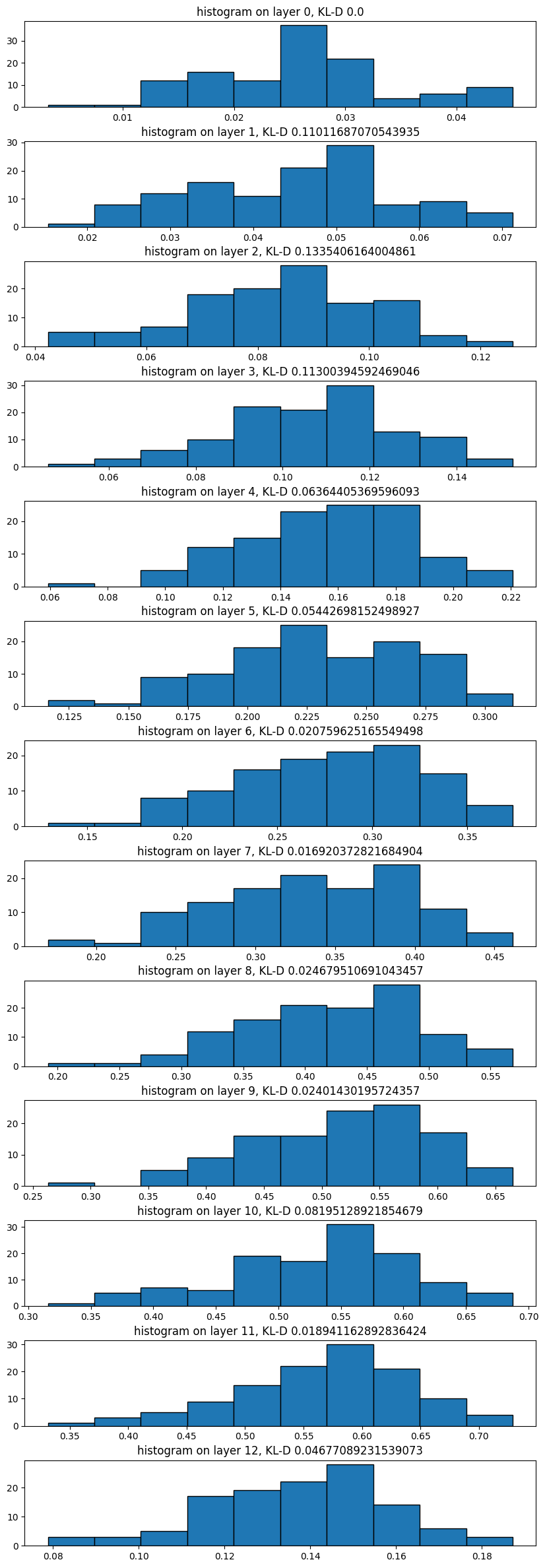}
        \caption{Pretrained}
        \label{fig:yelp0_hist_full}
    \end{subfigure}\hfill
    \begin{subfigure}{.20\linewidth}
        \includegraphics[width=\linewidth]{figs/YELP_bert-base-uncased_target_unknown_spectral_GW_scores_histo.png}
        \caption{Dense}
        \label{fig:yelp1_hist_full}
    \end{subfigure}\hfill
    \begin{subfigure}{.20\linewidth}
        \includegraphics[width=\linewidth]{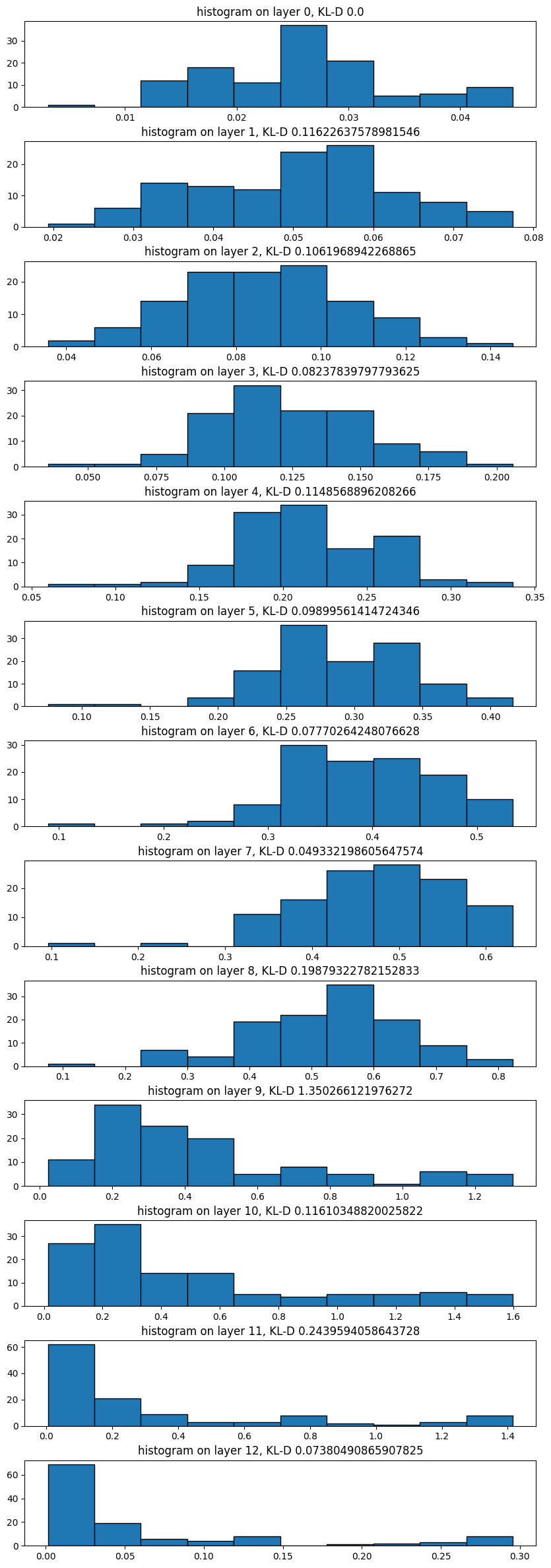}
        \caption{25\% Sparse}
        \label{fig:yelp2_hist_full}
    \end{subfigure}\hfill
    \begin{subfigure}{.20\linewidth}
        \includegraphics[width=\linewidth]{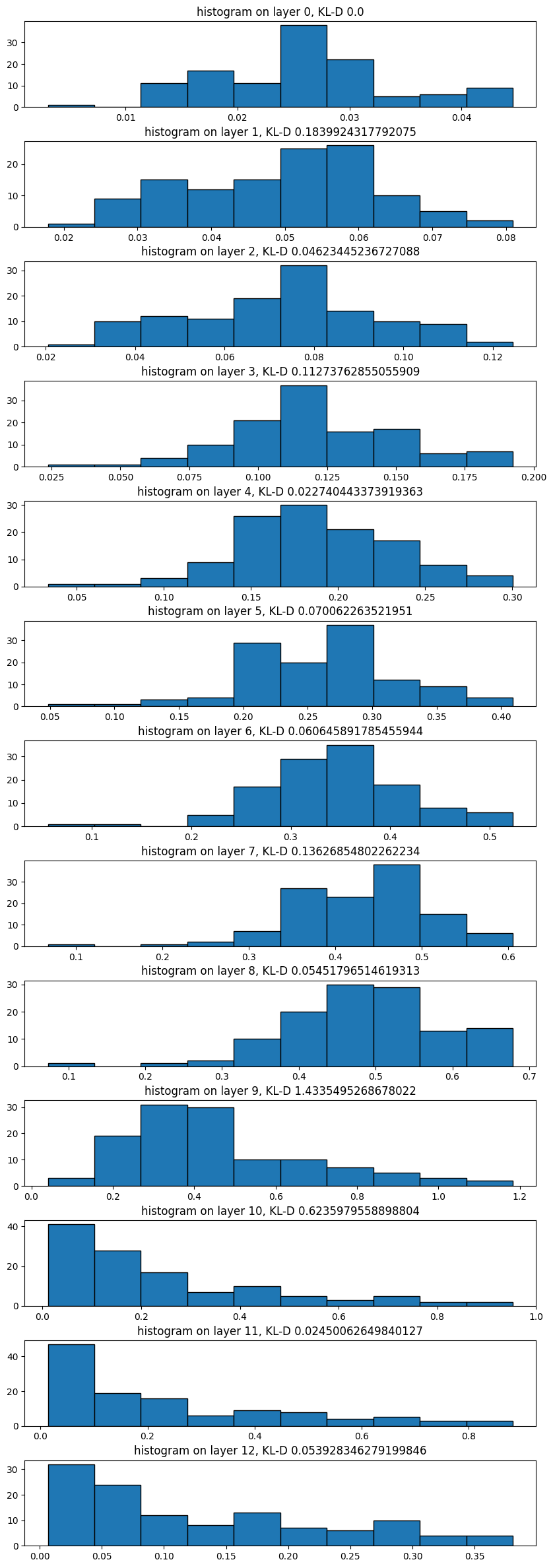}
        \caption{70\% Sparse}
        \label{fig:yelp3_hist_full}
    \end{subfigure}\hfill
    \begin{subfigure}{.20\linewidth}
        \includegraphics[width=\linewidth]{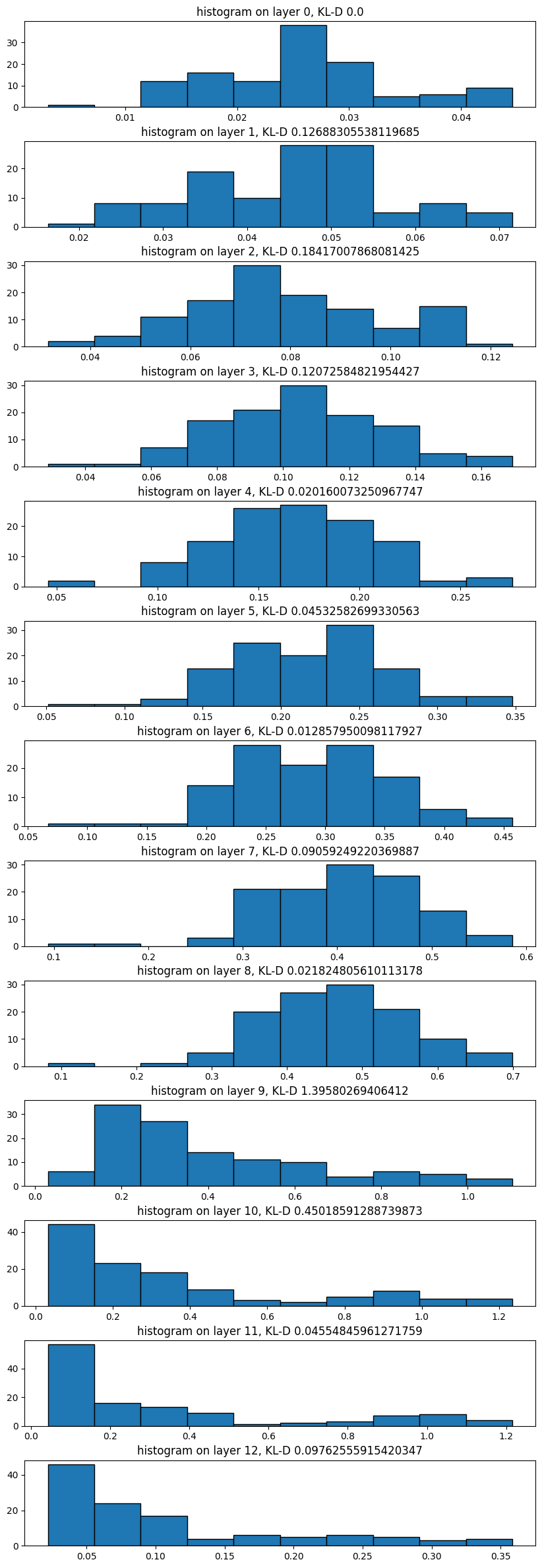}
        \caption{95\% Sparse}
        \label{fig:yelp4_hist_full}
    \end{subfigure}
    \caption{\small{Histogram on pairwise distances on Yelp, across  BERT models with different sparsity levels. $a)$ is the pre-trained BERT and the rest are fine tuned BERT models with increasing sparsity levels: ($b)$densely fine-tuned, $c)$ $25\%$, $d)$ $70\%$ and $e)$ $95\%$ sparsity.}}
    \label{fig:yelp_hist_full}
\end{figure}

To complement Figure~\ref{fig:yelp_hist} on other fine-tuned BERT models on YELP, we also plot all the histograms of pairwise distances between two samples in a batch, across all layers for each of 5 models in Figure~\ref{fig:yelp_hist_full}. Pre-trained models are publicly available models training on other datasets. Row b) to e) are the fine-tuned models on YELP, with different sparsity levels. As one can see, pretrained models do not have much differentiations across layers in the histograms, with maximal KL-divergence of 0.11 between histogram in consecutive layers. Fine tuned models, on the other hard, show larger KL-divergence values, in particular in later layers. For example, Layers 9 in the Dense BERT model contains KL distance of 1.58 from its previous layer. The results show that significant transformations in pairwise distances occur across layers and such distances would be captured by GW distances,  as show in Figure~\ref{fig:yelp_w} and Figure~\ref{fig:yelp_w_new}.

\section{Fine-tuning with Different Layers}
\label{sec:appendix_finetune}
Since the GW distance indicates significant changes occurs only at later layers in YELPS, we investigate performance of fine-tuning only partial layers from pretrained models, by freezing early layers during training and training only later layers alongside a classification layer (denoted as C) at the end.  In Table~\ref{table:freezing}, we can see that there is no significant performance differences between fine-tuning layer 8 to 12 and fine-tuning layer 9 to 12 (0.04\% drop). On the other hand, the accuracy drops 6 times more by freezing layer 1 to 9, with 0.25\%. Freezing layer 1 to 10 results 0.49\%  drop, and finally fine-tuning only 12 results $3.59\%$ drop. These findings validate that the later layers are crucial for significant functional changes.

\begin{table}[h]
\small
  \caption{Accuracy of fine-tuning partial layers in various BERT models. C denotes the classification layer on top of BERT models. }
%  \label{table:fine-tune}
  \centering
 \begin{tabular}{l||c|c|c|c|c|c}
    \toprule
Fine-tune & All & 8$\sim$12 + C & 9$\sim$12 + C  & 10$\sim$12 + C & 11$\sim$12 + C & Only 12 + C \\
\hline
Accuracy (\%) &  $97.87$   & 97.47 & 97.43  &97.19 & 96.7 & 93.11 \\
% \midrule
% Model 1&{GW-D for}  & Top Similar Layers& $D_{\text{min}}=$ \\
%  &     $c_1$  &  $z_1^2$ & $0.02$ \\
%  &   $c_2$ &  $z_1^2$ & $0.02$\\
% &     $c$ & $z_1^2$, 6 others, and Resid-Post$^3$  & $0.02$ \\
    \bottomrule
  \end{tabular}
  \label{table:freezing}
\end{table}

\section{Baseline Methods and Implementation Details}
\label{appendix:baselines}
 Besides the standard Euclidean and cosine distances, we compare a few other baselines, as discussed below.

 Wasserstein Distance \citep{dwivedi2019representation}: We use the POT, python optimal transport library   \url{pythonot} \citep{flamary2021pot}, with the algorithm proposed in \citep{bonneel2011displacement}.

 Representational similarity metric (RSM) \citep{klabunde2023similarity}: RSM compares two different spaces by  using the $L_2$ norms on differences in inter-instances distances. This can be seen as approximation to GW using the fixed and identity transportation plan (i.e., the samples map to itself). We use existing implementation at:  \url{https://github.com/mklabunde/llm_repsim/blob/main/llmcomp/measures/rsm_norm_difference.py}.

 Representational Similarity Analysis (RSA) \citep{klabunde2023similarity}: RSA is similar to RSM  but use correlation instead of $L_2$-norm to compute the final distance. Implementation at: \url{https://github.com/mklabunde/llm_repsim/blob/main/llmcomp/measures/rsa.py}

 Canonical Correlation analysis (CCA) \citep{morcos2018insights}: CCA compute distances based on variances and covariances. Implementation at: \url{https://github.com/google/svcca/blob/master/cca_core.py}

 Centered Kernel Alignment (CKA) \citep{kornblith2019similarity}: CKA is based on normalized Hilbert-Schmidt Independence Criterion (HSIC).  Implementation at:
 \url{https://github.com/mklabunde/llm_repsim/blob/main/llmcomp/measures/cka.py}

Multi-Scale Intrinsic Distance (MSID) \cite{tsitsulin2019shape}: MSID compute the intrinsic and multiple distance, and can be considered as a lower bound of the GW distance. Implementation at: \url{https://github.com/xgfs/imd/blob/master/msid/msid.py}. We have explored different hyperparameter settings with different neighbors k (5 or all batch data available) and number of iterations for SLQ, but results are all similar to the default parameter setting. 

 Augmented GW (AGW) \citep{demetci2023revisiting}: AGW considers feature alignment in addition to sample alignment. Its overall objective can be seen as a penalized GW distance. Implementation at:  \url{https://github.com/pinardemetci/AGW-AISTATS24/tree/main}. 

 For all methods, we use default parameter settings to obtain results in the paper. Note that RSM, RSA, CCA, MSID, and AGW, along with our proposed approach can handle different dimensions of inputs.

Gromo-Wasserstein Distance \citep{dwivedi2019representation}: We use the POT, python optimal transport library   \url{pythonot} \citep{flamary2021pot}. We use the solver based on the conditional gradient  \citep{titouan2019optimal}.

\section{More Baselines on YELP}
\label{appendix:yelp_morebaselines}
\begin{figure}[h]
   % \caption{\small{Pairwise GW distance on Yelp, across BERT models. (a) is pre-trained BERT and (b)-(e) are fine tuned BERT models with increasing sparsity (dense, $25\%$, $70\%$ and $95\%$ sparsity).}}
   % \label{fig:yelp}
% \end{figure}
% \begin{figure}[h]
 \vskip -3mm
  \begin{subfigure}{.20\linewidth}
        \includegraphics[width=\linewidth]{figs/euc_YELP_dense.png}
        %\caption{}
         %\label{fig:yelp0}
    \end{subfigure}\hfill
    \begin{subfigure}{.20\linewidth}
        \includegraphics[width=\linewidth]{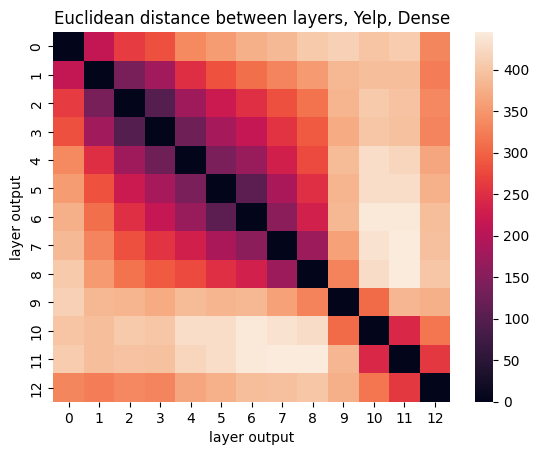}
        %\caption{}
         %\label{fig:yelp1}
    \end{subfigure}\hfill
    \begin{subfigure}{.20\linewidth}
        \includegraphics[width=\linewidth]{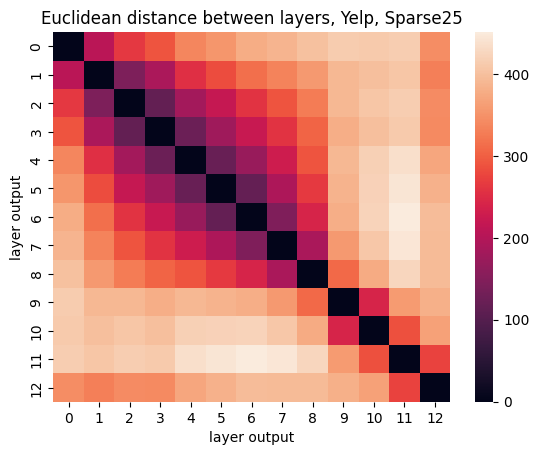}
        %\caption{}
         %\label{fig:yelp2}
    \end{subfigure}\hfill
    \begin{subfigure}{.20\linewidth}
        \includegraphics[width=\linewidth]{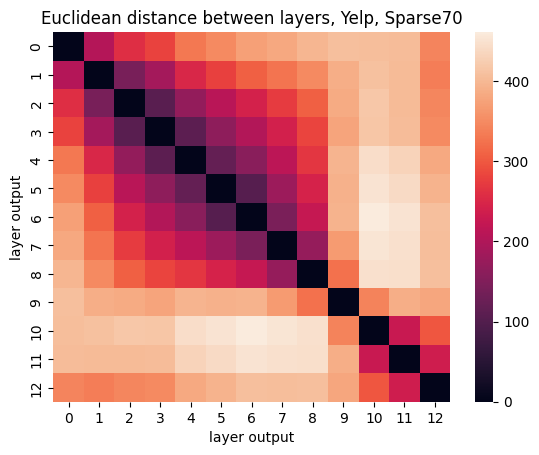}
        %\caption{}
         %\label{fig:yelp3}
    \end{subfigure}\hfill
    \begin{subfigure}{.20\linewidth}
        \includegraphics[width=\linewidth]{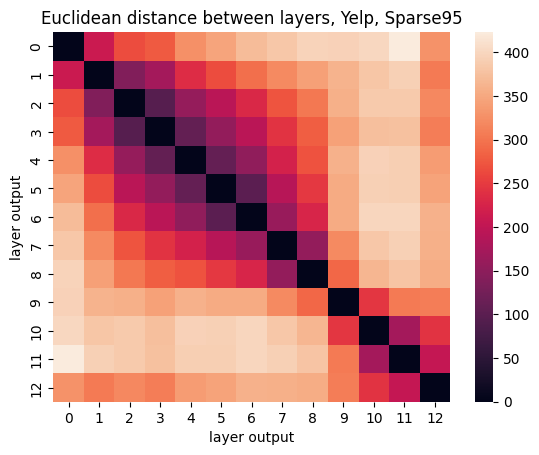}
        %\caption{}
        %\label{fig:yelp4}
    \end{subfigure}
%     \caption{\small{Pairwise Euclidean distance on Yelp, across BERT models. (a) is pre-trained BERT and (b)-(e) are fine tuned BERT models with increasing sparsity (dense, $25\%$, $70\%$ and $95\%$ sparsity).}}
%     \label{fig:yelp_euc}
% \end{figure}
% \begin{figure}[h]
\newline
 \begin{subfigure}{.20\linewidth}
        \includegraphics[width=\linewidth]{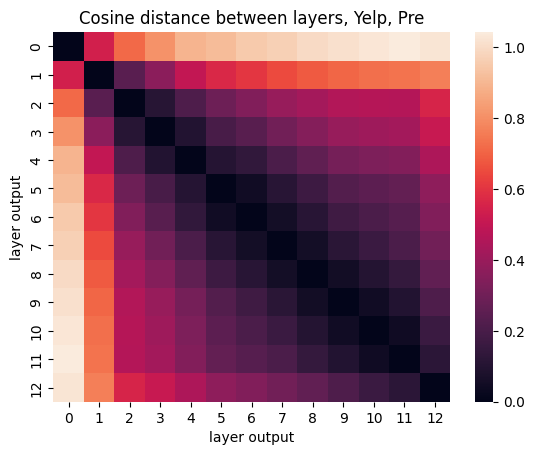}
        %\caption{}
        %\label{fig:yelp0}
    \end{subfigure}\hfill
    \begin{subfigure}{.20\linewidth}
        \includegraphics[width=\linewidth]{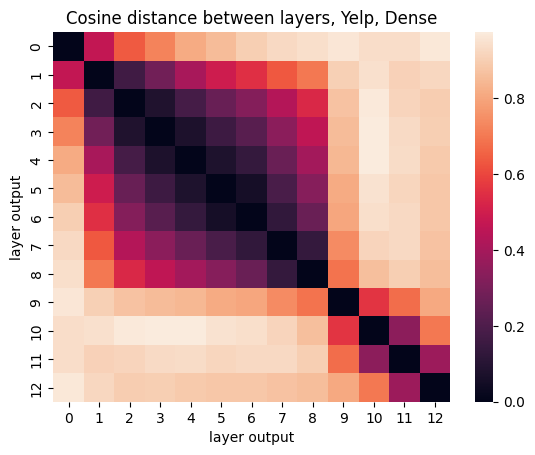}
        %\caption{}
         %\label{fig:yelp1}
    \end{subfigure}\hfill
    \begin{subfigure}{.20\linewidth}
        \includegraphics[width=\linewidth]{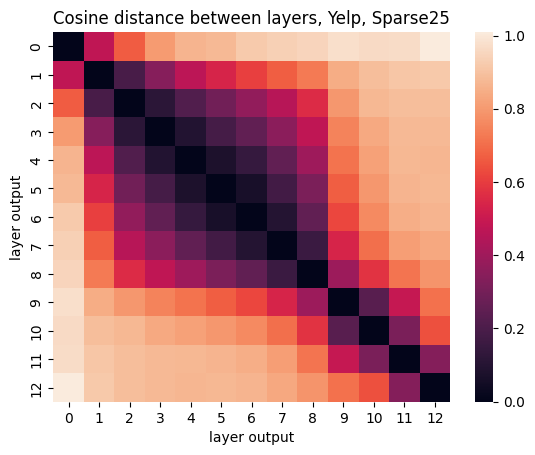}
        %\caption{}
         %\label{fig:yelp2}
    \end{subfigure}\hfill
    \begin{subfigure}{.20\linewidth}
        \includegraphics[width=\linewidth]{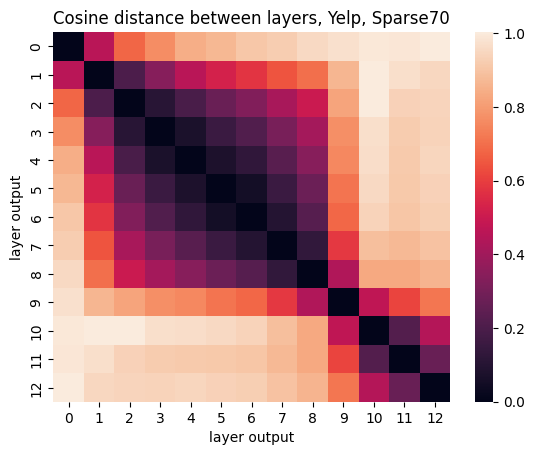}
        %\caption{}
         %\label{fig:yelp3}
    \end{subfigure}\hfill
    \begin{subfigure}{.20\linewidth}
        \includegraphics[width=\linewidth]{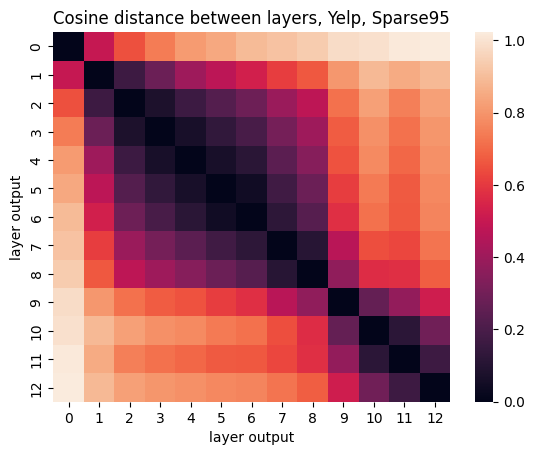}
        %\caption{}
         %\label{fig:yelp4}
    \end{subfigure}
\newline
  \begin{subfigure}{.20\linewidth}
        \includegraphics[width=\linewidth]{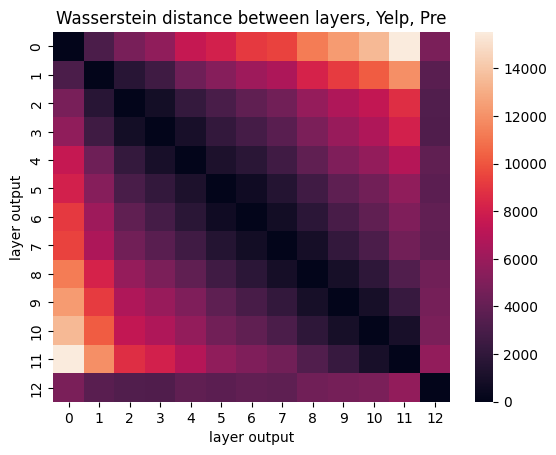}
        %\caption{}
        %\label{fig:yelp0}
    \end{subfigure}\hfill
    \begin{subfigure}{.20\linewidth}
        \includegraphics[width=\linewidth]{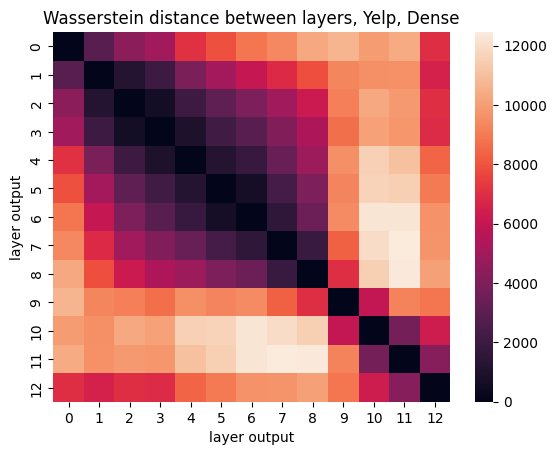}
        %\caption{}
         %\label{fig:yelp1}
    \end{subfigure}\hfill
    \begin{subfigure}{.20\linewidth}
        \includegraphics[width=\linewidth]{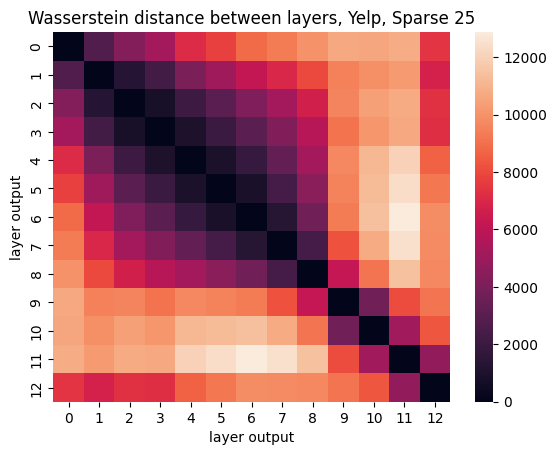}
        %\caption{}
         %\label{fig:yelp2}
    \end{subfigure}\hfill
    \begin{subfigure}{.20\linewidth}
        \includegraphics[width=\linewidth]{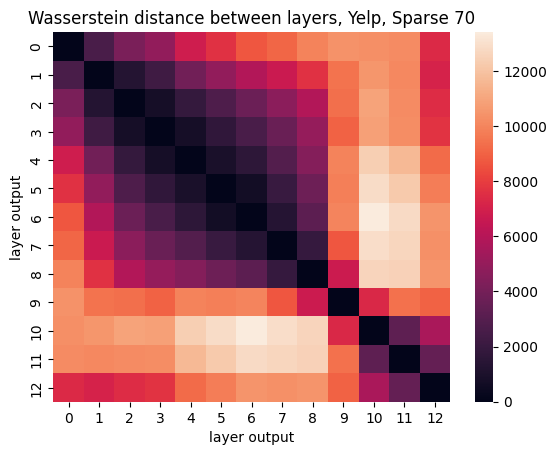}
        %\caption{}
         %\label{fig:yelp3}
    \end{subfigure}\hfill
    \begin{subfigure}{.20\linewidth}
        \includegraphics[width=\linewidth]{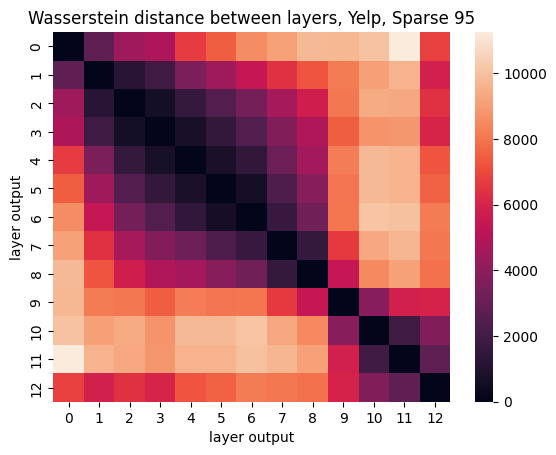}
        %\caption{}
         %\label{fig:yelp4}
    \end{subfigure}
%     \caption{\small{Pairwise Wasserstein distance on Yelp, across BERT models. (a) is pre-trained BERT and (b)-(e) are fine tuned BERT models with increasing sparsity (dense, $25\%$, $70\%$ and $95\%$ sparsity).}}
%     \label{fig:yelp_w}
% \end{figure}
% \begin{figure}[h]
\newline
     \begin{subfigure}{.20\linewidth}
        \includegraphics[width=\linewidth]{figs/gw_Yelp_pre.png}
        \caption{Pre-trained}
     %    \label{fig:yelp0}
    \end{subfigure}\hfill
    \begin{subfigure}{.20\linewidth}
        \includegraphics[width=\linewidth]{figs/gw_Yelp_dense.png}
        \caption{Dense}
      %   \label{fig:yelp1}
    \end{subfigure}\hfill
    \begin{subfigure}{.20\linewidth}
        \includegraphics[width=\linewidth]{figs/gw_Yelp_sparse25.png}
        \caption{25\% Sparse}
     %    \label{fig:yelp2}
    \end{subfigure}\hfill
    \begin{subfigure}{.20\linewidth}
        \includegraphics[width=\linewidth]{figs/gw_Yelp_sparse70.png}
        \caption{70\% Sparse}
     %   \label{fig:yelp3}
    \end{subfigure}\hfill
    \begin{subfigure}{.20\linewidth}
        \includegraphics[width=\linewidth]{figs/gw_Yelp_sparse95.png}
        \caption{95\% Sparse}
      %  \label{fig:yelp4}
    \end{subfigure}
    \caption{\small{Pairwise (layer) distances on Yelp, across different BERT models. Different Rows:  \textit{Euclidean, Cosine, Wasserstein, and the proposed GW distance}, from top to bottom. Different columns: first column is the pre-trained BERT and the rest are fine tuned BERT models with increasing sparsity (dense, $25\%$, $70\%$ and $95\%$ sparsity). As can be seen GW clearly demarcates the (functional) sub-network blocks.}}
    \label{fig:yelp_w_2old}
    \vskip -3mm
\end{figure}

\begin{figure}[h]
   % \caption{\small{Pairwise GW distance on Yelp, across BERT models. (a) is pre-trained BERT and (b)-(e) are fine tuned BERT models with increasing sparsity (dense, $25\%$, $70\%$ and $95\%$ sparsity).}}
   % \label{fig:yelp}
% \end{figure}
% \begin{figure}[h]
 \vskip -3mm
  \begin{subfigure}{.20\linewidth}
        \includegraphics[width=\linewidth]{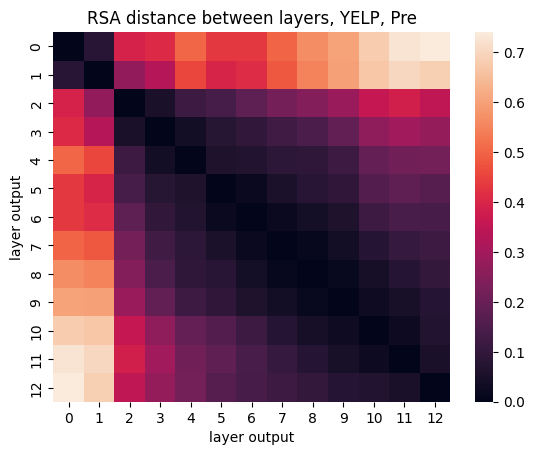}
        %\caption{}
        %\label{fig:yelp0}
    \end{subfigure}\hfill
    \begin{subfigure}{.20\linewidth}
        \includegraphics[width=\linewidth]{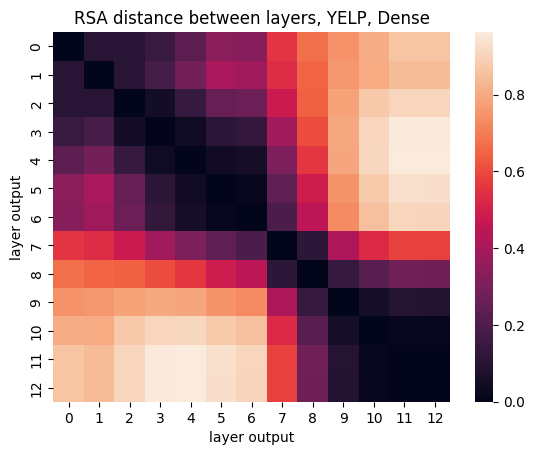}
        %\caption{}
         %\label{fig:yelp1}
    \end{subfigure}\hfill
    \begin{subfigure}{.20\linewidth}
        \includegraphics[width=\linewidth]{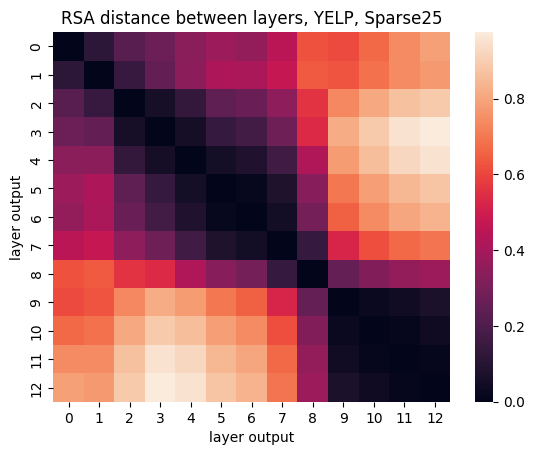}
        %\caption{}
         %\label{fig:yelp2}
    \end{subfigure}\hfill
    \begin{subfigure}{.20\linewidth}
        \includegraphics[width=\linewidth]{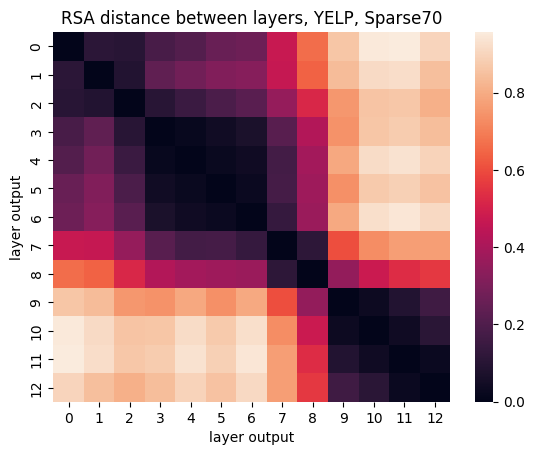}
        %\caption{}
         %\label{fig:yelp3}
    \end{subfigure}\hfill
    \begin{subfigure}{.20\linewidth}
        \includegraphics[width=\linewidth]{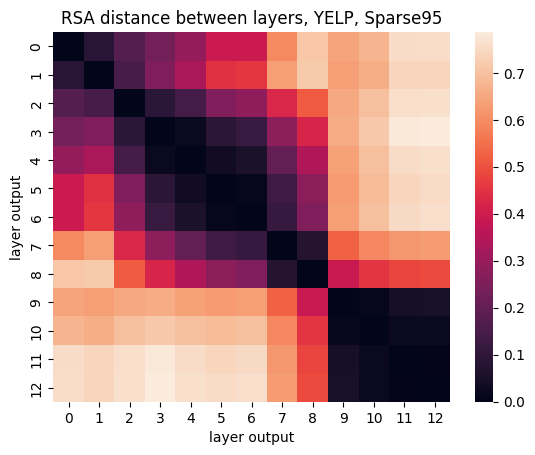}
        %\caption{}
         %\label{fig:yelp4}
    \end{subfigure}
\newline
  \begin{subfigure}{.20\linewidth}
        \includegraphics[width=\linewidth]{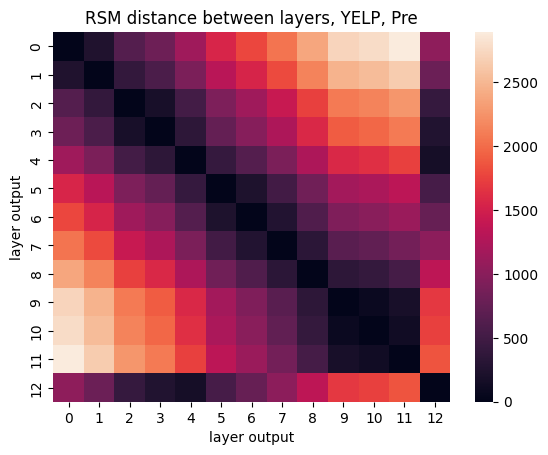}
        %\caption{}
        %\label{fig:yelp0}
    \end{subfigure}\hfill
    \begin{subfigure}{.20\linewidth}
        \includegraphics[width=\linewidth]{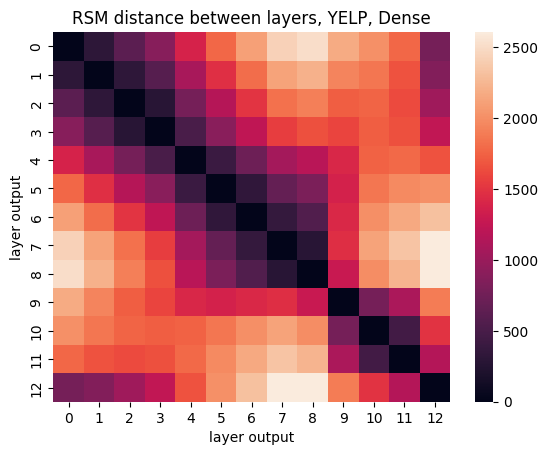}
        %\caption{}
         %\label{fig:yelp1}
    \end{subfigure}\hfill
    \begin{subfigure}{.20\linewidth}
        \includegraphics[width=\linewidth]{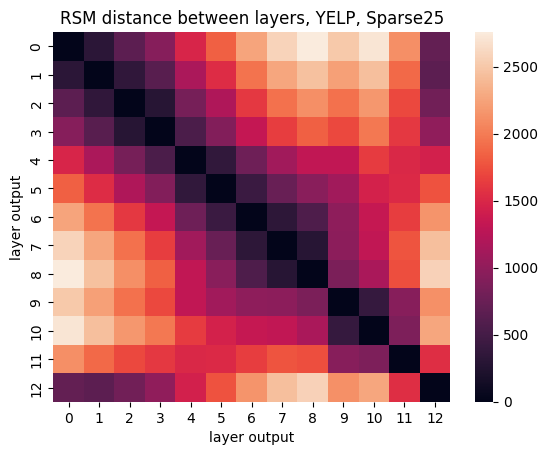}
        %\caption{}
         %\label{fig:yelp2}
    \end{subfigure}\hfill
    \begin{subfigure}{.20\linewidth}
        \includegraphics[width=\linewidth]{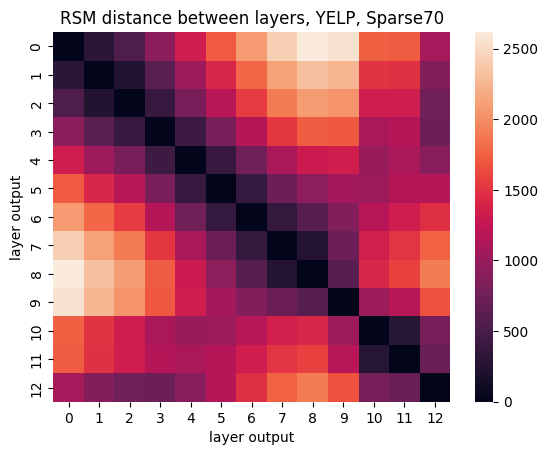}
        %\caption{}
         %\label{fig:yelp3}
    \end{subfigure}\hfill
    \begin{subfigure}{.20\linewidth}
        \includegraphics[width=\linewidth]{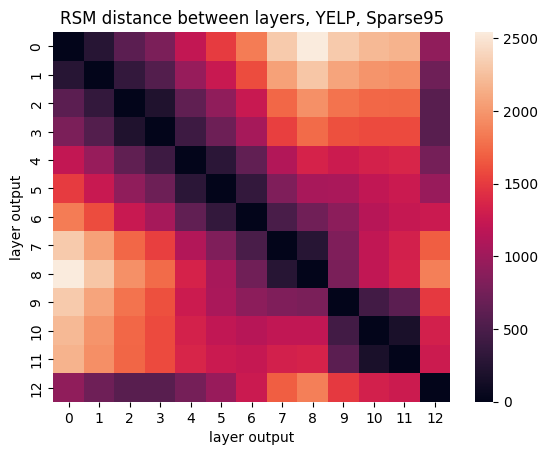}
        %\caption{}
         %\label{fig:yelp4}
    \end{subfigure}
\newline
  \begin{subfigure}{.20\linewidth}
        \includegraphics[width=\linewidth]{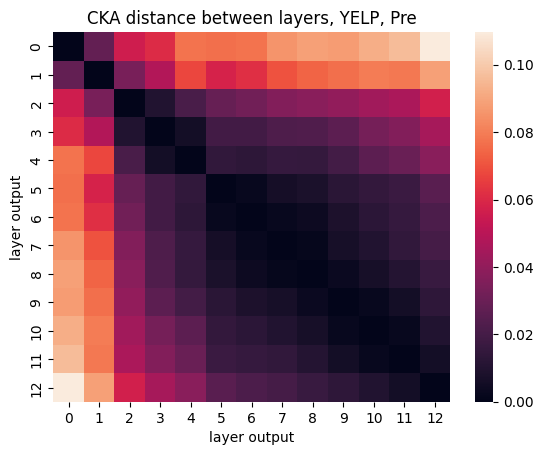}
        %\caption{}
        %\label{fig:yelp0}
    \end{subfigure}\hfill
    \begin{subfigure}{.20\linewidth}
        \includegraphics[width=\linewidth]{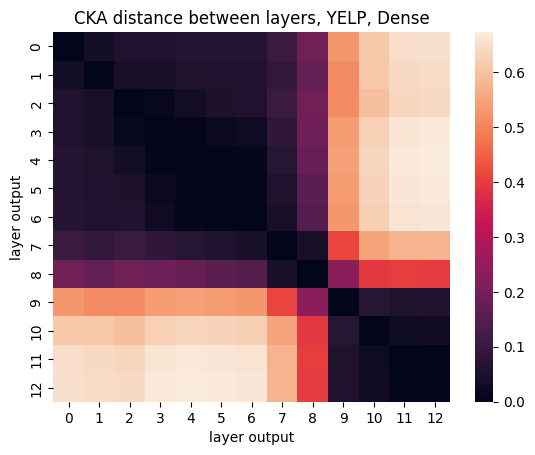}
        %\caption{}
         %\label{fig:yelp1}
    \end{subfigure}\hfill
    \begin{subfigure}{.20\linewidth}
        \includegraphics[width=\linewidth]{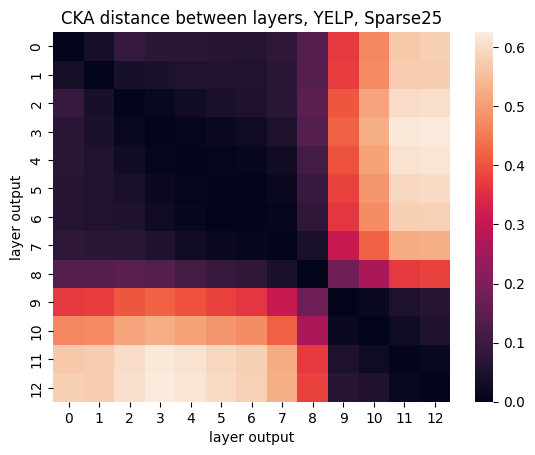}
        %\caption{}
         %\label{fig:yelp2}
    \end{subfigure}\hfill
    \begin{subfigure}{.20\linewidth}
        \includegraphics[width=\linewidth]{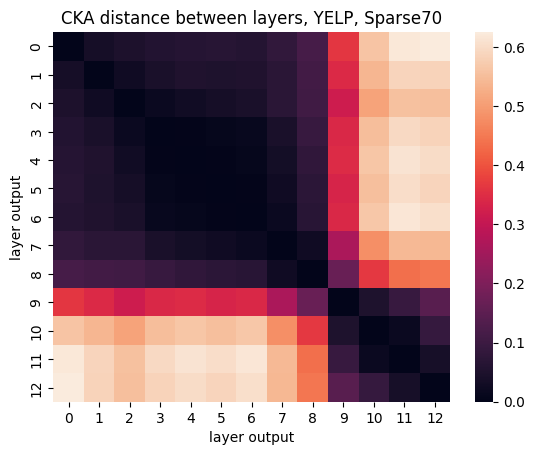}
        %\caption{}
         %\label{fig:yelp3}
    \end{subfigure}\hfill
    \begin{subfigure}{.20\linewidth}
        \includegraphics[width=\linewidth]{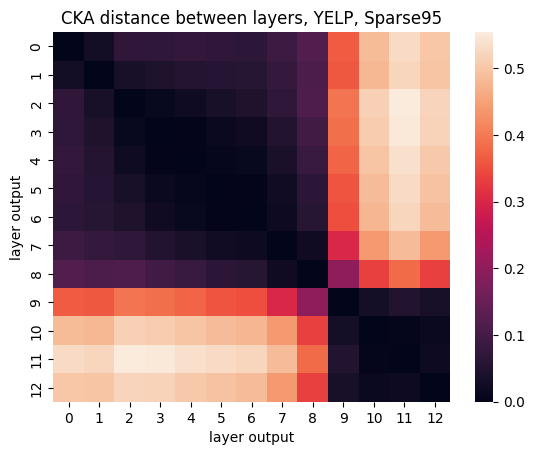}
        %\caption{}
         %\label{fig:yelp4}
    \end{subfigure}
\newline
  \begin{subfigure}{.20\linewidth}
        \includegraphics[width=\linewidth]{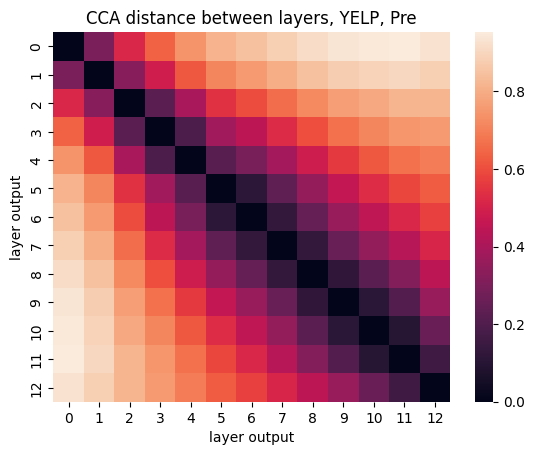}
        %\caption{}
        %\label{fig:yelp0}
    \end{subfigure}\hfill
    \begin{subfigure}{.20\linewidth}
        \includegraphics[width=\linewidth]{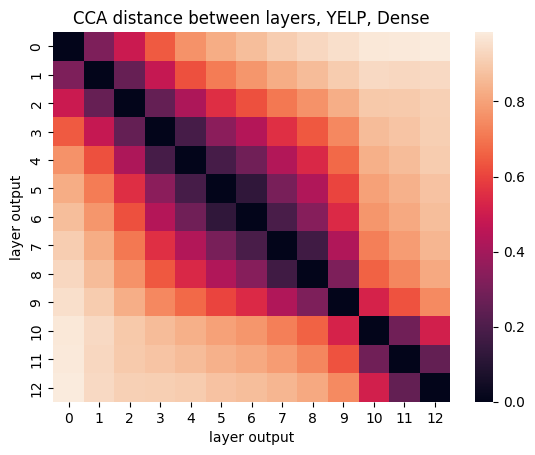}
        %\caption{}
         %\label{fig:yelp1}
    \end{subfigure}\hfill
    \begin{subfigure}{.20\linewidth}
        \includegraphics[width=\linewidth]{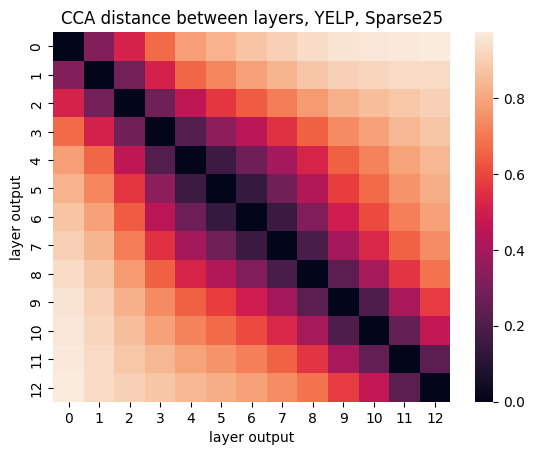}
        %\caption{}
         %\label{fig:yelp2}
    \end{subfigure}\hfill
    \begin{subfigure}{.20\linewidth}
        \includegraphics[width=\linewidth]{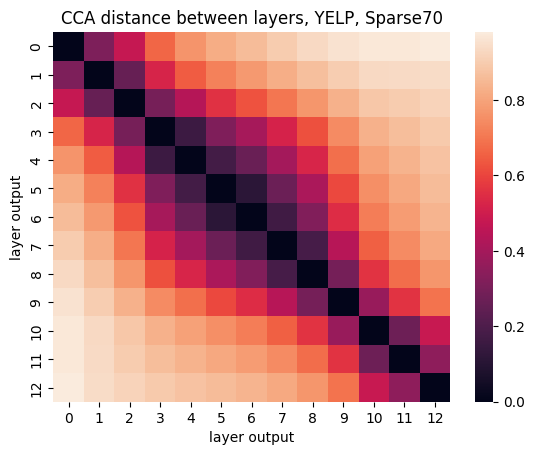}
        %\caption{}
         %\label{fig:yelp3}
    \end{subfigure}\hfill
    \begin{subfigure}{.20\linewidth}
        \includegraphics[width=\linewidth]{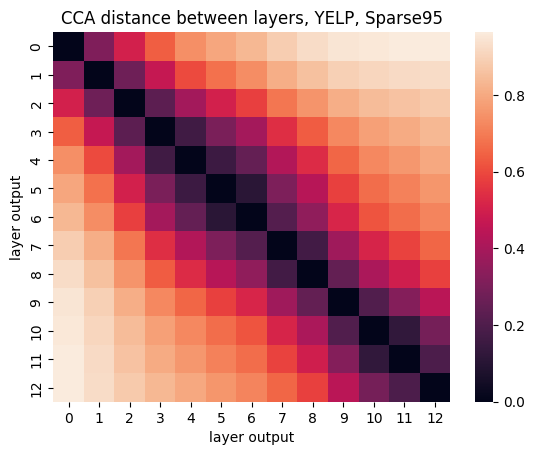}
        %\caption{}
         %\label{fig:yelp4}
    \end{subfigure}
\newline
  \begin{subfigure}{.20\linewidth}
        \includegraphics[width=\linewidth]{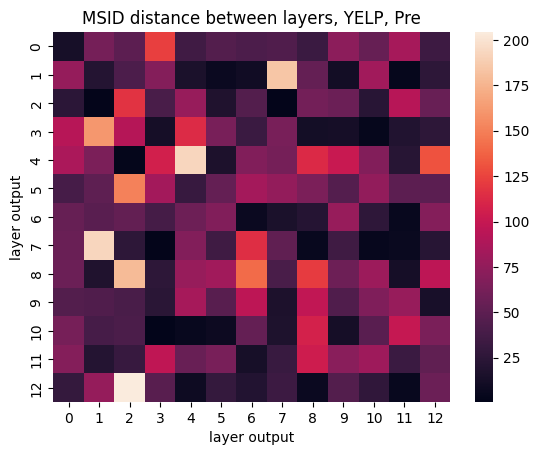}
        %\caption{}
        %\label{fig:yelp0}
    \end{subfigure}\hfill
    \begin{subfigure}{.20\linewidth}
        \includegraphics[width=\linewidth]{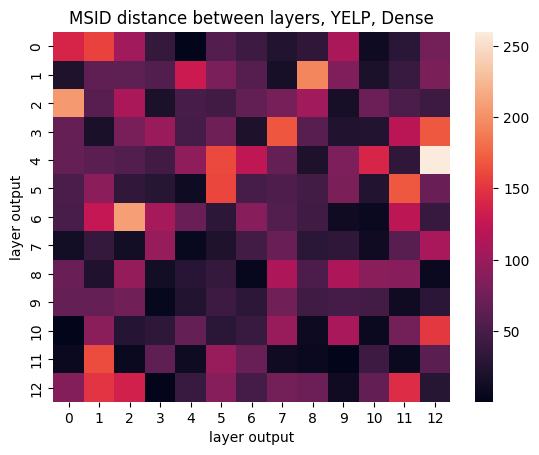}
        %\caption{}
         %\label{fig:yelp1}
    \end{subfigure}\hfill
    \begin{subfigure}{.20\linewidth}
        \includegraphics[width=\linewidth]{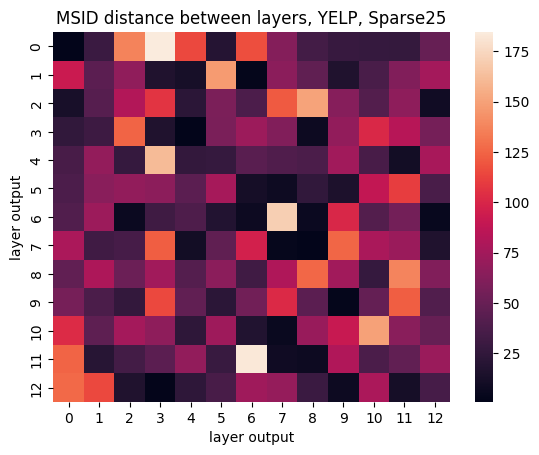}
        %\caption{}
         %\label{fig:yelp2}
    \end{subfigure}\hfill
    \begin{subfigure}{.20\linewidth}
        \includegraphics[width=\linewidth]{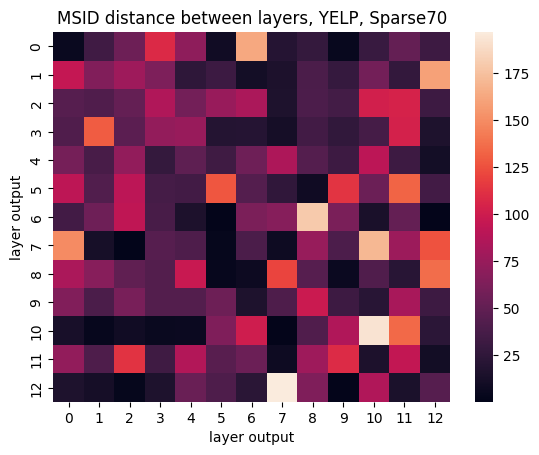}
        %\caption{}
         %\label{fig:yelp3}
    \end{subfigure}\hfill
    \begin{subfigure}{.20\linewidth}
        \includegraphics[width=\linewidth]{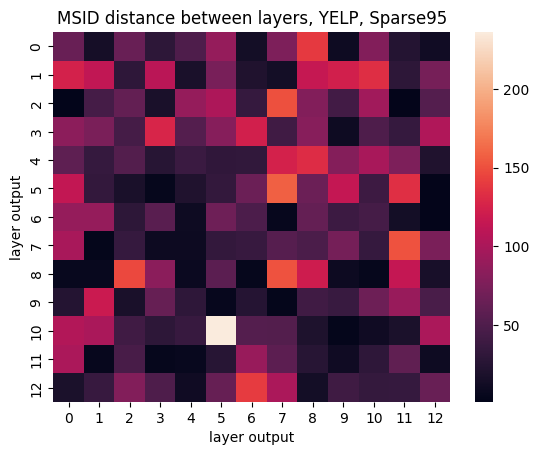}
        %\caption{}
         %\label{fig:yelp4}
    \end{subfigure}
\newline
  \begin{subfigure}{.20\linewidth}
        \includegraphics[width=\linewidth]{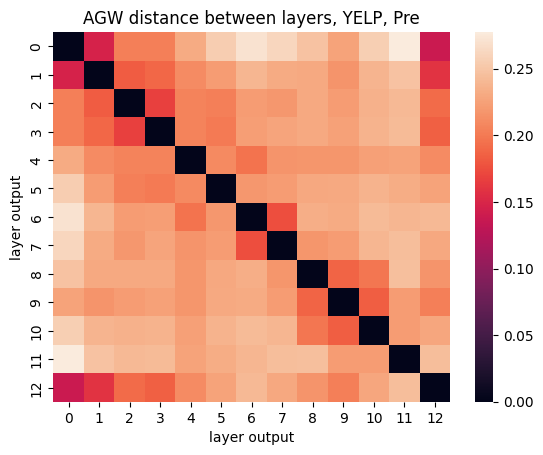}
        %\caption{}
        %\label{fig:yelp0}
    \end{subfigure}\hfill
    \begin{subfigure}{.20\linewidth}
        \includegraphics[width=\linewidth]{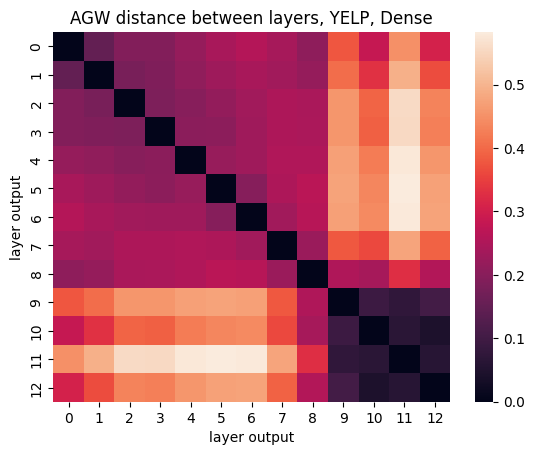}
        %\caption{}
         %\label{fig:yelp1}
    \end{subfigure}\hfill
    \begin{subfigure}{.20\linewidth}
        \includegraphics[width=\linewidth]{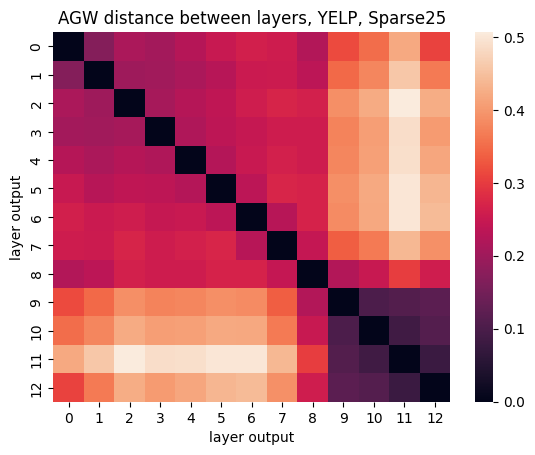}
        %\caption{}
         %\label{fig:yelp2}
    \end{subfigure}\hfill
    \begin{subfigure}{.20\linewidth}
        \includegraphics[width=\linewidth]{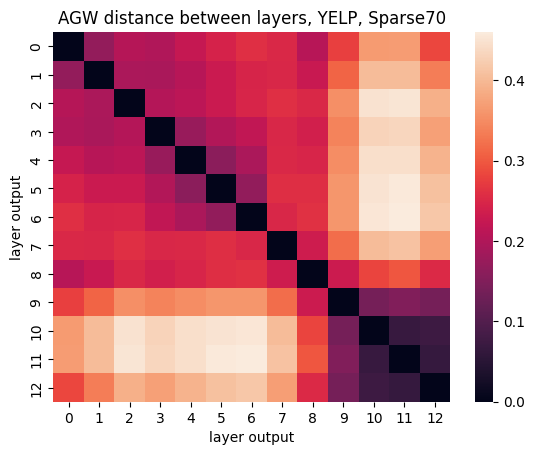}
        %\caption{}
         %\label{fig:yelp3}
    \end{subfigure}\hfill
    \begin{subfigure}{.20\linewidth}
        \includegraphics[width=\linewidth]{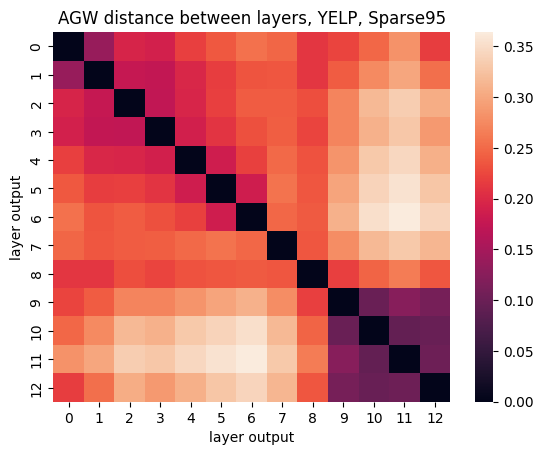}
        %\caption{}
         %\label{fig:yelp4}
    \end{subfigure}
% \newline
%   \begin{subfigure}{.20\linewidth}
%         \includegraphics[width=\linewidth]{figs/w_Yelp_pre.png}
%         %\caption{}
%         %\label{fig:yelp0}
%     \end{subfigure}\hfill
%     \begin{subfigure}{.20\linewidth}
%         \includegraphics[width=\linewidth]{figs/w_Yelp_dense.png}
%         %\caption{}
%          %\label{fig:yelp1}
%     \end{subfigure}\hfill
%     \begin{subfigure}{.20\linewidth}
%         \includegraphics[width=\linewidth]{figs/w_Yelp_sparse25.png}
%         %\caption{}
%          %\label{fig:yelp2}
%     \end{subfigure}\hfill
%     \begin{subfigure}{.20\linewidth}
%         \includegraphics[width=\linewidth]{figs/w_Yelp_sparse70.png}
%         %\caption{}
%          %\label{fig:yelp3}
%     \end{subfigure}\hfill
%     \begin{subfigure}{.20\linewidth}
%         \includegraphics[width=\linewidth]{figs/w_Yelp_sparse95.png}
%         %\caption{}
%          %\label{fig:yelp4}
%     \end{subfigure}
%     \caption{\small{Pairwise Wasserstein distance on Yelp, across BERT models. (a) is pre-trained BERT and (b)-(e) are fine tuned BERT models with increasing sparsity (dense, $25\%$, $70\%$ and $95\%$ sparsity).}}
%     \label{fig:yelp_w}
% \end{figure}
% \begin{figure}[h]
\newline
     \begin{subfigure}{.20\linewidth}
        \includegraphics[width=\linewidth]{figs/gw_Yelp_pre.png}
        \caption{Pre-trained}
        %\label{fig:yelp0}
    \end{subfigure}\hfill
    \begin{subfigure}{.20\linewidth}
        \includegraphics[width=\linewidth]{figs/gw_Yelp_dense.png}
        \caption{Dense}
        %\label{fig:yelp1}
    \end{subfigure}\hfill
    \begin{subfigure}{.20\linewidth}
        \includegraphics[width=\linewidth]{figs/gw_Yelp_sparse25.png}
        \caption{25\% Sparse}
        %\label{fig:yelp2}
    \end{subfigure}\hfill
    \begin{subfigure}{.20\linewidth}
        \includegraphics[width=\linewidth]{figs/gw_Yelp_sparse70.png}
        \caption{70\% Sparse}
        %\label{fig:yelp3}
    \end{subfigure}\hfill
    \begin{subfigure}{.20\linewidth}
        \includegraphics[width=\linewidth]{figs/gw_Yelp_sparse95.png}
        \caption{95\% Sparse}
        %\label{fig:yelp4}
    \end{subfigure}
%---------------

    \caption{\small{Pairwise (layer) distances on Yelp, across different BERT models. Different Rows:  \textit{RSA, RSM, CKA, CCA, MSID, AGW, and the proposed GW distance}, from top to bottom. Different columns: first column is the pre-trained BERT and the rest are fine tuned BERT models with increasing sparsity (dense, $25\%$, $70\%$ and $95\%$ sparsity). As one can be seen, GW clearly demarcates the (functional) sub-network blocks.}}
    \label{fig:yelp_w_new}
    \vskip -3mm
\end{figure}

Due to the page limit, here we include  baseline results on Yelp Datasets in Figure~\ref{fig:yelp_w_2old} and Figure~\ref{fig:yelp_w_new}.  

We  compare the proposed GW distance with Euclidean, Cosine, and Wasserstein distance as baselines in Figure~\ref{fig:yelp_w_2old}, on the same YELP dataset and with the same settings. 
Euclidean distance between two layers' outputs, shown in the first row of Figure~\ref{fig:yelp_w_2old}, can be seen as the GW distance with a fixed identity-mapping transportation plan for each sample. This validates the low-valued diagonal elements. Off-diagonal elements show greater variation, and it is less obvious there are two distinct sub-groups within layers. The similar pattern is also observed with Cosine and Wasserstein distances, with similar strong diagonal pattern but more pronounced  block structures than Euclidean distance.  we also include 6 other baseline similarity measure  in Figure~\ref{fig:yelp_w_new}. Overall, CKA produces also similar block structures to the proposed GW distance, though with greater variability within block structures. In contrast, other baselines fail to reveal such clear block structures. 

%\red{TODO switch baselines with the figure in the main paper}

\section{Cross Model Comparison} 
\label{appendix:cross}
\begin{figure}[h]
%     \caption{\small{Pairwise GW distance in SST2, across BERT models. (a) is pre-trained BERT and (b)-(e) are fine tuned BERT models with increasing sparsity (dense, $25\%$, $70\%$ and $95\%$ sparsity).}}
%     \label{fig:sst}
% \end{figure}
% \vskip -5mm
% \begin{figure}[h]
  \begin{subfigure}{.33\linewidth}
        \includegraphics[width=\linewidth]{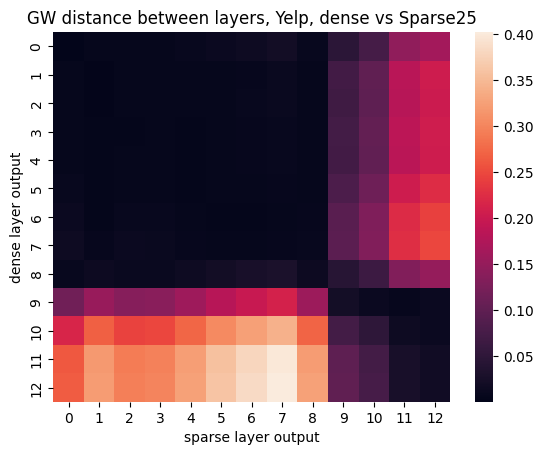}
        %\\caption{}
         %\label{fig:yelp0}
    \end{subfigure}\hfill
    \begin{subfigure}{.33\linewidth}
        \includegraphics[width=\linewidth]{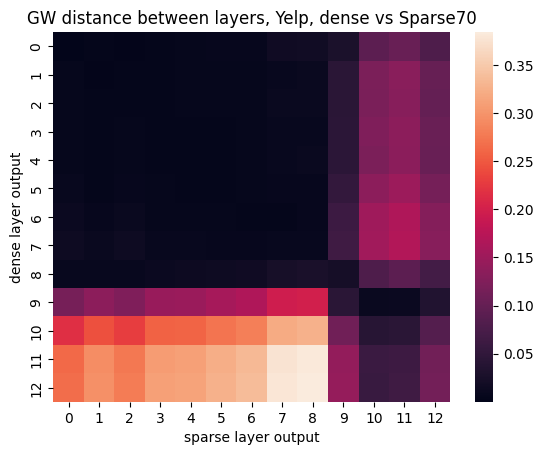}
      %\  \caption{}
         %\label{fig:yelp1}
    \end{subfigure}\hfill
    \begin{subfigure}{.33\linewidth}
        \includegraphics[width=\linewidth]{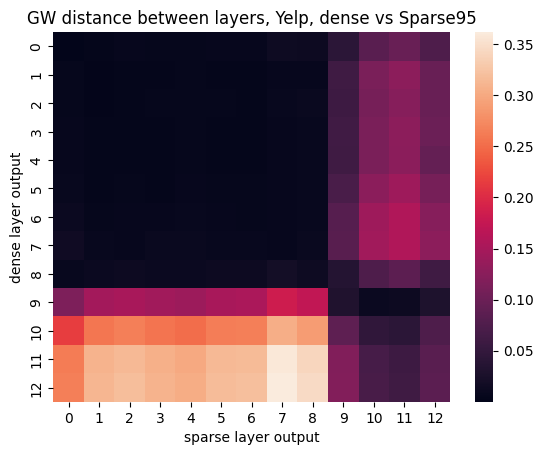}
       %\ \caption{}
         %\label{fig:yelp2}
    \end{subfigure}\hfill
    % \begin{subfigure}{.20\linewidth}
    %     \includegraphics[width=\linewidth]{figs/euc_SST_sparse70.png}
    %    %\ \caption{}
    %      %\label{fig:yelp3}
    % \end{subfigure}\hfill
    % \begin{subfigure}{.20\linewidth}
    %     \includegraphics[width=\linewidth]{figs/euc_SST_sparse95.png}
    %    %\ \caption{}
    %     %\label{fig:yelp4}
    % \end{subfigure}
%     \caption{\small{Pairwise Euclidean distance on SST2, across BERT models. (a) is pre-trained BERT and (b)-(e) are fine tuned BERT models with increasing sparsity (dense, $25\%$, $70\%$ and $95\%$ sparsity).}}
%     \label{fig:sst_euc}
% \end{figure}
% \begin{figure}[h]
\newline
 \begin{subfigure}{.33\linewidth}
        \includegraphics[width=\linewidth]{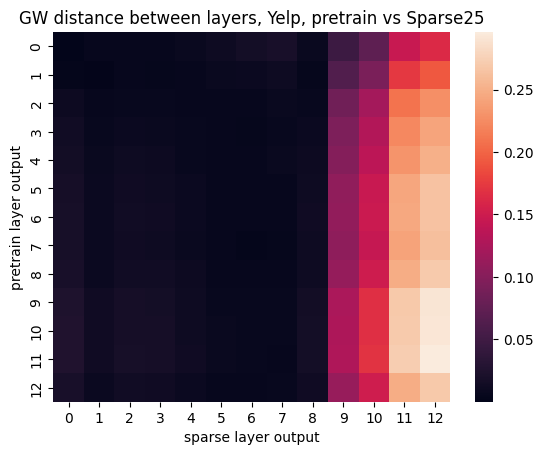}
        %\\caption{}
         %\label{fig:yelp0}
    \end{subfigure}\hfill
    \begin{subfigure}{.33\linewidth}
        \includegraphics[width=\linewidth]{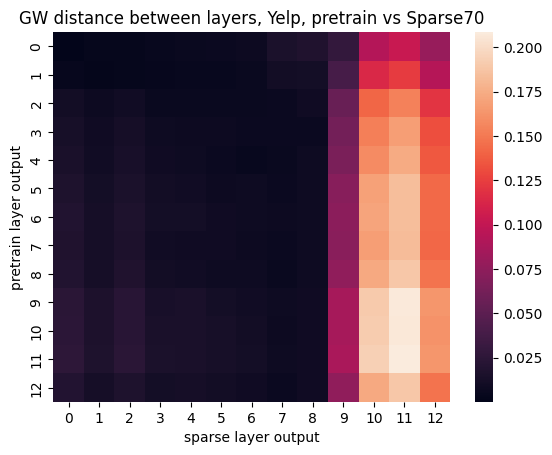}
        %\\caption{}
         %\label{fig:yelp1}
    \end{subfigure}\hfill
    \begin{subfigure}{.33\linewidth}
        \includegraphics[width=\linewidth]{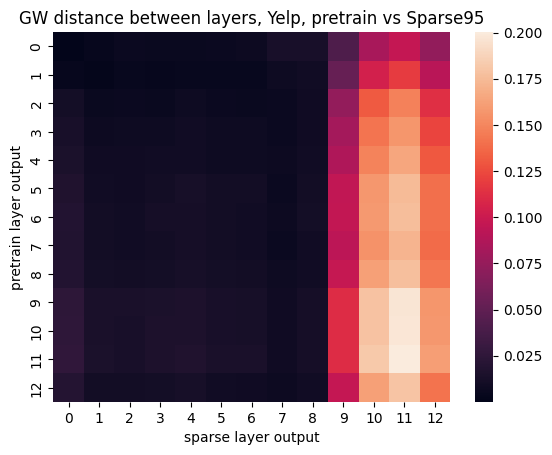}
        %\\caption{}
         %\label{fig:yelp2}
    \end{subfigure}\hfill
      \caption{\small{Pairwise distances on YELP dataset, of layers across two different BERT models. \textit{TOP}: Densely fine-tuned BERT model vs fine-tuned BERT models with different sparsity levels. \textit{Bottom}: Pretrained BERT model vs fine-tuned BERT models with different sparsity levels.}}
     \label{fig:yelp_gw_cross}
\end{figure}

We can also use GW distance to compare layers from different BERT models. Shown in Figure~\ref{fig:yelp_gw_cross}, pretrained and densely fine-tuned BERT models exhibit different similarity measures when compared to  fine-tuned BERT models with  different levels of sparsity.  

\section{SST2 Datasets}
\label{appendix:sst}
\begin{figure}[h]
%     \caption{\small{Pairwise GW distance in SST2, across BERT models. (a) is pre-trained BERT and (b)-(e) are fine tuned BERT models with increasing sparsity (dense, $25\%$, $70\%$ and $95\%$ sparsity).}}
%     \label{fig:sst}
% \end{figure}
% \vskip -5mm
% \begin{figure}[h]
  \begin{subfigure}{.20\linewidth}
        \includegraphics[width=\linewidth]{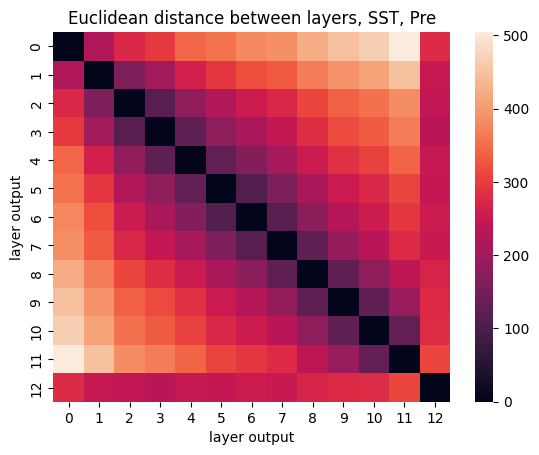}
        %\\caption{}
         %\label{fig:yelp0}
    \end{subfigure}\hfill
    \begin{subfigure}{.20\linewidth}
        \includegraphics[width=\linewidth]{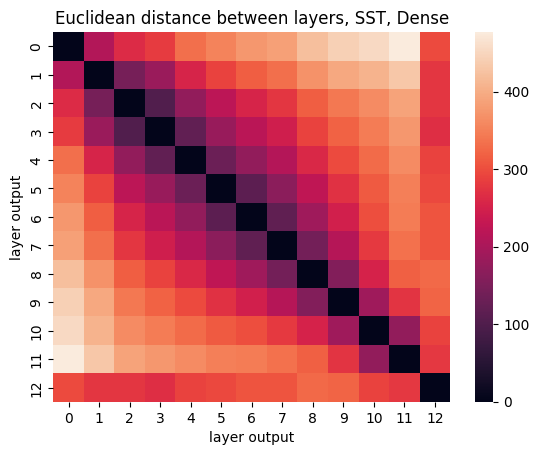}
      %\  \caption{}
         %\label{fig:yelp1}
    \end{subfigure}\hfill
    \begin{subfigure}{.20\linewidth}
        \includegraphics[width=\linewidth]{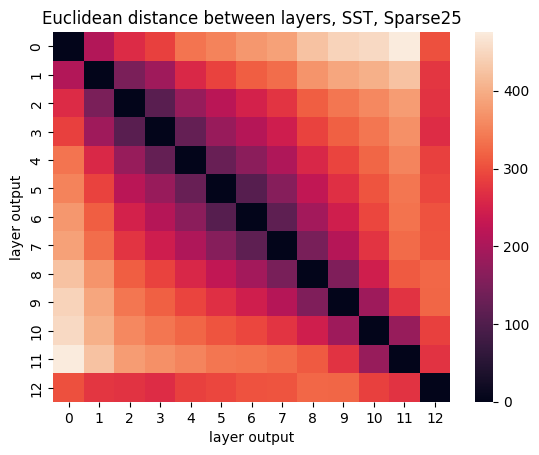}
       %\ \caption{}
         %\label{fig:yelp2}
    \end{subfigure}\hfill
    \begin{subfigure}{.20\linewidth}
        \includegraphics[width=\linewidth]{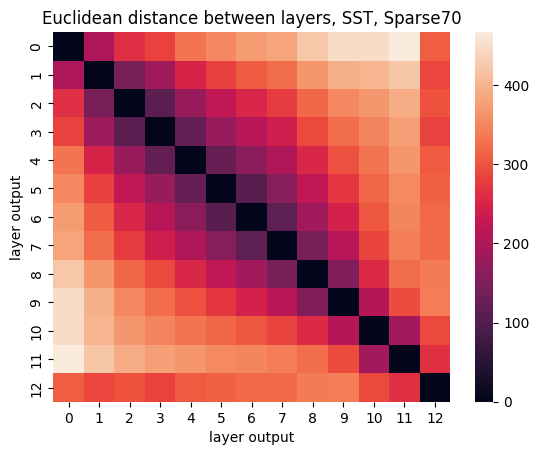}
       %\ \caption{}
         %\label{fig:yelp3}
    \end{subfigure}\hfill
    \begin{subfigure}{.20\linewidth}
        \includegraphics[width=\linewidth]{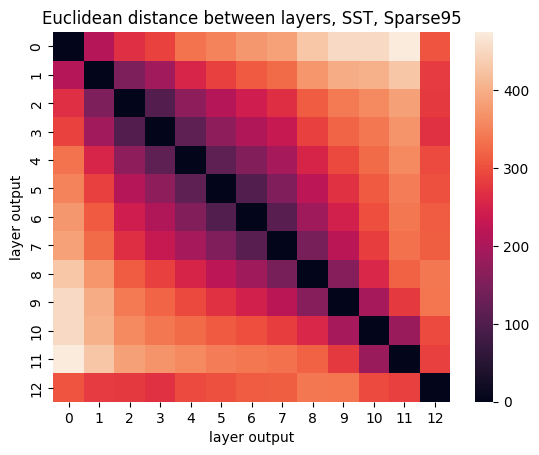}
       %\ \caption{}
        %\label{fig:yelp4}
    \end{subfigure}
%     \caption{\small{Pairwise Euclidean distance on SST2, across BERT models. (a) is pre-trained BERT and (b)-(e) are fine tuned BERT models with increasing sparsity (dense, $25\%$, $70\%$ and $95\%$ sparsity).}}
%     \label{fig:sst_euc}
% \end{figure}
% \begin{figure}[h]
\newline
 \begin{subfigure}{.20\linewidth}
        \includegraphics[width=\linewidth]{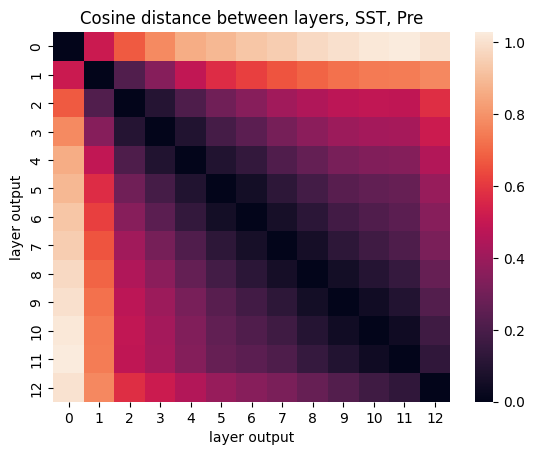}
        %\\caption{}
         %\label{fig:yelp0}
    \end{subfigure}\hfill
    \begin{subfigure}{.20\linewidth}
        \includegraphics[width=\linewidth]{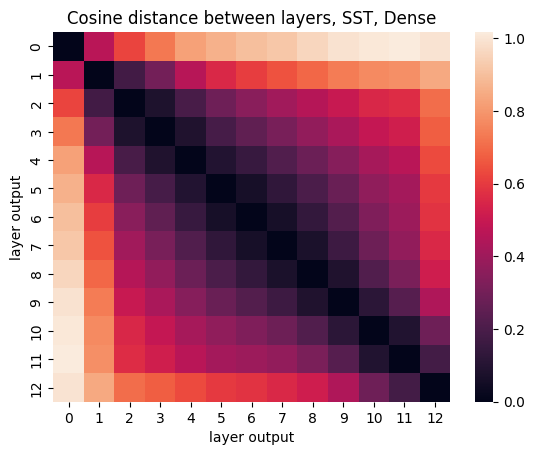}
        %\\caption{}
         %\label{fig:yelp1}
    \end{subfigure}\hfill
    \begin{subfigure}{.20\linewidth}
        \includegraphics[width=\linewidth]{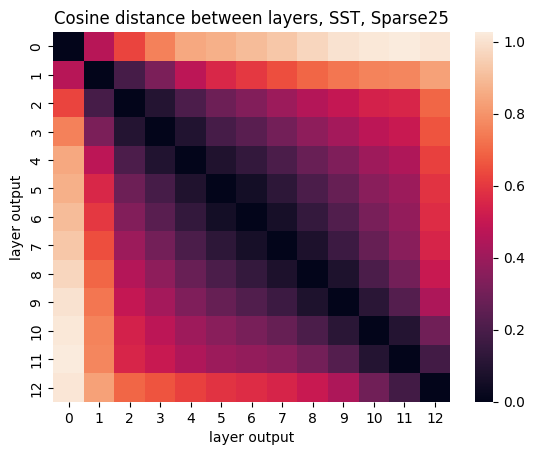}
        %\\caption{}
         %\label{fig:yelp2}
    \end{subfigure}\hfill
    \begin{subfigure}{.20\linewidth}
        \includegraphics[width=\linewidth]{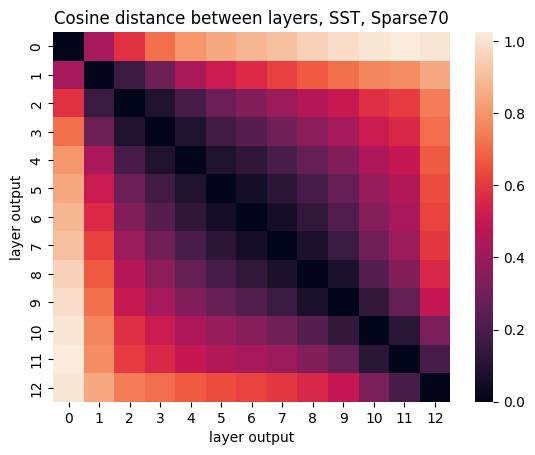}
        %\\caption{}
         %\label{fig:yelp3}
    \end{subfigure}\hfill
    \begin{subfigure}{.20\linewidth}
        \includegraphics[width=\linewidth]{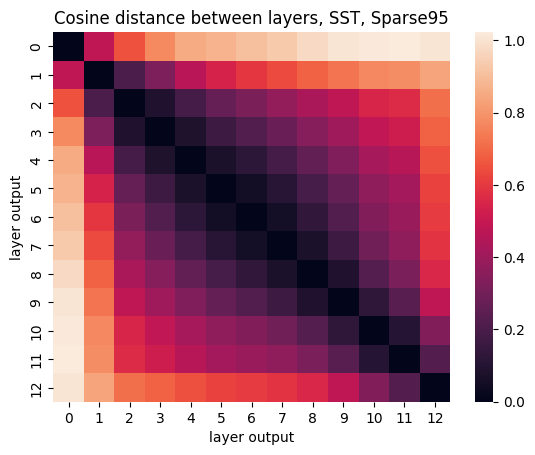}
       %\ \caption{}
        %\label{fig:yelp4}
    \end{subfigure}
\newline
  \begin{subfigure}{.20\linewidth}
        \includegraphics[width=\linewidth]{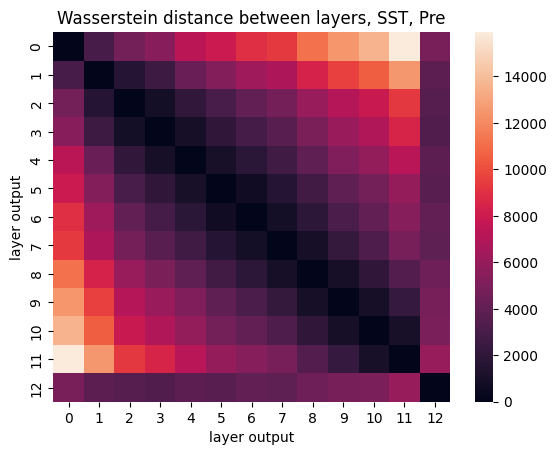}
       %\ \caption{}
        %\label{fig:yelp0}
    \end{subfigure}\hfill
    \begin{subfigure}{.20\linewidth}
        \includegraphics[width=\linewidth]{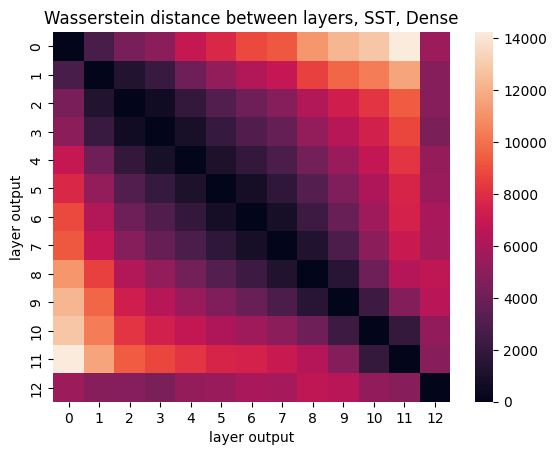}
        %\\caption{}
         %\label{fig:yelp1}
    \end{subfigure}\hfill
    \begin{subfigure}{.20\linewidth}
        \includegraphics[width=\linewidth]{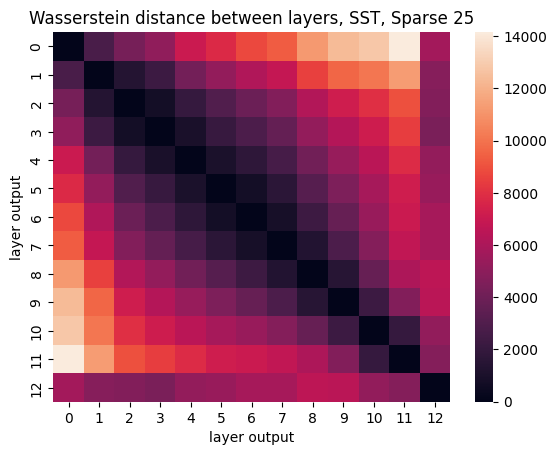}
       %\ \caption{}
         %\label{fig:yelp2}
    \end{subfigure}\hfill
    \begin{subfigure}{.20\linewidth}
        \includegraphics[width=\linewidth]{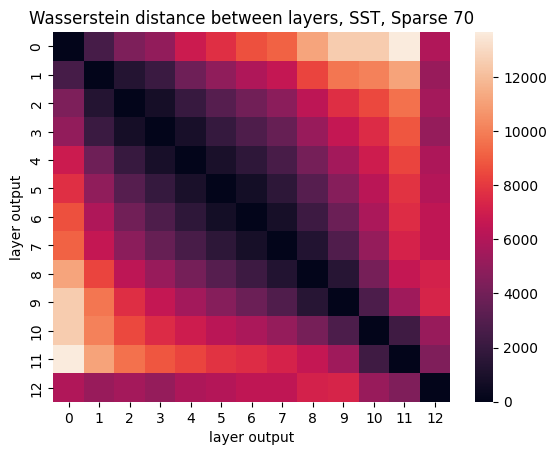}
        %\\caption{}
         %\label{fig:yelp3}
    \end{subfigure}\hfill
    \begin{subfigure}{.20\linewidth}
        \includegraphics[width=\linewidth]{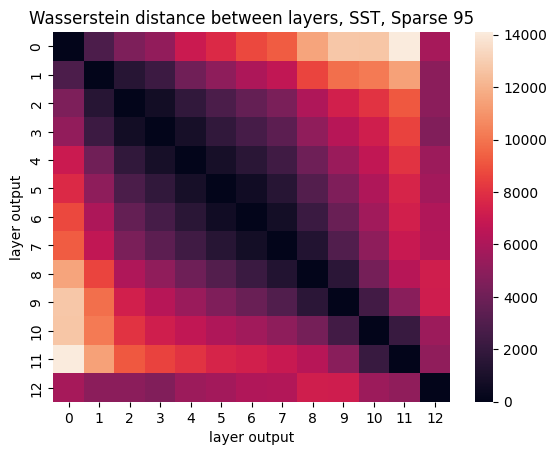}
        %\\caption{}
         %\label{fig:yelp4}
    \end{subfigure}
    \newline
    \begin{subfigure}{.20\linewidth}
        \includegraphics[width=\linewidth]{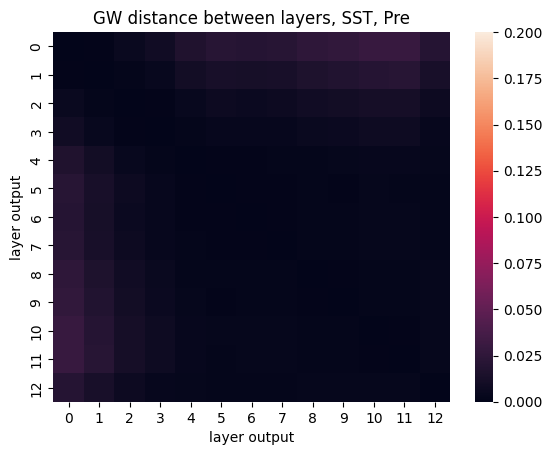}
         \caption{Pretrained}
        \label{fig:sst0}
    \end{subfigure}\hfill
    \begin{subfigure}{.20\linewidth}
        \includegraphics[width=\linewidth]{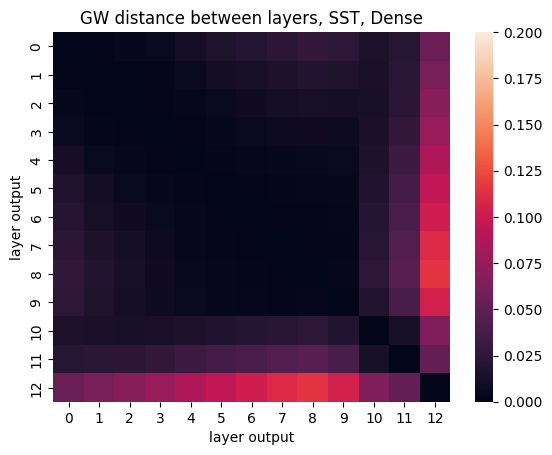}
        \caption{Dense}
        \label{fig:sst1}
    \end{subfigure}\hfill
    \begin{subfigure}{.20\linewidth}
        \includegraphics[width=\linewidth]{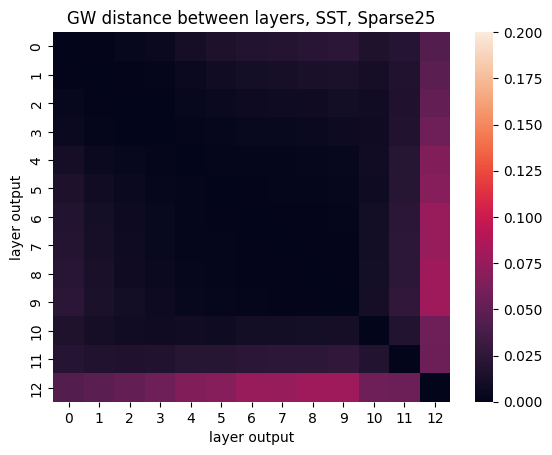}
        \caption{25\% Sparse}
        \label{fig:sst2}
    \end{subfigure}\hfill
    \begin{subfigure}{.20\linewidth}
        \includegraphics[width=\linewidth]{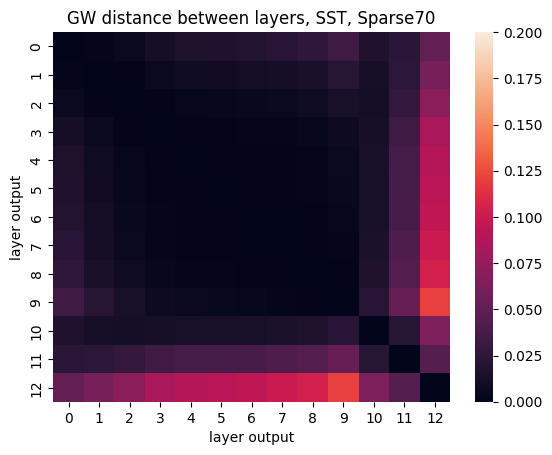}
        \caption{70\% Sparse}
        \label{fig:sst3}
    \end{subfigure}\hfill
    \begin{subfigure}{.20\linewidth}
        \includegraphics[width=\linewidth]{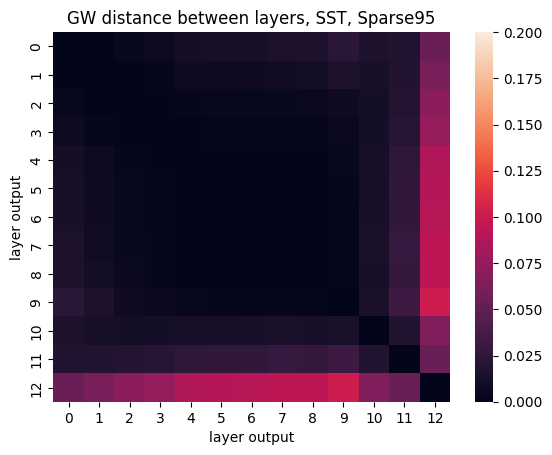}
        \caption{95\% Sparse}
        \label{fig:sst4}
    \end{subfigure}
      \caption{\small{Pairwise distances on SST dataset, across different BERT models. Different Rows:  \textit{Euclidean, Cosine, Wasserstein,  and the proposed GW distances}, from top to bottom. Different columns: first column is the pre-trained BERT and the rest are fine tuned BERT models with increasing sparsity (dense, $25\%$, $70\%$ and $95\%$ sparsity).}}
     \label{fig:sst_w}
\end{figure}

Besides YELP Datasets, we also tested the GW distance on SST2 dataset. Results on SST2 dataset are shown in Figure~\ref{fig:sst_w} again confirm there exist two-three different groups in terms of functional similarity. The first major difference is seen at layers $10$ and $11$, while layer $12$ forms its own block. When sparsifying these models, lesser differences are observed in general as also seen on the YELP dataset. Other baselines provide less clarity on the division of sub-components. 

\begin{figure}[h]
%     \caption{\small{Pairwise GW distance in SST2, across BERT models. (a) is pre-trained BERT and (b)-(e) are fine tuned BERT models with increasing sparsity (dense, $25\%$, $70\%$ and $95\%$ sparsity).}}
%     \label{fig:sst}
% \end{figure}
% \vskip -5mm
% \begin{figure}[h]
  \begin{subfigure}{.20\linewidth}
        \includegraphics[width=\linewidth]{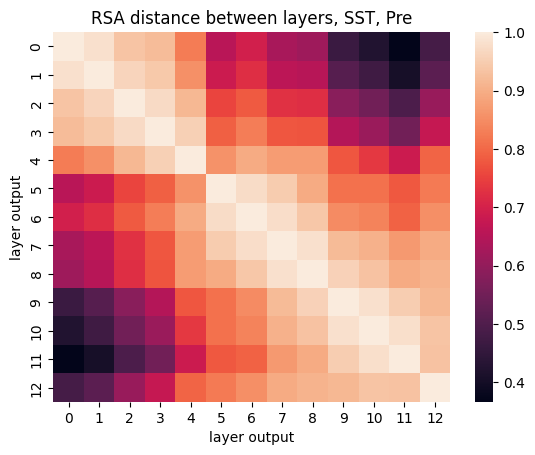}
        %\\caption{}
         %\label{fig:yelp0}
    \end{subfigure}\hfill
    \begin{subfigure}{.20\linewidth}
        \includegraphics[width=\linewidth]{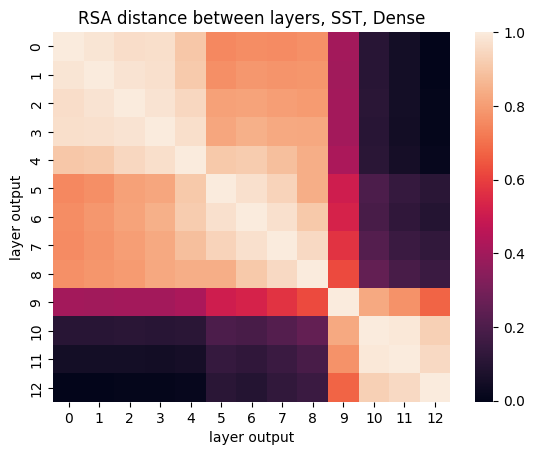}
      %\  \caption{}
         %\label{fig:yelp1}
    \end{subfigure}\hfill
    \begin{subfigure}{.20\linewidth}
        \includegraphics[width=\linewidth]{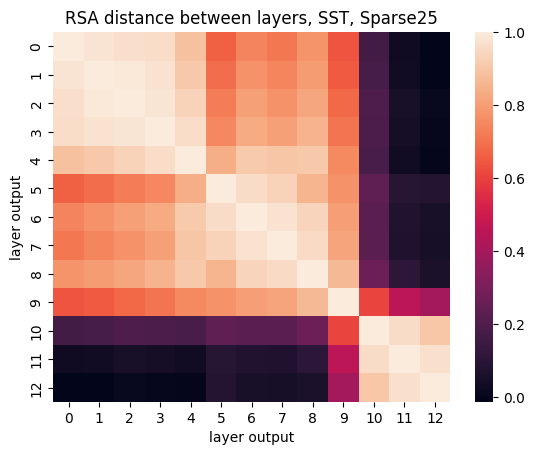}
       %\ \caption{}
         %\label{fig:yelp2}
    \end{subfigure}\hfill
    \begin{subfigure}{.20\linewidth}
        \includegraphics[width=\linewidth]{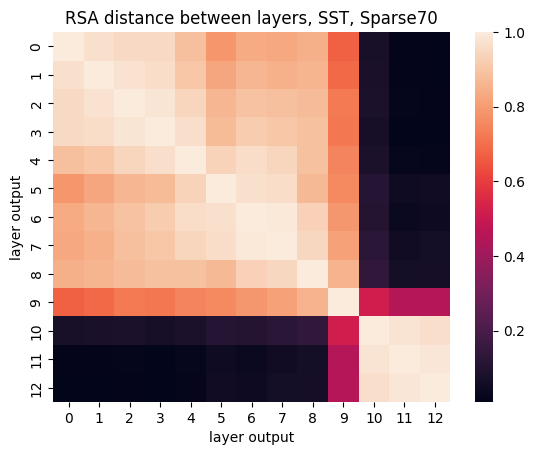}
       %\ \caption{}
         %\label{fig:yelp3}
    \end{subfigure}\hfill
    \begin{subfigure}{.20\linewidth}
        \includegraphics[width=\linewidth]{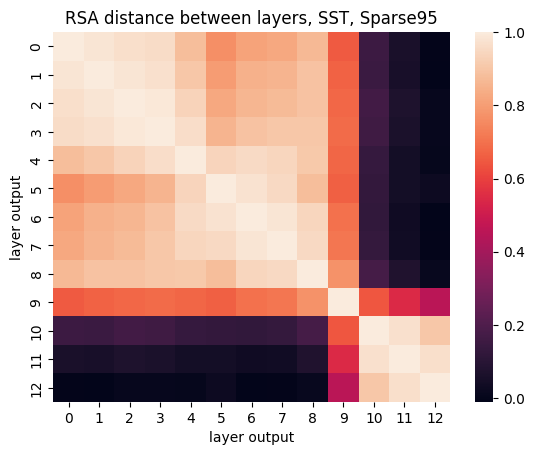}
       %\ \caption{}
        %\label{fig:yelp4}
    \end{subfigure}
%     \caption{\small{Pairwise Euclidean distance on SST2, across BERT models. (a) is pre-trained BERT and (b)-(e) are fine tuned BERT models with increasing sparsity (dense, $25\%$, $70\%$ and $95\%$ sparsity).}}
%     \label{fig:sst_euc}
% \end{figure}
% \begin{figure}[h]
\newline
 \begin{subfigure}{.20\linewidth}
        \includegraphics[width=\linewidth]{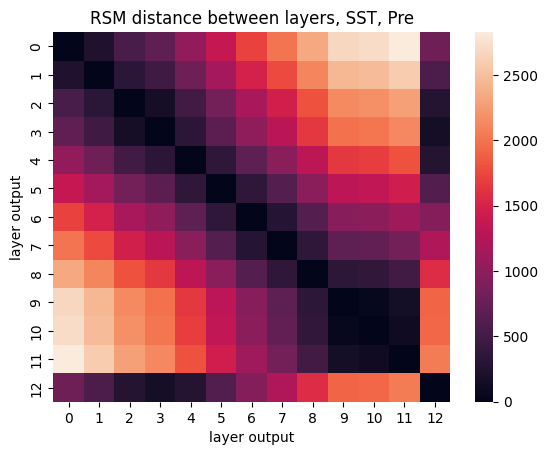}
        %\\caption{}
         %\label{fig:yelp0}
    \end{subfigure}\hfill
    \begin{subfigure}{.20\linewidth}
        \includegraphics[width=\linewidth]{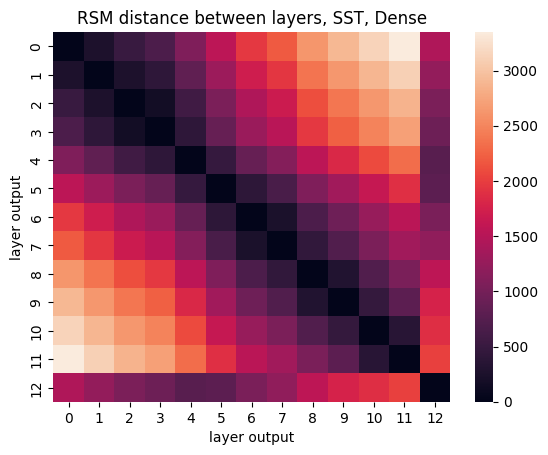}
        %\\caption{}
         %\label{fig:yelp1}
    \end{subfigure}\hfill
    \begin{subfigure}{.20\linewidth}
        \includegraphics[width=\linewidth]{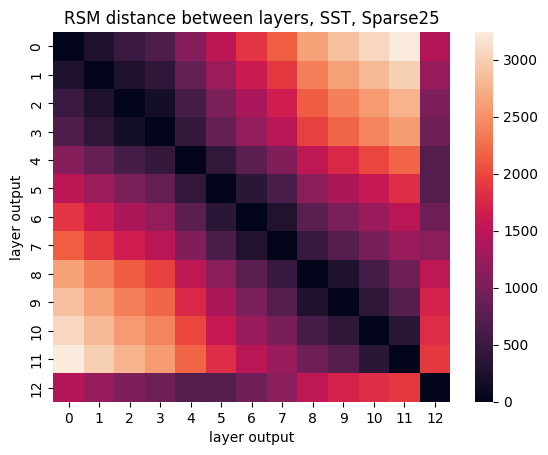}
        %\\caption{}
         %\label{fig:yelp2}
    \end{subfigure}\hfill
    \begin{subfigure}{.20\linewidth}
        \includegraphics[width=\linewidth]{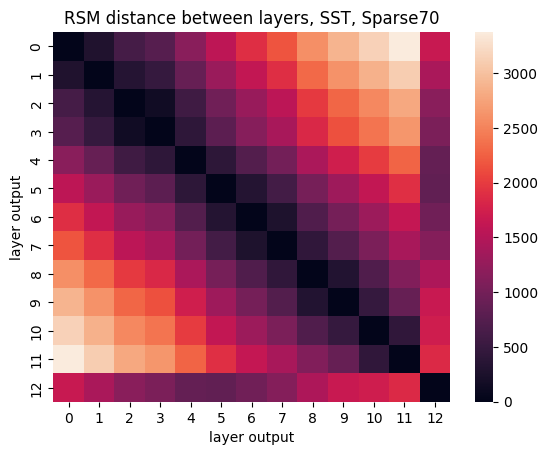}
        %\\caption{}
         %\label{fig:yelp3}
    \end{subfigure}\hfill
    \begin{subfigure}{.20\linewidth}
        \includegraphics[width=\linewidth]{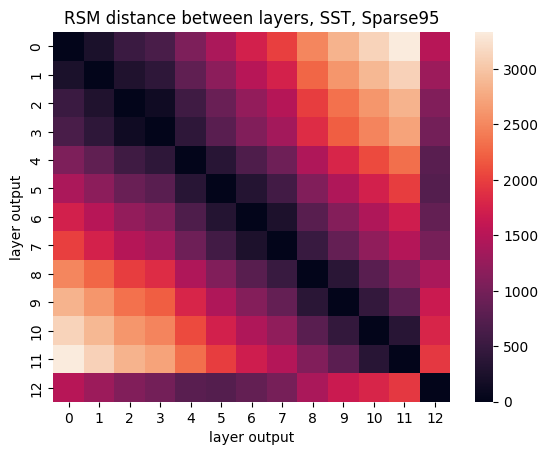}
       %\ \caption{}
        %\label{fig:yelp4}
    \end{subfigure}
\newline
  \begin{subfigure}{.20\linewidth}
        \includegraphics[width=\linewidth]{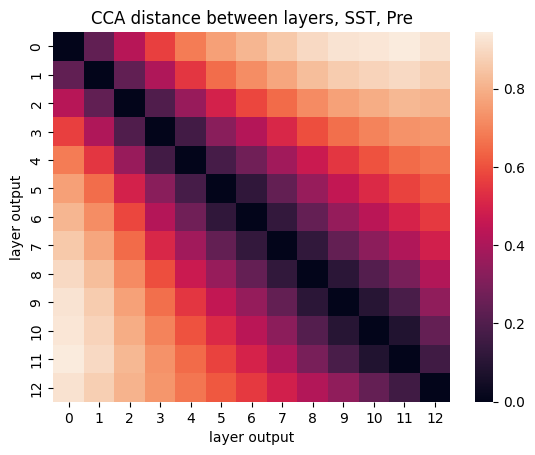}
       %\ \caption{}
        %\label{fig:yelp0}
    \end{subfigure}\hfill
    \begin{subfigure}{.20\linewidth}
        \includegraphics[width=\linewidth]{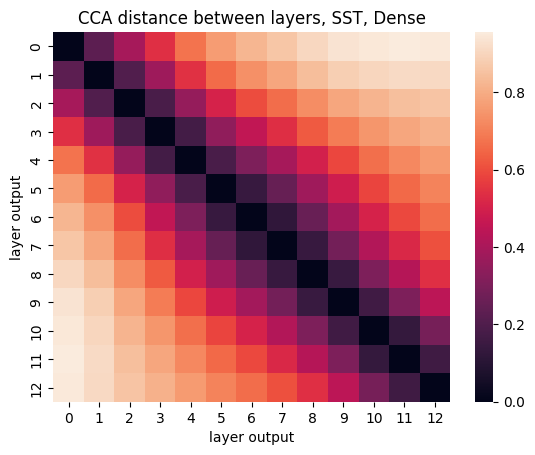}
        %\\caption{}
         %\label{fig:yelp1}
    \end{subfigure}\hfill
    \begin{subfigure}{.20\linewidth}
        \includegraphics[width=\linewidth]{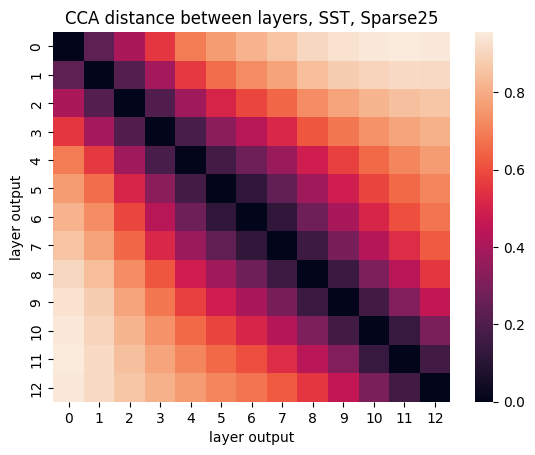}
       %\ \caption{}
         %\label{fig:yelp2}
    \end{subfigure}\hfill
    \begin{subfigure}{.20\linewidth}
        \includegraphics[width=\linewidth]{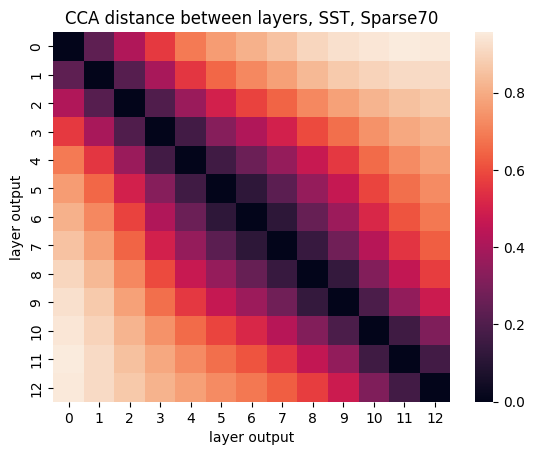}
        %\\caption{}
         %\label{fig:yelp3}
    \end{subfigure}\hfill
    \begin{subfigure}{.20\linewidth}
        \includegraphics[width=\linewidth]{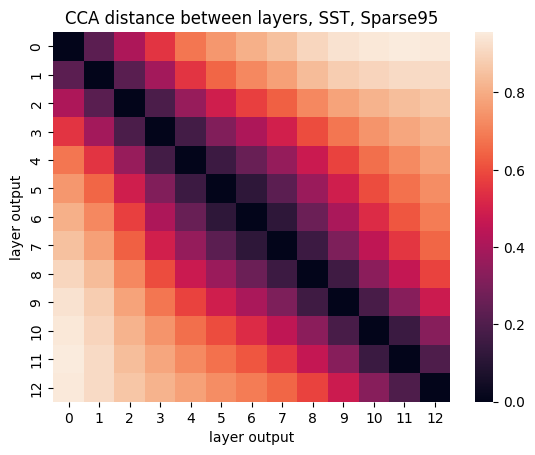}
        %\\caption{}
         %\label{fig:yelp4}
    \end{subfigure}
    \newline

    \begin{subfigure}{.20\linewidth}
        \includegraphics[width=\linewidth]{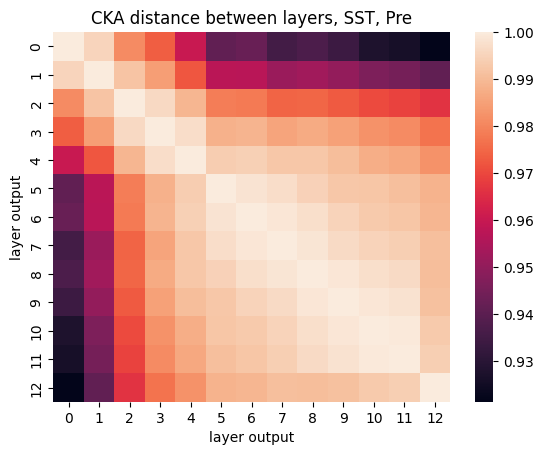}
         % \caption{Pretrained}
        % \label{fig:sst0}
    \end{subfigure}\hfill
    \begin{subfigure}{.20\linewidth}
        \includegraphics[width=\linewidth]{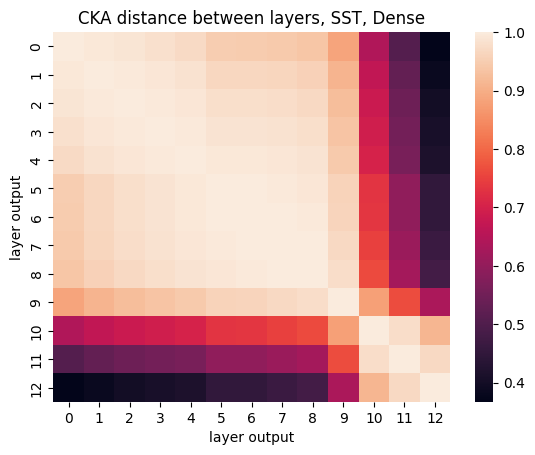}
        % \caption{Dense}
      %   \label{fig:sst1}
    \end{subfigure}\hfill
    \begin{subfigure}{.20\linewidth}
        \includegraphics[width=\linewidth]{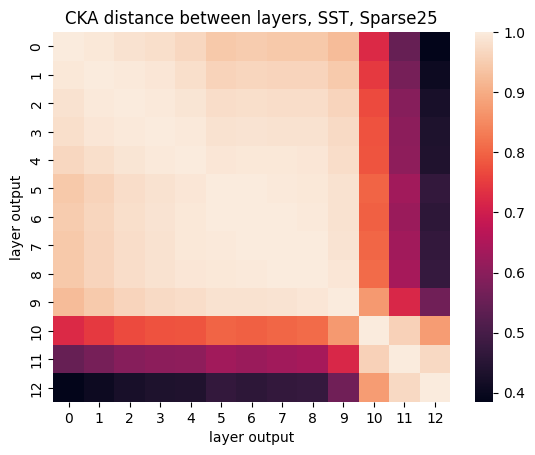}
        % \caption{25\% Sparse}
        % \label{fig:sst2}
    \end{subfigure}\hfill
    \begin{subfigure}{.20\linewidth}
        \includegraphics[width=\linewidth]{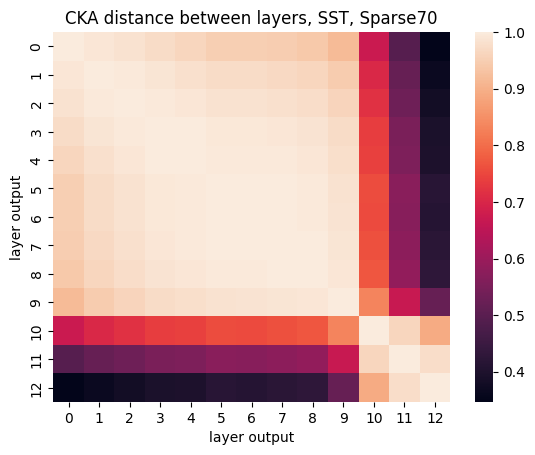}
         %\caption{70\% Sparse}
       %  \label{fig:sst3}
    \end{subfigure}\hfill
    \begin{subfigure}{.20\linewidth}
        \includegraphics[width=\linewidth]{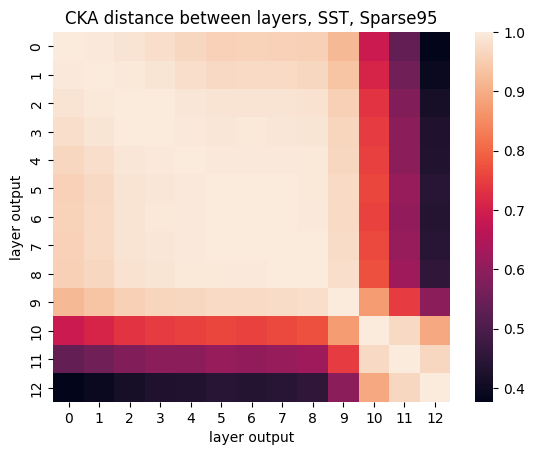}
        % \caption{95\% Sparse}
       %  \label{fig:sst4}
    \end{subfigure}
        \newline
    \begin{subfigure}{.20\linewidth}
        \includegraphics[width=\linewidth]{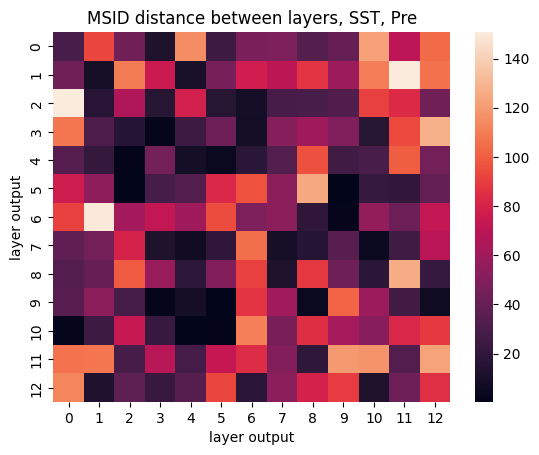}
       %\ \caption{}
        %\label{fig:yelp0}
    \end{subfigure}\hfill
    \begin{subfigure}{.20\linewidth}
        \includegraphics[width=\linewidth]{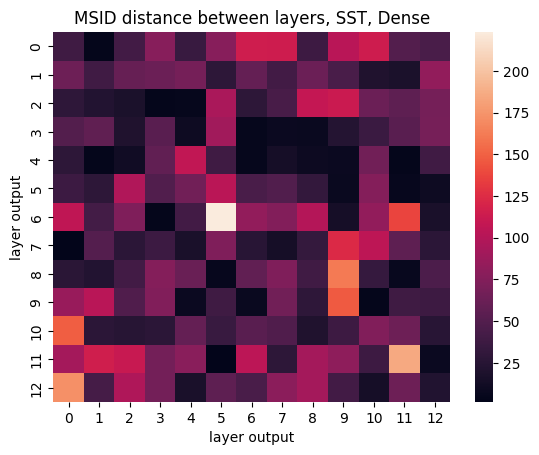}
        %\\caption{}
         %\label{fig:yelp1}
    \end{subfigure}\hfill
    \begin{subfigure}{.20\linewidth}
        \includegraphics[width=\linewidth]{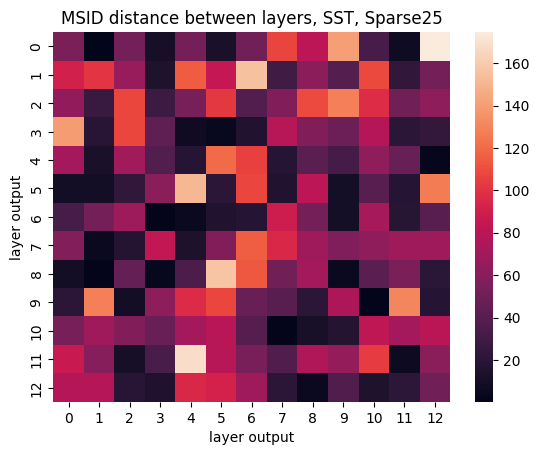}
       %\ \caption{}
         %\label{fig:yelp2}
    \end{subfigure}\hfill
    \begin{subfigure}{.20\linewidth}
        \includegraphics[width=\linewidth]{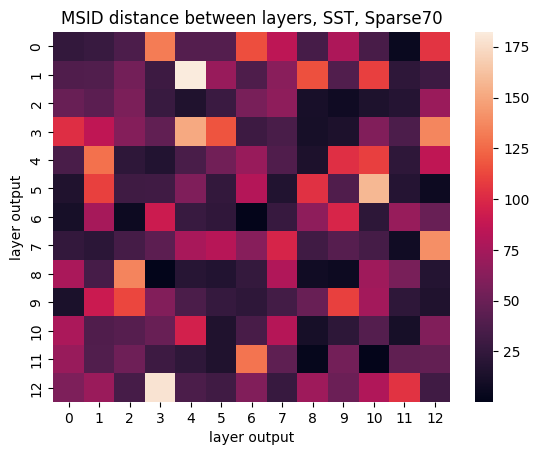}
        %\\caption{}
         %\label{fig:yelp3}
    \end{subfigure}\hfill
    \begin{subfigure}{.20\linewidth}
        \includegraphics[width=\linewidth]{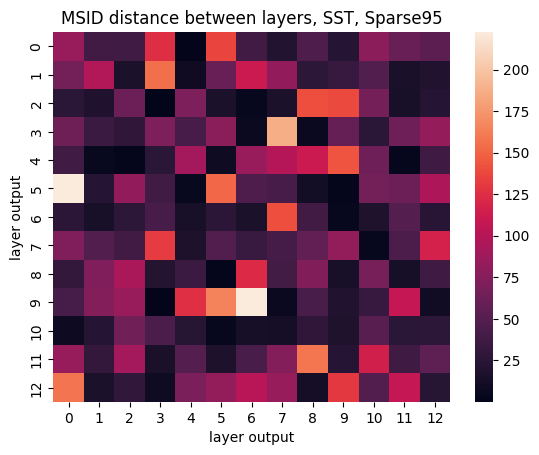}
        %\\caption{}
         %\label{fig:yelp4}
           \end{subfigure}\hfill
               \newline
  \begin{subfigure}{.20\linewidth}
        \includegraphics[width=\linewidth]{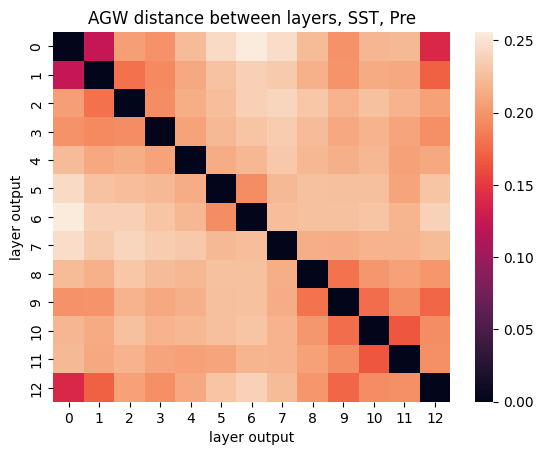}
       %\ \caption{}
        %\label{fig:yelp0}
    \end{subfigure}\hfill
    \begin{subfigure}{.20\linewidth}
        \includegraphics[width=\linewidth]{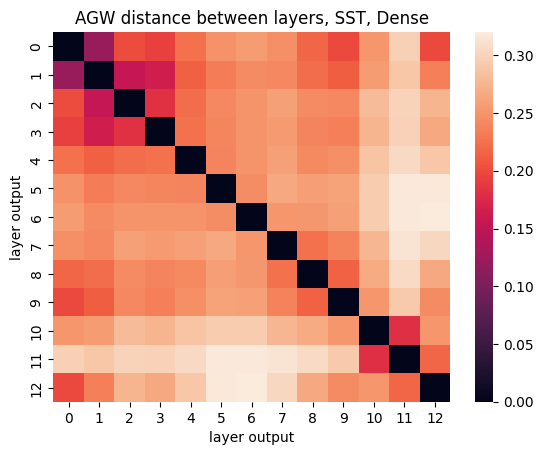}
        %\\caption{}
         %\label{fig:yelp1}
    \end{subfigure}\hfill
    \begin{subfigure}{.20\linewidth}
        \includegraphics[width=\linewidth]{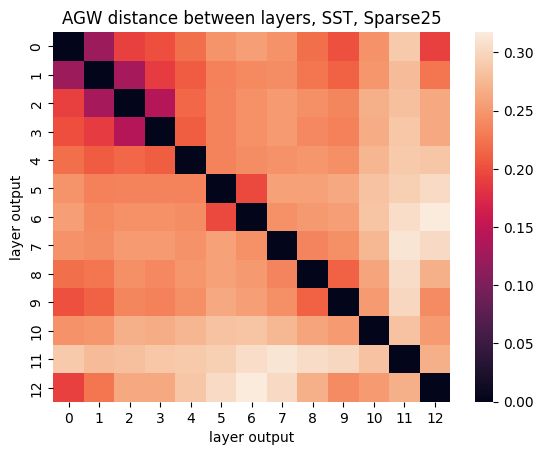}
       %\ \caption{}
         %\label{fig:yelp2}
    \end{subfigure}\hfill
    \begin{subfigure}{.20\linewidth}
        \includegraphics[width=\linewidth]{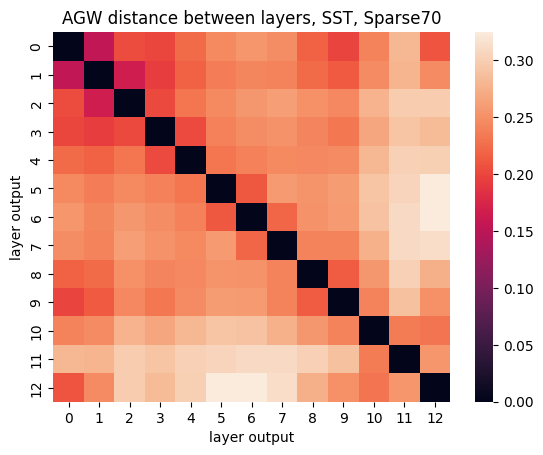}
        %\\caption{}
         %\label{fig:yelp3}
    \end{subfigure}\hfill
    \begin{subfigure}{.20\linewidth}
        \includegraphics[width=\linewidth]{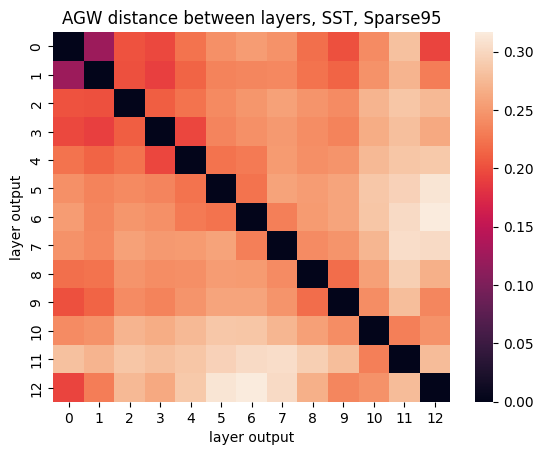}
        %\\caption{}
         %\label{fig:yelp4}
    \end{subfigure}
             \newline
    \begin{subfigure}{.20\linewidth}
        \includegraphics[width=\linewidth]{figs/gw_SST_pre.png}
         \caption{Pretrained}
        % \label{fig:sst0}
    \end{subfigure}\hfill
    \begin{subfigure}{.20\linewidth}
        \includegraphics[width=\linewidth]{figs/gw_SST_Dense.png}
        \caption{Dense}
        %\label{fig:sst1}
    \end{subfigure}\hfill
    \begin{subfigure}{.20\linewidth}
        \includegraphics[width=\linewidth]{figs/gw_SST_sparse25.png}
        \caption{25\% Sparse}
       % \label{fig:sst2}
    \end{subfigure}\hfill
    \begin{subfigure}{.20\linewidth}
        \includegraphics[width=\linewidth]{figs/gw_SST_sparse70.png}
        \caption{70\% Sparse}
      % \label{fig:sst3}
    \end{subfigure}\hfill
    \begin{subfigure}{.20\linewidth}
        \includegraphics[width=\linewidth]{figs/gw_SST_sparse95.png}
        \caption{95\% Sparse}
        %\label{fig:sst4}
    \end{subfigure}
      \caption{\small{More Pairwise distances on SST dataset, across different BERT models. Different Rows:  \textit{RSA, RSM, CCA, CKA,  MSID, AGW, and the proposed GW distance}, from top to bottom. Different columns: first column is the pre-trained BERT and the rest are fine tuned BERT models with increasing sparsity (dense, $25\%$, $70\%$ and $95\%$ sparsity).}}
    \label{fig:sst_w_new}
\end{figure}

More baselines are included in Figure~\ref{fig:sst_w_new}, as they do not all fit into the one page. Overall, RSA and CKA identify block structures but with larger 2nd block.

\section{Model Pruning/Compressing}
\label{appendix:pruning}

Another another potential application beside freezing-and-fine-tuning specific transformer blocks, we study the problem of model compress or pruning with the discovered subnetworks.

For each of desired block sizes, we take the original pre-trained BERT and only use the first $n = \{12, 8, 4, 2, 1, 0 \}$ transformer blocks while discarding the rest. Note that $n = 12$ means we use all the transformer blocks, resulting the same BERT model. $n=0$, on the other hand, means that we only use a (linear) classifier layer (after embedding layer) to predict the class label. The results are shown in Table~\ref{table:compress}. As a reminder, GW distance suggest the last 4 blocks in YELP (see Figure~\ref{fig:yelp_w}) and the last 2 blocks in SST (see Figure~\ref{fig:sst_w}) are mostly different, which is marked by star $(*)$ in the table. It shows that by using a limited number of layers, we can achieve similar performance with the full 12 block model, with $0.01\%$ and $0.54\%$ differences in YELP and SST, respectively. Using one fewer  transformer block can risk much worse reduction of performance, with $0.10\%$ and $8.60\%$ differences (about 10 times worse performance reduction). 

\begin{table}[h]
\small
  \caption{Accuracy of pruning BERT with a smaller number of blocks on YELP and SST. N denotes the number of transformer blocks in the new BERT models. }
%  \label{table:fine-tune}
  \centering
 \begin{tabular}{l||c|c|c|c|c|c}
    \toprule
Number of Transformer Blocks & 12 (all)  & 8  & 4& 2 & 1 & 0 (only classifier) \\
\hline
YELP &  $97.87$   & 97.87 & $97.86^{*}$  &97.76 & 97.11 & 60.3 \\
\hline
SST  &  $92.40$   & 90.25 & 90.25  & $91.86^{*}$ & 83.26 & 50.92 \\
% \midrule
% Model 1&{GW-D for}  & Top Similar Layers& $D_{\text{min}}=$ \\
%  &     $c_1$  &  $z_1^2$ & $0.02$ \\
%  &   $c_2$ &  $z_1^2$ & $0.02$\\
% &     $c$ & $z_1^2$, 6 others, and Resid-Post$^3$  & $0.02$ \\
    \bottomrule
  \end{tabular}
  \label{table:compress}
\end{table}

\section{GW Distance with Different Random Seeds}
\begin{figure}[h]
    \begin{subfigure}{.30\linewidth}
        \includegraphics[width=\linewidth]{figs/gw_Yelp_dense.png}
        \caption{Dense}
       % \label{fig:yelp1}
    \end{subfigure}\hfill
    \begin{subfigure}{.30\linewidth}
        \includegraphics[width=\linewidth]{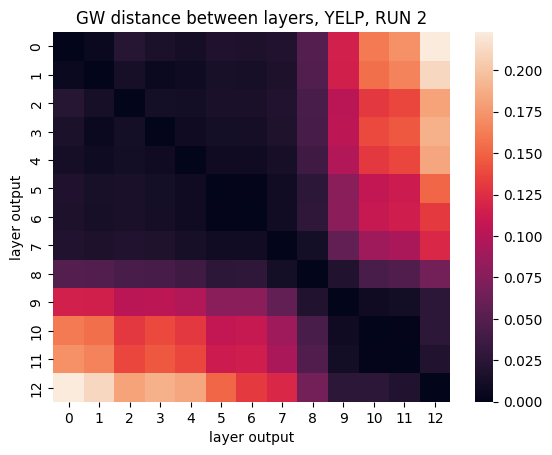}
        \caption{25\% Sparse}
     %   \label{fig:yelp2}
    \end{subfigure}\hfill
    \begin{subfigure}{.30\linewidth}
        \includegraphics[width=\linewidth]{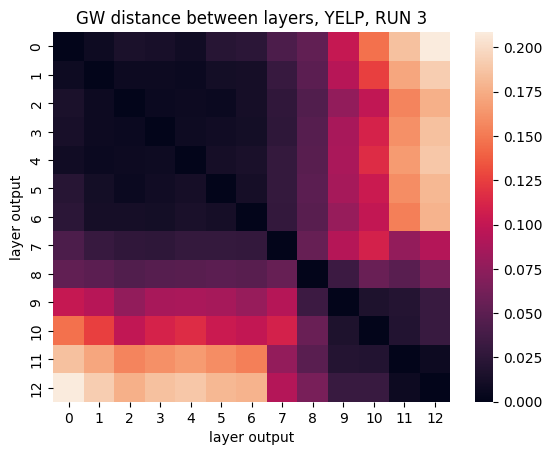}
        \caption{70\% Sparse}
      %  \label{fig:yelp3}
    \end{subfigure}\hfill
    \caption{\small{Pairwise GW (layer) distances on Yelp, across  BERT models trained with 3 different random seeds.  As one can be seen,  the (functional) sub-network blocks stay rather consistent with different seeds even though there is some variations among the models.}}
    \label{fig:yelp_seed}
\end{figure}

Neural networks initialized with different random seeds can converge to distinct  representations \citep{li2015convergent,morcos2018insights,kornblith2019similarity}, even when their performance is comparable. To study the impact of initialization seeds  on the learned representations, we train the same BERT model on YELP datasets with different seeds, with identical hyperparameters for a total of 27,000 iterations. As shown in Figure~\ref{fig:yelp_seed}, while the learned representations  vary across seeds, but the general block structures  remain consistent when analyzed using GW distances. 

\section{Computer Vision Application: CIFAR-10 Datasets}
\label{appendix:vision}
In addition to the attention-based architectures,  we  also test our approach on  ResNet9, 
 a popular convolutional neural network architecture\citep{he2016deep,park2023trak}. We compare a randomly initialized ResNet9 and a trained model on CIFAR 10 image dataset  CIFAR-10 \citep{krizhevsky2009learning}, achieving 91.63\% accuracy on the test data. CIFAR-10 dataset consists of 60000 32x32 color images in 10 image classes, with 6000 images per class. There are 50000 training images and 10000 testing images. The classes are completely mutually exclusive. ResNet is a convolutional neural network with many residual connections. ResNet9 specifically contains 9 convolution layers, each followed by BatchNorm and ReLU activation. The exact details of the ResNet 9 is listed in Table~\ref{table:location_resnet}.

\begin{figure}
\centering
 \vskip -3mm
  \begin{subfigure}{.25\linewidth}
        \includegraphics[width=\linewidth]{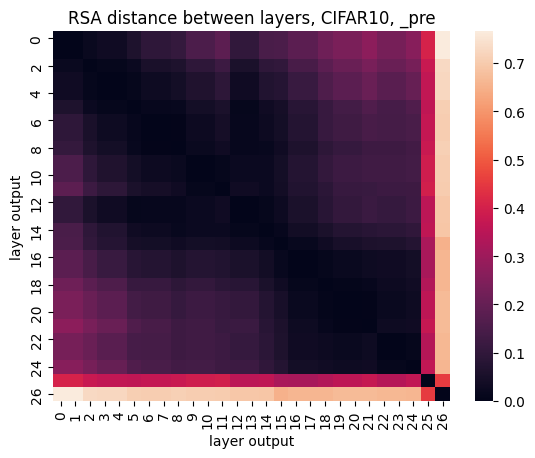}
        %\caption{}
         %\label{fig:yelp0}
    \end{subfigure}
    \begin{subfigure}{.25\linewidth}
        \includegraphics[width=\linewidth]{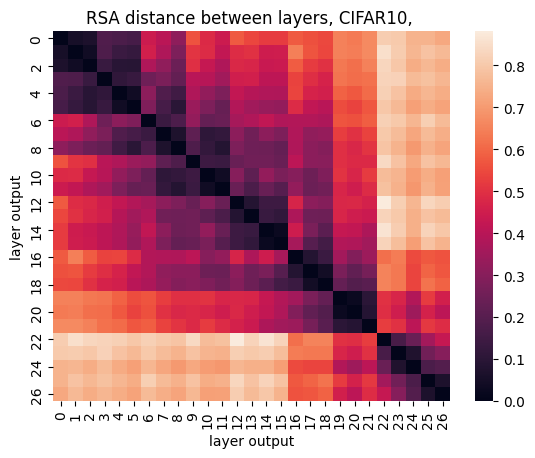}
        %\caption{}
         %\label{fig:yelp1}
    \end{subfigure}
\newline
  \begin{subfigure}{.25\linewidth}
        \includegraphics[width=\linewidth]{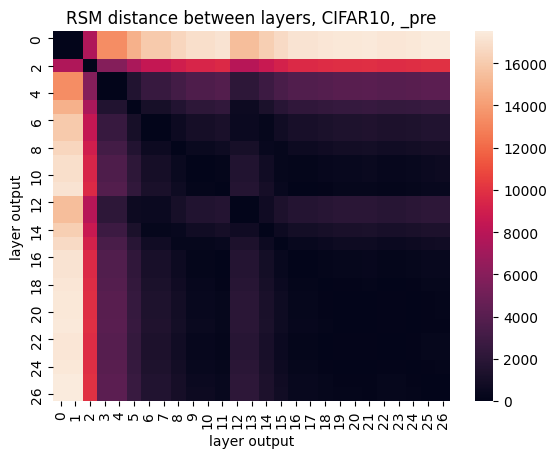}
        %\caption{}
        %\label{fig:yelp0}
    \end{subfigure}
    \begin{subfigure}{.25\linewidth}
        \includegraphics[width=\linewidth]{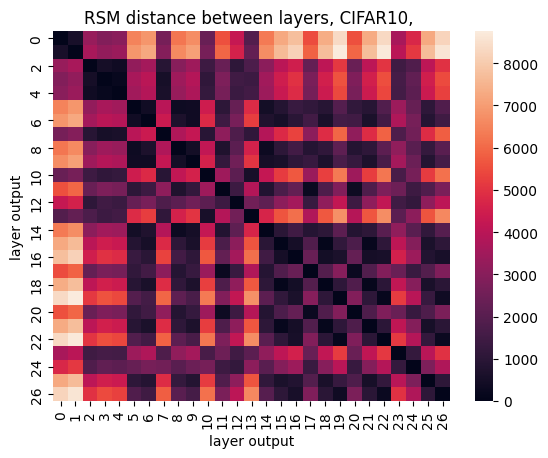}
        %\caption{}
         %\label{fig:yelp1}
    \end{subfigure}
\newline
  \begin{subfigure}{.25\linewidth}
        \includegraphics[width=\linewidth]{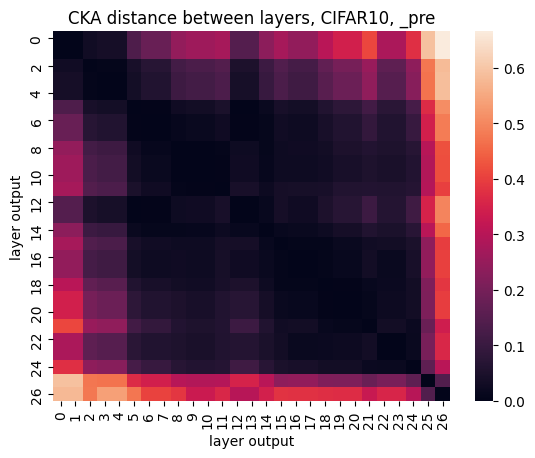}
        %\caption{}
        %\label{fig:yelp0}
    \end{subfigure}
    \begin{subfigure}{.25\linewidth}
        \includegraphics[width=\linewidth]{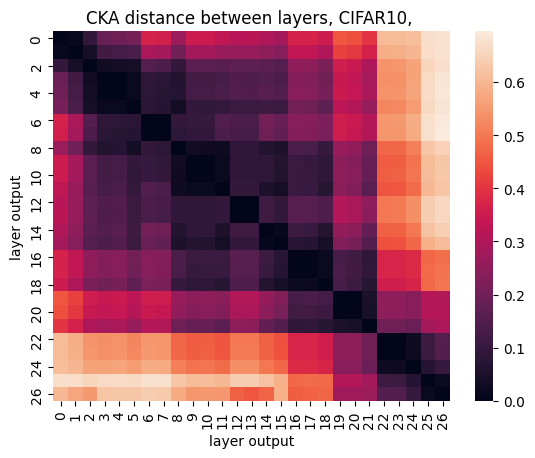}
        %\caption{}
         %\label{fig:yelp1}
    \end{subfigure}
\newline
  \begin{subfigure}{.25\linewidth}
        \includegraphics[width=\linewidth]{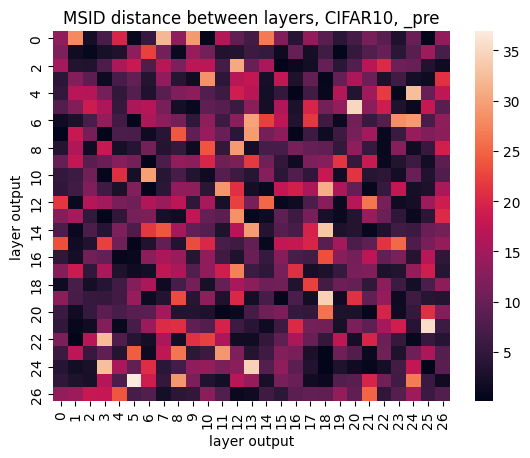}
        %\caption{}
        %\label{fig:yelp0}
    \end{subfigure}
    \begin{subfigure}{.25\linewidth}
        \includegraphics[width=\linewidth]{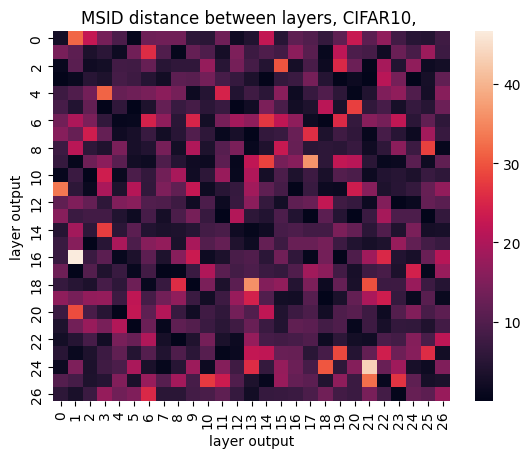}
        %\caption{}
         %\label{fig:yelp1}
    \end{subfigure}
\newline
  \begin{subfigure}{.25\linewidth}
        \includegraphics[width=\linewidth]{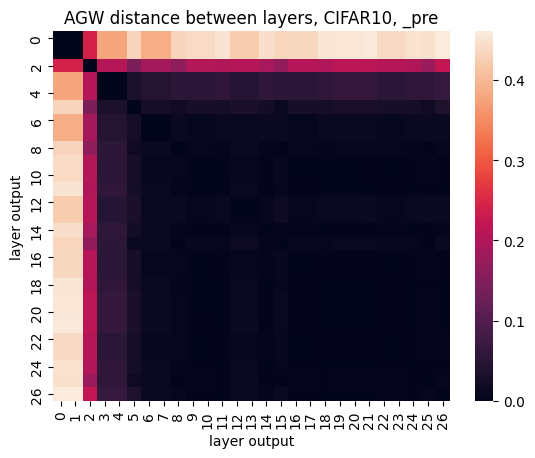}
        %\caption{}
        %\label{fig:yelp0}
    \end{subfigure}
    \begin{subfigure}{.25\linewidth}
        \includegraphics[width=\linewidth]{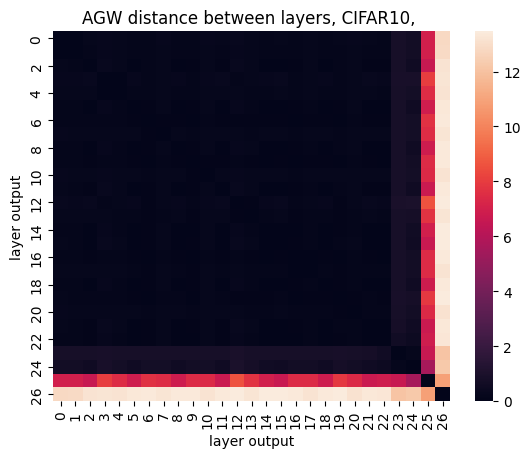}
        %\caption{}
         %\label{fig:yelp1}
    \end{subfigure}
\newline
  \begin{subfigure}{.25\linewidth}
        \includegraphics[width=\linewidth]{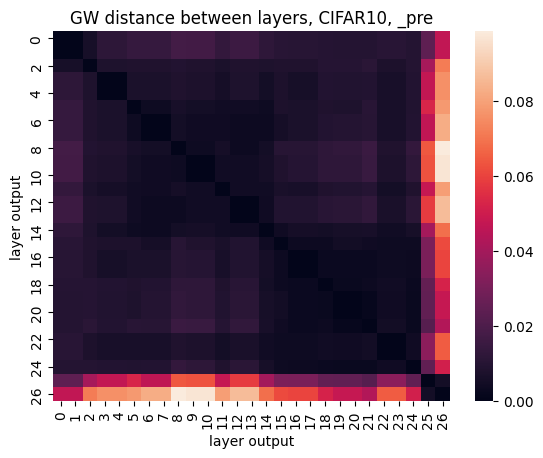}
        %\caption{}
        %\label{fig:yelp0}
    \end{subfigure}
    \begin{subfigure}{.25\linewidth}
        \includegraphics[width=\linewidth]{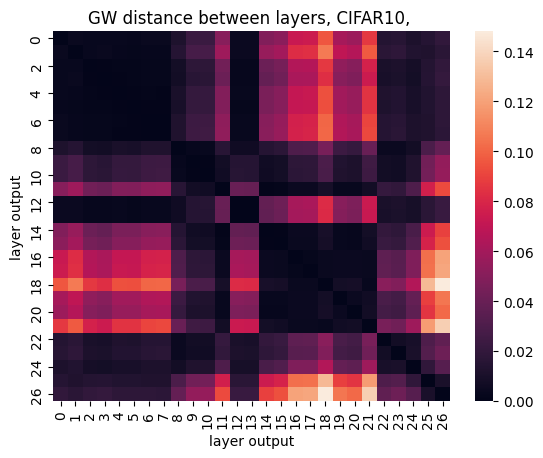}
    \end{subfigure}
    \newline
    \caption{\small{Pairwise (layer) distances on CIFAR-10, across different BERT models. Different Rows:  \textit{RSA, RSM, CKA, MSID, AGW, and the proposed GW distance}, from top to bottom. Different columns: the first column is the pre-trained ResNet9, and the 2nd column contains the fine tuned ResNet models.}}
    \label{fig:cifar_w_new}

\end{figure}

We show the pairwise distance of all layers in consideration using all methods,  that can handle difference dimensions of inputs, in Figure~\ref{fig:cifar_w_new}. The first column shows results from randomly initialized pre-trained models, and the second columns shows results from the trained ResNet. Pre-trained models generally do not show clear sub-network structures, while the trained models shows differences across layers. RSA, RSM, and CKA show progressive changes over the network layers, which is not too informative. AGW only shows the last a few layers contain significant changes, and MSID distance does not contain clear patterns. In comparison, GW distance shows clear division of 3 or 4 subnetworks.

%%%%%%%%%%%%%%%%%%%%%%%%%%%%%%%%%%%%%%%%%%%%%%%%%%%%%%%%%%%%%%%%%%%%%%%%%%%%%%%
%%%%%%%%%%%%%%%%%%%%%%%%%%%%%%%%%%%%%%%%%%%%%%%%%%%%%%%%%%%%%%%%%%%%%%%%%%%%%%%

\end{document}